\pdfoutput=1

\documentclass[11pt]{article}

\usepackage[final]{acl}

\usepackage{times}
\usepackage{latexsym}

\usepackage[T1]{fontenc}

\usepackage[utf8]{inputenc}

\usepackage{microtype}

\usepackage{inconsolata}

\usepackage{graphicx}

\usepackage{xcolor}
\usepackage{adjustbox}
\usepackage{amsmath}
\usepackage{multirow}
\usepackage{makecell}
\usepackage{booktabs}
\usepackage{makecell}
\usepackage{amsfonts}
\usepackage{dsfont}
\usepackage{caption}
\usepackage{colortbl}

\usepackage{etoc}
\DeclareUnicodeCharacter{2009}{\,}

\usepackage[edges]{forest}
\usepackage{tcolorbox}

\usepackage{tabularx}
\usepackage{amssymb} 

\usepackage{array,booktabs,tabularx,ragged2e}
\newcolumntype{Y}{>{\RaggedRight\arraybackslash}X}

\usepackage{hyperref}

\usepackage{algorithm}
\usepackage{algorithmic}

\definecolor{orangenode}{RGB}{255,200,180}
\definecolor{greennode}{RGB}{220,240,220}
\definecolor{bluenode}{RGB}{200,220,255}

\definecolor{lightblue}{HTML}{E0F2F7}
\definecolor{mycustomblue}{HTML}{003366}
\definecolor{tablehead}{HTML}{2C5E7B}
\definecolor{tablerow}{HTML}{F0F8FF}

\definecolor{ngreen}{HTML}{D5E8D4}
\definecolor{nblue}{HTML}{DAE8FC}
\definecolor{npurple}{HTML}{E1D5E7}

\definecolor{aclblue}{RGB}{214, 230, 245} 
\definecolor{aclgray}{RGB}{247, 247, 247} 
\definecolor{deepblue}{RGB}{43, 75, 126}  

\newcommand{\bolditalic}[1]{\textit{\textbf{#1}}}

\definecolor{hidden-red}{RGB}{205, 44, 36}
\definecolor{hidden-blue}{RGB}{194,232,247}
\definecolor{hidden-orange}{RGB}{243,202,120}
\definecolor{hidden-green}{RGB}{34,139,34}
\definecolor{hidden-pink}{RGB}{255,245,247}
\definecolor{hidden-black}{RGB}{20,68,106}
\definecolor{purple}{RGB}{144,153,196}
\definecolor{yellow}{RGB}{255,228,123}
\definecolor{hidden-yellow}{RGB}{255,248,203}
\definecolor{tkcolor}{RGB}{224,223,255}
\definecolor{mycustomblue}{rgb}{0, 0.40, 0.75}

\tikzstyle{my-box}=[
rectangle,
draw=hidden-black,
rounded corners,
text opacity=1,
minimum height=1.5em,
minimum width=5em,
inner sep=2pt,
align=center,
fill opacity=.5,
]
\tikzstyle{leaf}=[
my-box,
minimum height=1.5em,
fill=hidden-red!20,
text=black,
align=left,
font=\normalsize,
inner xsep=5pt,
inner ysep=4pt,
text width=46em,
]
\tikzstyle{leaf2}=[
my-box,
minimum height=1.5em,
fill=hidden-green!20,
text=black,
align=left,
font=\normalsize,
inner xsep=5pt,
inner ysep=4pt,
text width=46em,
]
\tikzstyle{leaf3}=[
my-box,
minimum height=1.5em,
fill=yellow!32,
text=black,
align=left,
font=\normalsize,
inner xsep=5pt,
inner ysep=4pt,
align=left,
text width=46em,
]
\tikzstyle{leaf4}=[
my-box,
minimum height=1.5em,
fill=hidden-blue!57,
text=black,
align=left,
font=\normalsize,
inner xsep=5pt,
inner ysep=4pt,
text width=46em,
]
\tikzstyle{leaf5}=[
my-box,
minimum height=1.5em,
fill=mycustomblue!15,
text=black,
align=left,
font=\normalsize,
inner xsep=5pt,
inner ysep=4pt,
]
\tikzstyle{leaf6}=[
my-box,
minimum height=1.5em,
fill=purple!30,
text=black,
align=left,
font=\normalsize,
inner xsep=5pt,
inner ysep=4pt,
]

\title{\textit{When LLM Meets Tree Search}: A Systematic View of Inference as Search in Large Language Models}

\author{
  \textbf{Jiaqi Wei}\textsuperscript{1,2*},
  \textbf{Xiang Zhang}\textsuperscript{3*},
  \textbf{Yuejin Yang}\textsuperscript{4,2*},
  \textbf{Wenxuan Huang}\textsuperscript{4,2*},
  \textbf{Juntai Cao}\textsuperscript{3*},
  \textbf{Sheng Xu}\textsuperscript{4,2}, \\
  \textbf{Xiang Zhuang}\textsuperscript{2},
  \textbf{Zhangyang Gao}\textsuperscript{2},
  \textbf{Muhammad Abdul-Mageed}\textsuperscript{3},
  \textbf{Laks V.S. Lakshmanan}\textsuperscript{3}, \\
  \textbf{Chenyu You}\textsuperscript{5},
  \textbf{Wanli Ouyang}\textsuperscript{6},
  \textbf{Siqi Sun}\textsuperscript{4,2}
\\
  \textsuperscript{1}Zhejiang University,
  \textsuperscript{2}Shanghai AI Laboratory,
  \textsuperscript{3}University of British Columbia, \\
  \textsuperscript{4}Fudan University, 
  \textsuperscript{5}Stony Brook University,
  \textsuperscript{6}The Chinese University of Hong Kong
\\
  \textsuperscript{*}Equal Contribution
\\
  \small{
    \textbf{Correspondence:} \href{mailto: jiaqi.wei@zju.edu.cn}{jiaqi.wei@zju.edu.cn}, \href{mailto: siqisun@fudan.edu.cn}{siqisun@fudan.edu.cn}
  }
}

\begin{document}
\maketitle

\begin{abstract}
As pretraining scaling laws approach saturation, \textbf{Test-Time Scaling (TTS)} has emerged as an important
direction for improving reasoning by allocating inference-time compute to a fixed model prior.
Viewed at a high level, TTS reframes inference as \emph{search} over a space of partial reasoning states.
While Chain-of-Thought (CoT) exposes intermediate steps, common instantiations rely on
\emph{single-trajectory decoding}, limiting recovery from early errors and exploration.
This survey systematizes recent progress in \textbf{tree-search-based reasoning}, viewing inference as
\emph{instance-specific optimization} rather than decoding.
We trace the evolution from uninformed search to Monte Carlo Tree Search (MCTS), highlighting how
sampling-based control supports principled exploration--exploitation trade-offs.
To unify a fragmented literature, we introduce a \textbf{Unified Design Space} spanning search topology,
evaluation signals, and control dynamics, and advocate a \textbf{standardized compute-reporting abstraction}
to make compute--accuracy trade-offs explicit and comparable.
\end{abstract}

\begin{figure*}[!t]
    \centering
    \includegraphics[width=\linewidth]{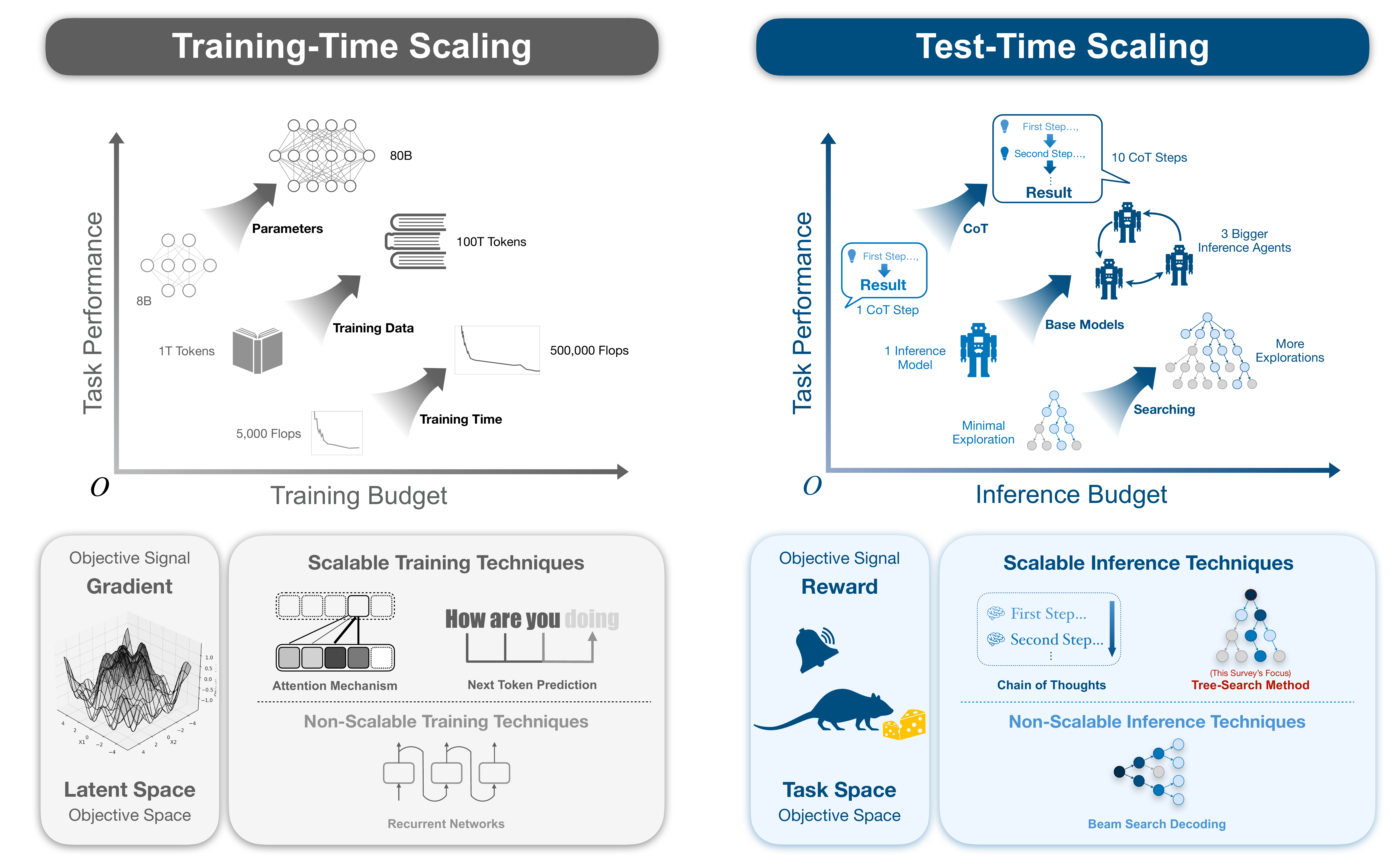}
    \caption{\textbf{Comparison of Training-time and Test-time Scaling}: Highlighting Budget Allocation, Techniques, and Their Impact on Task Performance.}
    \label{fig:scaling}
\end{figure*}

\section{Introduction}
As scaling laws for large language models (LLMs) enter a regime of diminishing returns
~\citep{kaplan2020scaling},
research attention is increasingly shifting from parameter growth toward
\emph{algorithmic strategies for reasoning}.
A central response is \textbf{Test-Time Scaling (TTS)}~\citep{brown2024large},
which allocates adaptive inference-time compute to improve problem solving from a fixed model prior.
At a high level, TTS reframes inference as \emph{search}: the system navigates a combinatorial space
of partial reasoning states under a task-defined objective.

Chain-of-Thought (CoT) prompting demonstrates that externalized reasoning can improve performance
~\citep{wei2022chain}, but most implementations rely on single-trajectory decoding, limiting recovery
from early errors and restricting exploration.
Recent work therefore revisits classical planning and tree search, maintaining a frontier of partial
solutions and reallocating compute based on intermediate feedback
~\citep{yao2023tree, hao-etal-2023-reasoning, wang2023selfconsistency}.
These \emph{search-augmented} methods form the algorithmic core of modern test-time reasoning.

Importantly, the same search mechanisms also enable \textbf{self-improvement}:
high-quality reasoning trajectories discovered through search (e.g., via MCTS) can be distilled into
training data or reward models, internalizing deliberative behavior into model parameters
~\citep{silver2017mastering, gulcehre2023reinforced, wan2024alphazerolike, guo2025deepseek}.
Tree search thus plays a dual role—as a transient inference-time optimizer and as a generator of
durable parametric knowledge.

Despite rapid progress, the literature remains fragmented across search strategies, reward or value
estimation, and evaluation protocols, obscuring general design principles~\citep{wei2025unifying}.
This survey provides a \textbf{Systematization of Knowledge (SoK)} for search-based reasoning in LLMs,
introducing a unified framework that connects test-time scaling and self-improvement through a shared
set of algorithmic components. Our goal is not to introduce new algorithms or benchmarks, but to consolidate existing methods under a shared conceptual and reporting framework.

\textbf{Our contributions are:}
\begin{itemize}
    \item \textbf{Unified Formalism:} A compact decomposition of search-based reasoning into
    \emph{search mechanism}, \emph{reward/value estimation}, and \emph{transition dynamics},
    clarifying their roles in both test-time optimization and parametric learning.
    \item \textbf{Component-Based Taxonomy:} A systematic organization of existing methods along
    search strategy, evaluation signal, and application paradigm.
    \item \textbf{Synthesis and Outlook:} A synthesis of empirical insights and open challenges in
    scaling deliberative search and designing effective reward signals.
\end{itemize}

\paragraph{Organization.} 
Section~\ref{sec:main_search_in_general_ai} reviews classical search paradigms and their relevance to long-horizon reasoning beyond greedy decoding. Section~\ref{sec:main_mcts} presents a unified framework for MCTS in LLMs, focusing on the integration of policy priors and value or reward signals. Section~\ref{sec:main_Informed} contrasts this with heuristic-guided search. Section~\ref{sec:main_evaluation_framework} discusses a lightweight compute-reporting framework for making test-time search results more interpretable across studies, and Section~\ref{sec:main_challenges} discusses open challenges and future directions in adaptive allocation and search-to-training distillation.

\section{Search in General AI}
\label{sec:main_search_in_general_ai}
Reasoning problems can be framed as search over trees or graphs, where large
branching factors make exhaustive exploration infeasible. Classical AI search
addresses this challenge by trading off exploration cost and solution quality.
We briefly summarize major paradigms to motivate LLM-based reasoning methods;
formal details are deferred to Appendix~\ref{appendix:Paradigms}.

\paragraph{Uninformed and Informed Search.}
Classical search ranges from \emph{uninformed} methods (e.g., BFS, DFS, UCS) to
\emph{heuristic} search, which incorporates domain knowledge via a heuristic
function $h(n)$. While uninformed methods serve as conceptual baselines, they scale
poorly in large reasoning spaces. Heuristic approaches such as A* and beam search
can reduce search cost when reliable heuristics exist, but designing accurate,
domain-general heuristics remains challenging for open-ended LLM reasoning.

\paragraph{Monte Carlo Tree Search (MCTS).}
Monte Carlo Tree Search (MCTS) offers a complementary paradigm that avoids explicit
heuristic design by learning value estimates through sampling. Originally developed
for game playing, MCTS has been adapted to single-agent reasoning by balancing
exploration and exploitation through accumulated search statistics. This makes
MCTS well suited to LLM reasoning tasks where evaluators are noisy, sparse, or only
available at terminal states.

\begin{figure*}[!t]
    \centering
    \includegraphics[width=\linewidth]{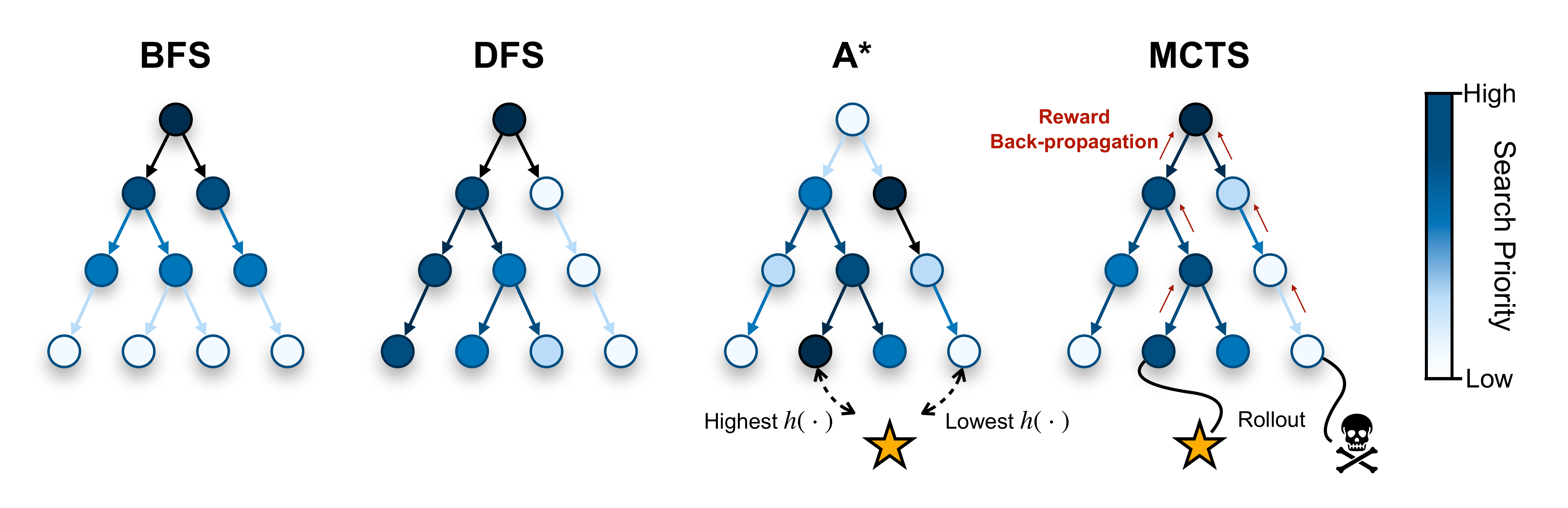}
    \caption{\textbf{A visual comparison of four fundamental tree search algorithms, where node color intensity represents search priority.} \textbf{BFS} explores exhaustively level by level, while \textbf{DFS} commits to a single path until a leaf is reached. In contrast, informed search like \textbf{A*} uses a heuristic function $h(\cdot)$ to prioritize nodes with the lowest estimated total cost, regardless of their depth. \textbf{MCTS} introduces a statistical approach, using simulated rollouts from leaf nodes and backpropagating the outcomes to dynamically guide the search toward high-reward regions of the tree.}
    \label{fig:search}
\end{figure*}

\begin{figure*}[!t]
    \centering
    \includegraphics[width=0.7\linewidth]{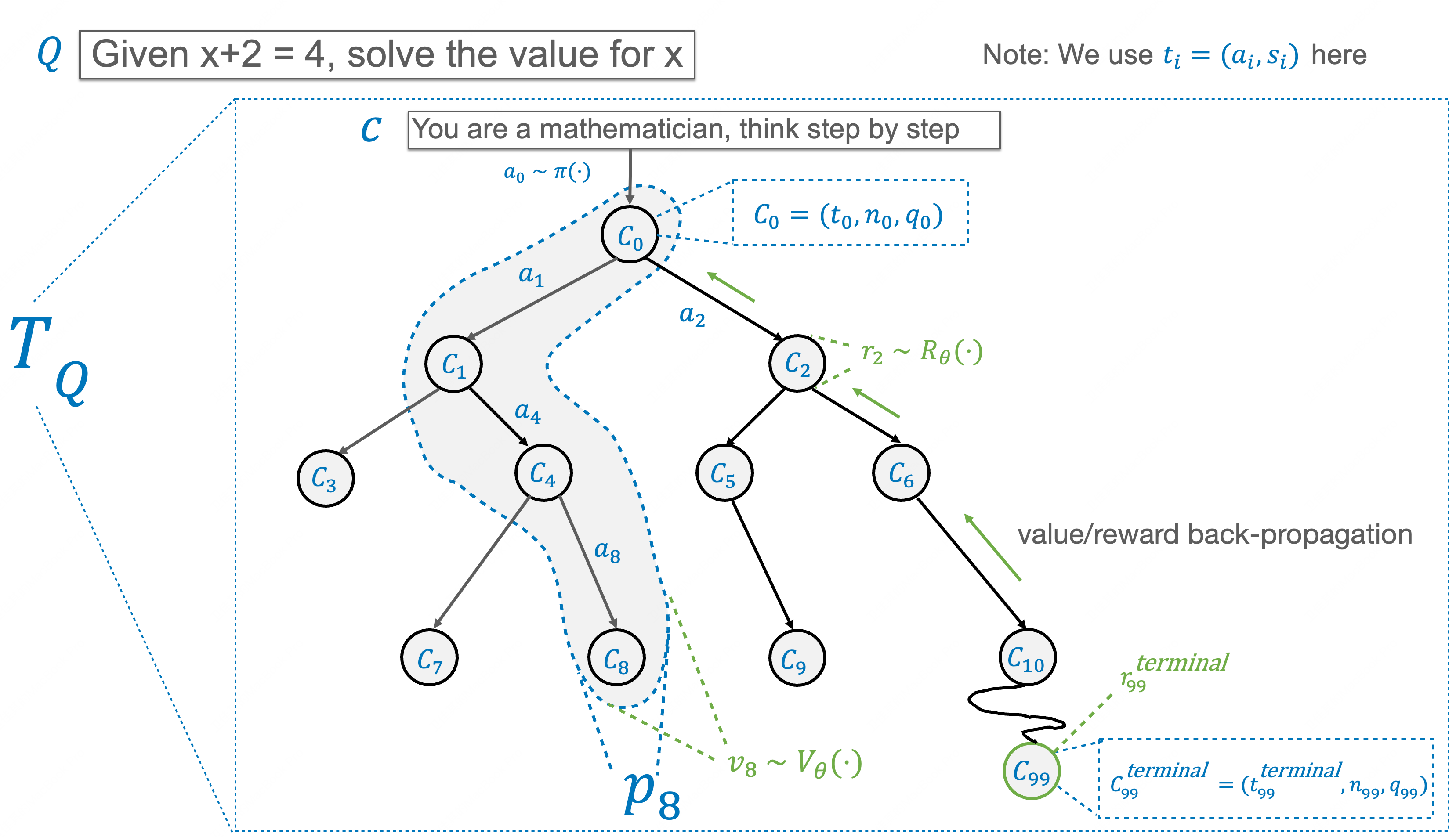}
    \caption{\textbf{Unified Notations for MCTS-Based Methods in LLM}.}
    \label{fig:notations}
\end{figure*}

\begin{figure*}[!t]
    \centering
    \includegraphics[width=1.0\linewidth]{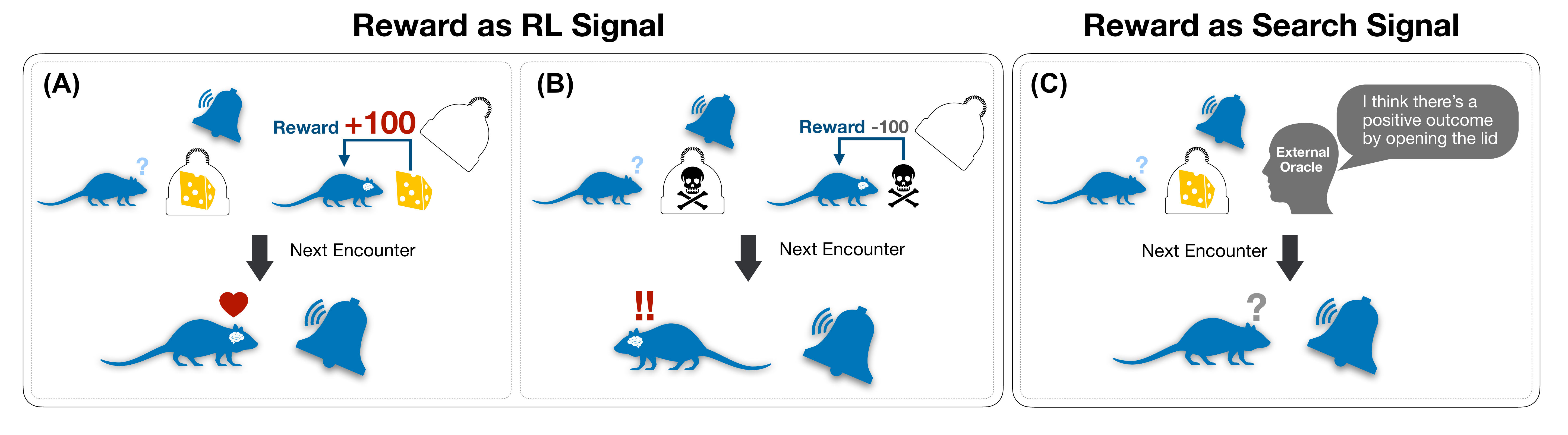}
    \caption{\textbf{Reward Design: Search vs. RL.} (A) In RL, a positive reward updates the agent's policy, making it more likely to repeat the action. (B) A negative reward also updates the policy, discouraging the behavior. The change is \textbf{durable}. (C) In search, an external oracle provides a reward signal to guide the current decision process without altering the agent's underlying parameters.}
    \label{fig:reward_design}
\end{figure*}

\paragraph{Reward as a Guiding Signal: Search vs. RL.}
\label{sec:rl_reward}
The notion of a ``reward'' plays a central role in both search and Reinforcement
Learning (RL), yet its function and implementation differ fundamentally, as
illustrated in Figure~\ref{fig:reward_design}. Although both are commonly referred
to simply as ``reward'' in the literature, they serve distinct purposes, and this
ambiguity can obscure the relationship between test-time planning and
training-time optimization. Clarifying this distinction is essential for the
unified framework developed in this paper; additional details are provided in
Appendix~\ref{appendix:Reward}.

\noindent \textbf{In Reinforcement Learning}, rewards are assimilated into model
parameters via gradient-based updates, inducing durable changes to the underlying
policy ($\pi_\theta$). This makes RL well suited for learning
\textbf{generalizable, reusable skills} that transfer across related tasks.

\noindent \textbf{In Test-Time Search}, rewards act as external, transient signals
that guide planning for a single problem instance. Such rewards, often provided by
non-differentiable oracles (e.g., verifiers or execution environments), influence
the current search trajectory without modifying model parameters. As a result,
search enables \textbf{task-specific, on-the-fly optimization} without risking
policy degradation or catastrophic forgetting, to which RL can be vulnerable when
learning sequential tasks.

In summary, RL employs rewards for long-term \textit{policy optimization}, whereas
search uses them for immediate, instance-level \textit{planning and guidance}. A
search reward defines a local objective for a single inference episode, while an
RL reward serves as a global training signal that reshapes model parameters over
many episodes. This distinction is crucial for interpreting hybrid MCTS--training
approaches that leverage test-time search signals for long-term model improvement.

\section{MCTS for LLMs}
\label{sec:main_mcts}

\subsection{Unified Problem Formulation}
To provide a clear comparative framework for MCTS-based LLM reasoning, we adopt a unified notation for consistency across methods.
\textbf{Note:} as in recent LLM planning work, the ``environment’’ is simply the evolving text trace, and transitions are deterministic: each action $a_i$ (a reasoning step) uniquely yields the next state $s_{i+1}$. This is a planning—not stochastic MDP—formulation used in RAP~\citep{hao2023rap}, ReST-MCTS~\citep{zhang2024rest}, AlphaLLM~\citep{tian2024alphallm}, rStar-Math~\citep{guan2025rstar}, and LLaMA-Berry~\citep{zhang2025llamaberry}.

Importantly, states are partial reasoning traces while actions represent only the next incremental step; the two spaces are therefore not equivalent. This asymmetry is intrinsic to deterministic planning and contrasts with RL’s environment-driven MDPs. The objective is to find an optimal reasoning trace $p' = [s_1, \ldots, s_n]$ for a problem $Q$. This formulation enables us to unify insights across papers and surface shared structural principles (e.g., how node granularity interacts with evaluation), which prior works have discussed only in isolation.

\begin{table}[!t]
\centering

\label{tab:notations}
\renewcommand{\arraystretch}{1.2} 
\footnotesize 
\rowcolors{2}{aclgray}{white} 
\begin{tabularx}{\linewidth}{ c >{\raggedright\arraybackslash}X }
    \toprule
    \rowcolor{aclblue} 
    \textbf{Symbol} & \textbf{Definition} \\
    \midrule
    $Q, c$ & Problem question and conditioning prompt \\
    $s_i, a_i$ & Reasoning state and action at step $i$ \\
    $p_i$ & Partial reasoning trace $[s_1, s_2, \ldots, s_i]$ \\
    $v_i, r_{s_i}$ & Value of trace $p_i$ and reward for state $s_i$ \\
    $\pi, V_\theta, R_\theta$ & Policy (LLM), value, and reward models \\
    $T_Q, \mathcal{A}$ & Search tree for problem $Q$ and the action space \\
    $C_i$ & Tree node tuple $(t_i, n_i, q_i)$ (id, count, quality) \\
    \bottomrule
\end{tabularx}
\caption{\textbf{Definition of unified notations.}}
\end{table}

\subsection{Structuring the Search: Node Representation and Granularity}

A fundamental design choice is the definition of a node in the search tree $T_Q$, which dictates the granularity of the search. We identify three primary strategies:

\textbf{Trace-based nodes}, employed in step-driven frameworks like ReST-MCTS* \citep{zhang2024rest}, define each node as a complete partial reasoning trace $p_i = [s_1, \ldots, s_i]$. This representation allows the value function $v_i = V_\theta(p_i)$ to capture the full context of the preceding reasoning path when assessing a node's potential.

\textbf{State-Action nodes}, used in methods such as RAP \citep{hao2023reasoning} and ALPHALLM \citep{tian2024toward}, represent each node as a state-action pair $(s_i, a_i)$. This more localized view focuses evaluation on the immediate quality of a single step, simplifying the input to the reward model.

\textbf{Terminal-State nodes}, a hallmark of purely goal-driven approaches like LLaMA-Berry \citep{zhang2024llama} and MCTSr \citep{zhang2024accessing}, radically restructure the search space. Here, each node represents a complete, terminal solution $s^{\text{terminal}}$. The tree does not model the sequential generation of a single solution but rather a space of candidates, where edges correspond to refinement or rewriting operations. This transforms the problem from finding an optimal path to finding an optimal node.

In practice, these node definitions correspond to different textual granularities: trace-based nodes typically bundle multiple sentences or "reasoning steps", state-action nodes can align with a single reasoning step or short segment, and terminal-state nodes treat entire solutions as atomic. Finer granularity provides more flexible guidance but increases branching and cost, while coarser granularity reduces tree size at the cost of less precise feedback.

\subsection{Designing the Evaluation Function}

The primary differentiator among MCTS-based methods lies in the design of the
evaluation function, which assigns a quality signal to each node—capturing either
the likelihood of reaching a correct solution ($v_i$) or the immediate reward of
a single action ($r_i$). This signal steers the entire search process and reflects
the overarching strategy of the framework.

\subsubsection{Evaluation Locus: Process vs.\ Outcome Rewards}
Evaluation signals can target either the reasoning process or the final outcome.
Methods aimed at improving reasoning trajectories, such as ReST-MCTS*, employ
\textbf{Process Reward Models (PRMs)} or value functions $V_\theta(p_i)$ to score
intermediate, non-terminal states, providing fine-grained guidance that supports
the discovery of high-quality reasoning paths for downstream training.

In contrast, methods focused on final correctness rely on \textbf{Outcome Reward
Models (ORMs)}, where intermediate nodes receive minimal reward and only
terminal states $s^{\text{terminal}}$ are evaluated. Terminal rewards may be
derived from majority voting (rStar \citep{qi2024mutual}), execution-based
verification (PG-TD \citep{zhang2023planning}, RethinkMCTS
\citep{li2024rethinkmcts}), or LLM-based judging (MCTSr, TS-LLM
\citep{feng2023alphazero}). Hybrid approaches such as HiAR-ICL support both PRM and
ORM signals, highlighting the flexibility of this design.

\subsubsection{Evaluator Architecture: External Models vs.\ Self-Evaluation}
Another key choice concerns how rewards are generated. Many systems train
dedicated evaluators, such as specialized value and reward models
($V_\theta, R_\theta$) in ReST-MCTS* and TS-LLM, or pairwise preference models as in
LLaMA-Berry, which ranks candidate solutions.

Alternatively, some methods reuse the policy LLM itself as the evaluator. RAP, for
example, treats the LLM as a world model that predicts both next states and
associated rewards, while MCTSr scores solutions via robust resampling. This
self-evaluation strategy reduces reliance on external data and separate model
maintenance.

\subsubsection{Multi-Critic and Composite Reward Functions}
Several frameworks further combine multiple evaluation signals into a single
composite score. ALPHALLM, for instance, computes node values as a weighted sum of
value estimates, process-level rewards, and outcome-level rewards,
\[
Q_i \leftarrow \beta_1 V_i + \beta_2 R^{\text{PRM}}_i + \beta_3 R^{\text{ORM}}_i .
\]
Related designs follow the same principle: RethinkMCTS augments execution-based
signals with LLM self-evaluation, while LLaMA-Berry combines local pairwise
preferences with a global win--loss signal. Such multi-critic formulations trade
simplicity for robustness by balancing step quality, long-term potential, and
final correctness.

\subsection{Adapting the MCTS Algorithm}
Beyond evaluation design, many methods adapt the core MCTS phases—selection,
expansion, and backpropagation—to better suit LLM reasoning.

In the \textbf{selection} phase, approaches such as PG-TD and rStar augment
standard UCB-style criteria with policy priors, prioritizing nodes that are both
historically promising and likely under the base LLM. 

The \textbf{expansion} phase is often extended with refinement operators.
For example, LLaMA-Berry performs critique-and-rewrite during expansion, while
RethinkMCTS applies verbal feedback to revise failing reasoning steps.

Finally, \textbf{backpropagation} is adapted to the reward structure. While many
methods rely on standard averaging or maximization, others propose smoother update
rules; for instance, MCTSr combines a node’s current value with the best-performing
child to stabilize value propagation. Overall, these adaptations illustrate the
flexibility of MCTS in accommodating the challenges of generative reasoning.

\begin{figure*}[!t]
    \centering
    \includegraphics[width=\linewidth]{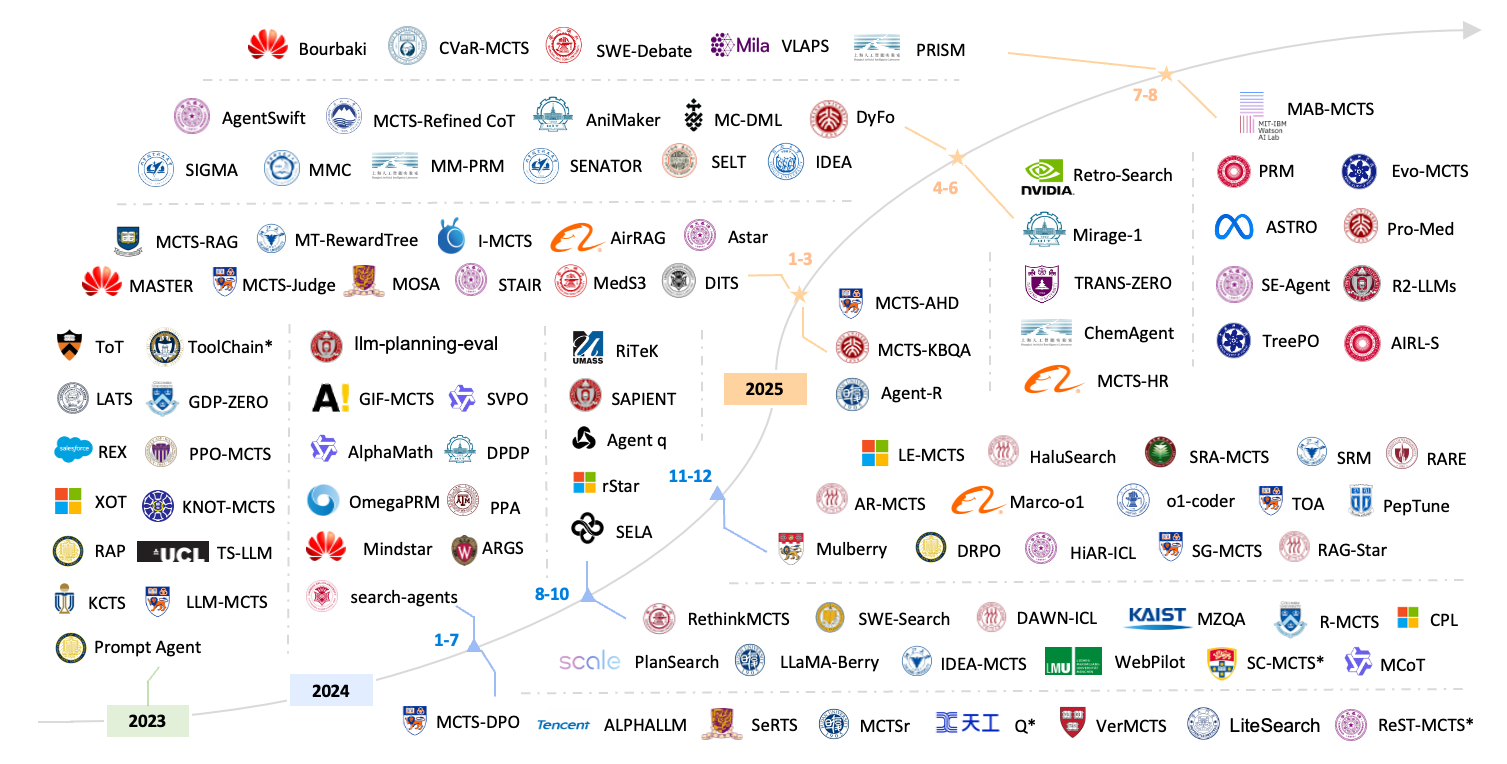}
    \caption{A map of the field's \textbf{rapid growth} on tree search algorithms.}
    \label{fig:roadmap}
\end{figure*}

\subsection{Advanced Topics and Hybrid Approaches}
\label{sec:advanced_topics}

Recent work extends tree-search-based reasoning beyond single-agent settings and
static algorithms. A prominent direction is \textbf{multi-agent and collaborative
search}, where multiple LLM agents coordinate, debate, or specialize to explore the
solution space more effectively, alleviating the limitations of a monolithic agent
in complex tasks such as software engineering and hierarchical planning
\citep{gan2025master, li2025swe, yang2025multi, hou2025halo, zhang2026postergen, Cao2025Multi2MT}.

Another line of progress focuses on improving the \textbf{reward signal} and
\textbf{search efficiency}. To reduce reliance on sparse terminal feedback,
recent methods favor process-supervised reward models (PRMs) that provide step-level
guidance \citep{yu2023ovm, ma2023let}. MCTS itself is increasingly used to
automatically generate such fine-grained supervision, enabling scalable reward
model training without manual annotation
\citep{luo2024improve, ma2025step, jin2025your, brandfonbrener2024vermcts, wu2023knot}.
In parallel, efficiency-oriented advances aim to control search cost through
adaptive dynamics, including information-directed exploration, dynamic node
selection, abstraction control, and test-time architectural adaptation
\citep{chandak2024information, wang2024litesearch, asai2025bilevel,
schmocker2025time, zhang2025booststep, li2025skip}.
Together, these hybrid approaches push tree search toward greater scalability and
practicality \citep{agarwal2025open}.

\begin{table*}[!t]
    \centering
    \small
    \resizebox{\textwidth}{!}{%
    \begin{tabular}{l l l l l l}
    \toprule
    \rowcolor{aclblue}
    \textbf{Task Domain} &
    \textbf{Topology} &
    \textbf{Evaluation} &
    \textbf{Backup} &
    \textbf{Typ. Hyperparams} &
    \textbf{Ref. Methods} \\ 
    \midrule

    \multirow{3}{*}{\makecell[l]{\textbf{Math \& Logic}\\
    \footnotesize (verifiable, long-horizon)}} &
    \textbf{Trace-based} &
    \makecell[l]{\textbf{PRM / PPRM}\\
    \footnotesize (requires high fidelity)} &
    Avg / Sum &
    $c_{puct}\!\in\![1,4]$ &
    ReST-MCTS* \citep{zhang2024restmcts} \\
    & (Step / Solution trees) &
    or Self-Refine &
    (value-driven) &
    Rollouts: $16$--$128$ &
    rStar-Math \citep{guan2025rstar} \\
    & & & & Depth: $8$--$20$ &
    LLaMA-Berry \citep{zhang2025llamaberry} \\
    
    \addlinespace[0.5em]
    \arrayrulecolor{black!15}\hline\arrayrulecolor{black}
    \addlinespace[0.5em]

    \multirow{3}{*}{\makecell[l]{\textbf{Code Generation}\\
    \footnotesize (test-based)}} &
    \textbf{Terminal-state} &
    \makecell[l]{\textbf{ORM (execution)}\\
    \footnotesize (binary, reliable)} &
    Max &
    Rollouts: $16$--$64$ &
    PG-TD \citep{zhang2023planning} \\
    & (Block / Function level) &
    + verbal feedback &
    (success-driven) &
    $k$ samples: $5$--$50$ &
    RethinkMCTS \citep{li2025rethinkmcts} \\
    & & & & Temp: $0.6$--$0.8$ & \\

    \addlinespace[0.5em]
    \arrayrulecolor{black!15}\hline\arrayrulecolor{black}
    \addlinespace[0.5em]

    \multirow{2}{*}{\makecell[l]{\textbf{RAG / Knowledge}\\
    \footnotesize (partial verification)}} &
    \textbf{Hierarchical} &
    \makecell[l]{\textbf{Hybrid (PRM + ORM)}\\
    \footnotesize (fragile if noisy)} &
    Min / AND &
    Retrieval $k$: $3$--$10$ &
    RAG-Star \citep{jiang2025ragstar} \\
    & (Retrieve $\to$ Reason) &
    & (weakest-link) &
    Depth: $3$--$5$ & \\

    \addlinespace[0.5em]
    \arrayrulecolor{black!15}\hline\arrayrulecolor{black}
    \addlinespace[0.5em]

    \multirow{3}{*}{\makecell[l]{\textbf{Autonomous Agents}\\
    \footnotesize (mixed feedback)}} &
    \textbf{State--Action} &
    \makecell[l]{\textbf{Composite}\\
    \footnotesize (cost-sensitive)} &
    Max-of-Avg &
    Depth: task horizon &
    RAP \citep{hao2023rap} \\
    & (World-model tree) &
    (success + shaping) &
    (planning) &
    Rollouts: $20$--$50$ &
    LATS \citep{zhou2023language} \\
    & & & & High $c_{puct}$ & \\
    
    \bottomrule
    \end{tabular}%
    }
    \caption{\textbf{Practitioner’s Guide}: Task-oriented MCTS configurations.
    Parenthetical notes summarize verification reliability and common failure sensitivities,
    clarifying when tree search is most effective and when returns diminish.}
    \label{tab:task_guide}
\end{table*}

\subsection{Applications of MCTS}

Monte Carlo Tree Search (MCTS) has been widely adopted in large language model
(LLM) systems across diverse task domains. At a high level, existing applications
fall into two paradigms: \emph{direct test-time enhancement} and
\emph{self-improvement via data generation}. Together with the algorithmic
variants summarized in Appendix~\ref{appendix:MCTS}, this categorization provides
a practical guide for selecting appropriate MCTS configurations.

In direct test-time enhancement, MCTS is applied during inference to explore
alternative reasoning paths or action sequences, improving output quality without
updating model parameters. This paradigm has been studied in general reasoning
and problem solving \citep{chen2024tree, gao2024interpretable, kang2024mindstar},
mathematical reasoning \citep{zhang2024accessing, xu2023no, yang2024markov}, code
generation and software engineering with compiler- or test-based feedback
\citep{brandfonbrener2024verified, li2024rethinkmcts, wang2025mcts}, agentic
planning in interactive environments \citep{koh2024tree, li2024planning,
hou2025halo}, retrieval-augmented generation and knowledge-intensive tasks
\citep{wu2023knot, jiang2024rag, feng2025airrag, weiretrieval}, and emerging multimodal reasoning
settings \citep{yao2024mulberry, dong2024progressive}. Beyond purely learned or
environmental feedback, a complementary line of work imposes \emph{formal structure}
on the agent's action space: CEDAR restricts LLM agent behaviour with regular-language
specifications and repairs violations in a counter-example driven loop
\citep{chen2025cedar}, while recent work studies the persistent, noise-tolerant active
learning of such regular constraints from class queries \citep{chen2026towards}. Such
symbolic restrictions prune infeasible branches in a sound and inexpensive way, and
therefore complement the learned reward signals discussed above. These requirements
are most stringent in embodied domains, where planning must be coupled with
contact-rich control that can be acquired only from limited real-world interaction
\citep{qiao2026focus, qiao2025signbot}.

In self-improvement via data generation, MCTS is used to generate high-quality
reasoning trajectories that serve as synthetic data for fine-tuning policies or
reward models, drawing inspiration from reinforcement learning and self-play.
Foundational work establishes iterative MCTS-based self-training loops for
improving general reasoning ability \citep{feng2023alphazero, guan2025rstar,
wang2024towards}. This paradigm has since been extended to instruction tuning,
alignment and safety \citep{liu2023don, khanov2024args, zhang2025stair}, scientific
and specialized domains such as medicine and chemistry
\citep{guo2024can, jiang2025meds, pan2025lemma, wei2025ai}, and more recently to multimodal
data generation for vision--language models \citep{wang2025sota, liu2025mmc}.

\subsection{Applicability, Trade-offs, and Practitioner’s Guide}
\label{sec:discussion}

\paragraph{Decision Criteria: When to Use Search.}
Tree search is most effective when the task admits \emph{reliable terminal verification}
or high-fidelity discrimination, enabling exploration of combinatorial solution spaces
without excessive reward noise. This setting is common in mathematics and program
synthesis, where numeric checkers or unit tests provide stable supervision
\citep{qi2024mutual, zhang2024llama}. In such domains, MCTS-based methods routinely
report substantial gains (e.g., 10--40\%) over greedy decoding. In contrast, for
open-ended generation tasks without verifiable correctness, search typically yields
marginal improvements over strong decoding baselines.

\paragraph{When Search Is Unwarranted.}
Empirical evidence suggests several recurring regimes where search offers poor
returns or degrades performance.
(i) When reward models or discriminators are weakly correlated with final correctness,
search over-exploits spurious signals and can exhibit inverse scaling
\citep{chen2024tree_search_discriminator, gao2023scaling}.
(ii) For short-horizon or easy instances, large inference-time budgets lead to
``overthinking'' with little benefit \citep{chen2024overthinking}.
(iii) When evaluation or verification dominates the compute budget, exploration
depth is severely constrained, yielding unfavorable accuracy--latency trade-offs
\citep{brown2024large}.
In these settings, lightweight alternatives such as re-ranking or small-$n$
self-consistency are often preferable.

\paragraph{Configuration Trade-offs When Search Applies.}
When search is appropriate, performance is governed by two primary trade-offs.
First, \emph{backup strategy}: max backups suit binary-verifier tasks (e.g., code),
while average backups stabilize high-variance domains such as mathematics
\citep{zhang2024restmcts, li2025rethinkmcts}.
Second, \emph{evaluation cost}: high-fidelity PRMs reduce reward noise but limit
search depth, whereas lightweight self-evaluation enables broader exploration.
Across studies, allocating roughly 20--30\% of inference compute to evaluation
yields robust gains.

\paragraph{MCTS vs.\ Heuristic Search.}
Heuristic methods such as Tree-of-Thoughts rely on LLM-generated intermediate
heuristics and excel under tight latency budgets or well-calibrated signals.
MCTS instead accumulates experience-driven statistics, making it more robust
in sparse-reward or deceptive-intermediate regimes common in long-horizon
reasoning tasks \citep{hao2023rap}.

\section{Informed Search with LLM-Generated Heuristics}
\label{sec:main_Informed}

Informed search guides Large Language Model (LLM) reasoning by leveraging
heuristics to navigate large solution spaces. Unlike classical approaches with
manually designed heuristics, modern methods generate guidance dynamically using
the LLM itself or auxiliary signals. Existing approaches largely fall into two
paradigms: \emph{direct state evaluation} and \emph{composite A* cost functions}.
Additional details are provided in Appendix~\ref{appendix:Informed}.

Direct state evaluation, exemplified by Tree-of-Thoughts (ToT)
\citep{yao2023tree}, treats the LLM as an on-the-fly heuristic.
The model first proposes multiple candidate next steps (“thoughts”), which are
then scored by an LLM-based evaluator. These scores guide classical search
procedures, such as beam search that retains the top-$b$ states or pruned DFS
that discards low-scoring branches.

A complementary paradigm adapts A* search by constructing composite heuristics
for the cost function $f(n)=g(n)+h(n)$. Here, $g(n)$ captures progress along the
current reasoning path, while $h(n)$ estimates the remaining cost to the goal.
Methods such as ToolChain* \citep{zhuang2024toolchain} and Q*
\citep{wang2024qimprovingmultistepreasoning} derive $g(n)$ and $h(n)$ from multiple
LLM-relevant signals, enabling more informed prioritization of partial solutions.
The key heuristic components are summarized in Table~\ref{tab:astar_heuristics}.

\begin{table}[!t]
\centering
\renewcommand{\arraystretch}{1.15} 
\footnotesize
\rowcolors{2}{aclgray}{white}
\begin{tabularx}{\columnwidth}{l c X}
\toprule
\rowcolor{aclblue}
\textbf{Heuristic} & \textbf{Comp.} & \textbf{Mechanism / Signal} \\
\midrule
Process Rewards & $g(n)$ & Aggregates step-level feedback (e.g., execution results and logits). \\
Stat. Consistency & $g(n)$ & Prioritizes steps frequently sampled across generations. \\
Memory Comp. & $g, h$ & Measures similarity to high-quality examples (e.g., LCS). \\
Learned Value & $h(n)$ & Predicts cost-to-goal via a trained proxy (e.g., Q-function). \\
\bottomrule
\end{tabularx}
    \caption{\textbf{A* Heuristic Components}.}
    \label{tab:astar_heuristics}
\vspace{-0.5em}
\end{table}

\section{Evaluation Framework}
\label{sec:main_evaluation_framework}

Recent progress in tree-structured decoding highlights test-time compute as an
important axis for scaling reasoning performance. However, cross-paper
comparisons remain difficult, as reported gains often conflate policy size,
evaluation cost, verification overhead, and hardware assumptions (see
Appendix~\ref{appendix:evaluation_compute}).

Rather than introducing a new metric or benchmark, we propose a lightweight
\emph{reporting abstraction}, termed the \emph{Standardized Compute-Reporting
Protocol} (SCRP), to make inference-time compute expenditures explicit and
comparable. SCRP is descriptive rather than prescriptive: it does not rank methods
or define optimality, but standardizes what quantities are reported.

SCRP decomposes inference-time resources into a vector
$\mathbf{B} = (C_{\text{policy}}, C_{\text{eval}}, C_{\text{verify}}, T_{\text{wall}})$.
For hardware-agnostic comparison, we adopt a first-order approximation of
per-instance cost:
\begin{equation}
\begin{aligned}
\mathcal{C}_{\text{total}}(x) \approx\;&
2P_{\text{policy}}T_{\text{policy}}(x) \\
&+ 2P_{\text{eval}}T_{\text{eval}}(x) \\
&+ C_{\text{verify}}(x),
\end{aligned}
\end{equation}
which serves as a monotonic proxy rather than a systems-level model.

Based on this decomposition, we encourage reporting performance as a function of
compute budget (e.g., Pass@FLOPs or Tokens-per-Solved), \emph{in addition to}
task-specific metrics such as accuracy, to surface trade-offs otherwise obscured
by raw performance numbers.

\section{Challenges, Future and Conclusion}
\label{sec:main_challenges}

Despite clear gains in reasoning, tree-search methods face two major bottlenecks:
\textbf{compute} and \textbf{reward quality}. Compared to greedy decoding, search
introduces substantial overhead~\citep{wang2024litesearch}, which is exacerbated by
strong models that often \emph{overthink} simple queries~\citep{chen2024not,
zeng2024scaling, wei2026think}. Structural constraints further limit parallelism and slow the
self-play cycles used to distill search behavior into base models
\citep{xiang2025towards}. A complementary lever is to shorten the reasoning trace
itself rather than only pruning the tree: DRAFT-RL couples concise chain-of-draft
reasoning with multi-agent coordination and reinforcement learning, preserving much
of the accuracy of verbose reasoning at a fraction of the token budget
\citep{li2026draftrl}. Addressing these issues will require more adaptive,
selectively activated search procedures with dynamic resource allocation and
aggressive pruning.

A second fundamental challenge lies in constructing reliable reward models. While
process reward models (PRMs) provide finer-grained supervision than outcome reward
models (ORMs), they rely on costly and hard-to-scale annotations
\citep{uesato2022solving, lightman2023let}, and current automated approaches remain
confined to narrow domains such as mathematics~\citep{wang2024math,
luo2024improve}. Imperfect rewards can misguide search and even induce
\emph{inverse inference scaling}, where additional rollouts degrade accuracy
\citep{gao2023scaling, zeng2024scalingsearchlearningroadmap}. The persistent gap
between learned PRMs and oracle verifiers
\citep{xiang2025towards} highlights the need for scalable
methods to generate high-fidelity process rewards. A promising alternative is to
replace or augment learned rewards with symbolic, sound-by-construction constraints,
such as regular-language restrictions on agent behaviour \citep{chen2025cedar}, and
to acquire such specifications automatically from noisy interaction data
\citep{chen2026towards}.

Overall, this survey unified classical and MCTS-style approaches around node
representation, reward design, and algorithmic adaptation for LLMs. Future
progress will depend on lighter-weight search dynamics and scalable, high-quality
reward signals to establish tree search as a general-purpose reasoning mechanism.

\section*{Limitation}
This study focuses on presenting a coherent framework and empirical analysis under a fixed set of experimental assumptions, rather than exhaustively exploring all possible model variants, hyperparameter configurations, or alternative implementation choices. While different design decisions—such as search depth, evaluation signals, or compute allocation strategies—may lead to variations in quantitative performance, these factors are not expected to alter the central observations or conclusions of this work. In addition, experiments are conducted on commonly used benchmarks and controlled settings, which may not fully capture the diversity, noise, and constraints encountered in real-world applications. Extending the evaluation to broader tasks, larger model families, and more heterogeneous environments is left for future work.

\section*{Acknowledgments}
Supported by Shanghai Artificial Intelligence Laboratory.

\bibliography{anthology}

\onecolumn

\appendix
\section*{Appendix}

\setcounter{tocdepth}{2}
\tableofcontents
\newpage

\section{Organization of the Appendix}
\label{app:roadmap}

The appendix is organized to provide a structured and progressively layered extension of the main paper, moving from foundational concepts to methodological taxonomies, and finally to evaluation standards and open challenges. This organization is designed to support both \emph{conceptual clarity} and \emph{practical usability}, enabling readers to navigate the rapidly growing landscape of inference-time tree search through a coherent and unified framework. To mitigate the fragmentation of prior literature, the appendix emphasizes visual taxonomies, comparative tables, and standardized abstractions that facilitate cross-method comparison and selective reading. The supplementary material is divided into six interrelated modules.

\paragraph{Foundational Paradigms (Appendix~\ref{appendix:Paradigms}).}
We begin by revisiting three foundational search paradigms that underlie modern LLM-based reasoning methods: uninformed search (e.g., BFS and DFS), informed search (heuristic-guided methods such as $A^{*}$), and Monte Carlo Tree Search (MCTS). This appendix establishes the algorithmic primitives, representational assumptions, and computational trade-offs shared across these paradigms, providing a common vocabulary and historical grounding for the adaptations introduced in later sections.

\paragraph{Theoretical Distinctions (Appendices~\ref{appendix:test_time_scaling} and~\ref{appendix:Reward}).}
These appendices formalize key conceptual axes that motivate inference-time search.
\begin{itemize}
    \item \textbf{Appendix~\ref{appendix:test_time_scaling} (Test-Time Optimization):}
    This section reframes tree search as a computation-centric alternative to parameter-centric training. We introduce a task-defined objective space that decomposes inference into a \emph{Prompt Space} (algorithm and policy selection) and an \emph{Answer Space} (solution generation). This abstraction clarifies how MCTS and related methods operate as structured test-time optimization procedures rather than training-time learning algorithms.

    \item \textbf{Appendix~\ref{appendix:Reward} (Reward as Guidance vs.\ Learning Signal):}
    Here we disentangle the overloaded notion of ``reward'' by contrasting its role in Reinforcement Learning with its role in deliberative search. We show that, in inference-time search, rewards act as transient, instance-specific guidance signals rather than persistent learning objectives, providing a conceptual foundation for understanding process rewards, outcome rewards, and hybrid designs without conflating search with policy optimization.
\end{itemize}

\paragraph{Methodological Taxonomy (Appendices~\ref{appendix:MCTS} and~\ref{appendix:Informed}).}
These modules constitute the methodological core of the appendix, mapping the design space of tree-search-based reasoning.
\begin{itemize}
    \item \textbf{Appendix~\ref{appendix:MCTS} (Monte Carlo Tree Search):}
    This appendix provides a comprehensive and hierarchical treatment of MCTS for LLMs. We first present a unified notation and visual taxonomy that systematizes node representations, evaluation loci, and backup strategies across prior work. Building on this foundation, we offer a \emph{practitioner-oriented guide} that distills empirically effective configurations across task domains into comparative tables. We further organize advanced topics and applications into two functional paradigms: \emph{direct test-time enhancement}, where MCTS improves inference without parameter updates, and \emph{self-improvement}, where search-generated trajectories are used for data synthesis and model refinement.

    \item \textbf{Appendix~\ref{appendix:Informed} (Heuristic-Guided Search):}
    This appendix analyzes informed search methods, including LLM-augmented BFS/DFS and $A^{*}$-style approaches. We focus on heuristic construction, cost decomposition, and admissibility–efficiency trade-offs, positioning heuristic-guided search as a complementary alternative to MCTS in settings with reliable intermediate guidance or strict latency constraints.
\end{itemize}

\paragraph{Standardized Evaluation Protocols (Appendix~\ref{appendix:evaluation_compute}).}
To enable reproducible and hardware-agnostic comparison across studies, this appendix introduces a unified protocol for reporting test-time compute. We provide practical recipes for FLOP estimation and wall-clock profiling, and formalize evaluation metrics such as Budgeted Accuracy and Tokens-per-Solved. These standards address inconsistencies in prior reporting and establish a principled basis for benchmarking inference-time search methods.

\paragraph{Practitioner’s Guide with Unified Notation (Appendix~\ref{appendix:Walkthrough}).}
To bridge theoretical abstraction and practical deployment, this appendix provides a consolidated practitioner’s guide grounded in the unified notation introduced throughout the survey. Rather than presenting methods as isolated algorithms, we re-express representative approaches using a shared set of symbols for states, actions, node definitions, evaluation functions, and backup rules. This normalization enables direct, side-by-side comparison of design choices and highlights common structural patterns that are obscured by paper-specific notation. The guide further distills recurring configurations into task-oriented templates, offering concrete recommendations for node granularity, reward design, and search dynamics across domains such as mathematics, code generation, retrieval-augmented reasoning, and agentic planning.

\paragraph{Challenges and Future Directions (Appendix~\ref{appendix:Challenges}).}
We conclude by synthesizing open challenges revealed throughout the survey, including overthinking on simple tasks, efficiency bottlenecks in deep or wide search, and the heavy reliance on high-quality reward models. These limitations motivate future research directions at the intersection of adaptive search dynamics, scalable reward modeling, and selective computation.

\noindent
Taken together, this organization enables readers to progress from conceptual foundations and theoretical distinctions to algorithmic taxonomies, standardized evaluation, and finally practitioner-oriented synthesis, supporting both principled understanding and informed adoption without requiring traversal of long, sequential method listings.

\section{Foundational Search Paradigms in General AI}
\label{appendix:Paradigms}

Solving complex problems can be formalized as a search task: finding an optimal path from an initial state to a goal state within a state-action space, conventionally represented as a tree $T_Q$. While classical AI has developed a rich toolkit for navigating such trees, the state spaces implicit in language model reasoning present unique challenges. They are not merely large; they are combinatorially vast, high-dimensional, and semantically structured, rendering exhaustive exploration computationally infeasible. This section revisits three foundational paradigms of tree search---uninformed, informed, and Monte Carlo-based---to establish a conceptual vocabulary for understanding their modern adaptations for LLM-based reasoning, where the goal is to identify optimal reasoning paths efficiently.

\subsection{Uninformed Search: Blind Exploration}
Traditional search algorithms, such as Breadth-First Search (\citet{moore1959shortest}, BFS), Depth-First Search (DFS), and Uniform Cost Search (UCS, or Dijkstra's algorithm), are \textbf{uninformed search} algorithms that operate with minimal knowledge about the goal. These algorithms can recognize the goal state when reached but lack any additional information to guide them toward it efficiently \citep{DBLP:books/aw/RN2020}. While some uninformed search algorithms, like UCS, consider the cost of the path taken so far, none can estimate the remaining distance to the goal or determine which paths are more promising. 

The key characteristic of uninformed search is that it must rely solely on the problem's basic definition - the available actions, their costs, and the goal recognition criteria - to systematically explore the search space. As a result, these algorithms differentiate between possible solution paths primarily through their order of exploration and accumulated costs. Each algorithm offers different guarantees: BFS finds the shortest path in terms of steps, while UCS finds the lowest-cost path. Additional variants like Depth-Limited Search (DLS) and Iterative Deepening Search (IDS) address memory limitations of basic DFS while maintaining completeness. The choice between these algorithms often depends on the problem's characteristics and computational constraints, particularly memory requirements.

\begin{table*}[!t]

\centering
\small 
\setlength{\tabcolsep}{4pt} 
\renewcommand{\arraystretch}{1.1}
\begin{tabularx}{\linewidth}{l l >{\raggedright\arraybackslash}X l}
\toprule
\rowcolor{mycustomblue}
\textcolor{white}{\textbf{Family}} & \textcolor{white}{\textbf{Algorithm}} & \textcolor{white}{\textbf{Guiding Signal / Principle}} & \textcolor{white}{\textbf{Typical Use Case}} \\
\midrule
\multirow{4}{*}{Uninformed}
& BFS & Explores layer-by-layer; guarantees shortest path in steps. & Shortest path, unweighted graphs. \\
& DFS & Explores a single branch to its depth before backtracking. & Path existence, memory efficiency. \\
& UCS & Expands node with the lowest accumulated path cost $g(n)$. & Optimal path, weighted graphs. \\
& IDS & Depth-first search with an incrementally increasing depth limit. & Optimal path, low memory overhead. \\
\midrule
\multirow{5}{*}{Informed}
& Greedy BeFS & Expands node closest to goal via heuristic $h(n)$ alone. & Quick, non-optimal solutions. \\
& A* Search & Balances path cost $g(n)$ \& heuristic $h(n)$. & General-purpose optimal planning. \\
& Weighted A* & Biases toward heuristic via $g(n)+w \cdot h(n), w>1$. & Speed-optimality trade-offs. \\
& IDA* & Iterative deepening applied to the A* cost function $f(n)$. & Memory-efficient optimal search. \\
& Beam Search & Keeps top-$k$ most promising candidates at each step. & High branching factor problems. \\
\midrule
\multirow{3}{*}{\shortstack[l]{Monte Carlo \\ (Sampling)}}
& UCT-MCTS & UCT balances exploitation ($q_j$) \& exploration. & Games/planning in vast state spaces. \\
& LLM-MCTS & LLM acts as policy prior $\pi$ and/or rollout policy. & Test-time deliberative reasoning. \\
& PUCT Variants & Integrates a policy network's prior $\pi$ into UCT bonus. & Integrating learned priors into search. \\
\bottomrule
\end{tabularx}
\caption{A comparative taxonomy of foundational search algorithms in AI. Notation: $g(n)$ is the accumulated path cost to node $n$; $h(n)$ is the heuristic estimate of the cost from $n$ to the goal; $q_i$ is the estimated quality value of a search tree node $C_i$.}
\label{tab:search_general_ai_compact}
\end{table*}

\subsection{Informed Search: Heuristic-Guided Exploration}
\textbf{Informed search}, or \textbf{heuristic search}, leverages domain-specific knowledge to guide the exploration toward the goal \citep{DBLP:books/aw/RN2020}. This knowledge is encoded in a \textbf{heuristic function} $h(n)$, which estimates the minimum cost from a node $n$ to a target state. Let $h^*(n)$ denote the \textit{true} optimal cost-to-go; the heuristic is formally defined as an estimator:
\begin{equation}
\label{eq:heuristicdef}
h(n) \approx h^*(n).
\end{equation}
By incorporating $h(n)$, algorithms can prioritize promising paths to reduce computational cost. The theoretical guarantees of these algorithms depend on the properties of $h(n)$. Let $c(n, n')$ be the cost of the edge between $n$ and its successor $n'$. A heuristic is considered:
\begin{itemize}
    \item \textbf{Admissible} if it never overestimates the true cost, i.e., $0 \le h(n) \le h^*(n)$ for all $n$.
    \item \textbf{Consistent} (or monotone) if it satisfies the triangle inequality: $h(n) \le c(n, n') + h(n')$.
\end{itemize}
The choice of $h(n)$ directly impacts efficiency. A heuristic $h_1$ is said to be \textit{more informed} (or dominant) than $h_2$ if $h_1(n) \ge h_2(n)$ for all $n$ (assuming both are admissible). Dominant heuristics generally prune the search space more effectively by providing tighter bounds on $h^*(n)$.

However, there is often a trade-off between the computational cost of calculating the heuristic and the savings it provides in search efficiency.
Common informed search algorithms include Greedy Best-First Search (BeFS), A* Search, Weighted A* Search, Iterative Deepening A* (IDA*), Beam Search, and Recursive Best-First Search (RBFS) . These algorithms vary in how they balance the heuristic estimates with path costs, leading to different trade-offs between optimality and efficiency. For instance, A* search, when used with an admissible heuristic, guarantees finding an optimal solution if one exists. The success of these algorithms in practical applications often depends on designing effective problem-specific heuristics. Common techniques for developing heuristics include relaxing problem constraints, pattern databases, and learning from experience \citep{DBLP:books/aw/RN2020}. While informed search algorithms generally outperform uninformed search in practice, their effectiveness relies heavily on the quality of their heuristic functions and the specific characteristics of the problem domain.

\subsection{Monte Carlo Tree Search: Learning from Experience}
Monte Carlo Tree Search (MCTS) was first introduced by \citet{Coulom2006EfficientSA} in the context of computer Go as an \textbf{adversarial search} algorithm, which aims to maximize winning probability against an optimal opponent. While adversarial MCTS alternates between players and models opponent responses, the MCTS variant used in LLM's inference-time search is a \textit{single-agent} formulation, where the algorithm explores different action sequences without modeling opposing players. This adaptation maintains MCTS's core strengths in balancing exploration and exploitation through statistical sampling, while refocusing the objective from competitive game-playing to finding optimal sequences of actions in a non-adversarial environment.

Inference-time MCTS (hereafter referred to simply as MCTS) retains the four fundamental phases of the original algorithm: selection, expansion, simulation, and backpropagation. During selection, the algorithm traverses the tree using the \textit{Upper Confidence bounds applied to Trees (UCT) policy}, which balances exploration and exploitation by selecting nodes (states) that maximize:
\begin{equation}
\label{eq:uct}
a^* = \arg\max_{a\in A(s)}\left[Q_i + c\sqrt{\frac{\ln n_i}{N(s,a)}}\right]
\end{equation}
where $Q(s,a)$ estimates the expected future reward of taking action $a$ in node $s$, $N(s)$ is the number of times node $s$ has been visited, $N(s,a)$ is the number of times action $a$ has been selected in node $s$, $c$ is an exploration constant, and $A(s)$ is the set of available actions at node $s$ \citep{Kocsis2006BanditBM}. In the expansion phase, new nodes sampled by LLMs (e.g. subsequent steps in reasoning) are added to the tree to gradually build a model of the search space. The simulation phase performs rollouts from leaf nodes using a default policy to estimate long-term rewards, replacing the win/loss outcomes of adversarial MCTS with domain-specific reward measures.

Unlike traditional uninformed search algorithms such as BFS or DFS that systematically explore the state space, MCTS offers a statistical sampling approach that can handle much larger search spaces. Compared to informed search algorithms like A*, which rely on pre-defined heuristics, MCTS builds its evaluation function through experience. This makes it particularly suitable for LLM inference where defining accurate heuristics is challenging. The algorithm's ability to balance between exploration and exploitation, combined with its flexibility in handling large state spaces, makes it a powerful tool for guiding LLM inference, though its effectiveness depends on carefully managing the trade-offs between computational resources and search depth.

\subsection{Comparison of Exploration Strategies}
Figure \ref{fig:search} provides a conceptual illustration of these distinct exploration strategies. Uninformed algorithms like BFS and DFS are governed by rigid, topology-driven expansion protocols. Informed search, exemplified by A*, introduces goal-directedness by prioritizing search based on a heuristic cost-to-go estimate, $h(\cdot)$, allowing it to focus on promising regions irrespective of tree topology. Finally, MCTS replaces the static heuristic with a dynamically learned value function, estimated via statistical sampling. This adaptive, self-correcting mechanism allows it to focus computational resources on the most promising regions of the search space without requiring prior domain knowledge encoded in a heuristic. This very property makes it the preeminent search paradigm for navigating the vast and ill-defined reasoning spaces of large language models.

\section{Test-time Scaling via Search}
\label{appendix:test_time_scaling}

As the scaling of model parameters and training data yields diminishing returns, a new frontier has emerged: \textbf{test-time scaling}. This paradigm investigates how to optimally allocate computational resources during inference to enhance a model's effective reasoning capabilities. Unlike training-time scaling, which refines a global, amortized policy by encoding knowledge into a model's weights, test-time scaling performs instance-specific optimization for a given problem $Q$. This section provides a detailed, mathematically-grounded analysis of these two orthogonal paradigms, contrasting how they operate in fundamentally different optimization landscapes: the latent parameter space for training versus the task-defined objective space for inference.

\subsection{A Tale of Two Optimizations for LLM Scaling: Training-Time vs. Test-Time}
The figure referenced illustrates two distinct approaches for improving model performance, each defined by its unique objective signal and the space over which it optimizes.

\paragraph{Training-Time Scaling: Optimization in Latent Parameter Space.}
During training, the primary goal is to learn a set of parameters $\theta^*$ that minimizes an expected loss function $\mathcal{L}$ over a data distribution $\mathcal{D}$. The optimization problem is formally stated as:
$$
\theta^* = \arg\min_{\theta \in \Theta} \mathbb{E}_{(i,o) \sim \mathcal{D}}[\mathcal{L}(f_\theta(i), o)],
$$
where $\Theta \subseteq \mathbb{R}^N$ is the high-dimensional \textbf{latent parameter space}. The \textbf{objective signal} in this paradigm is the gradient of the loss with respect to the parameters, $\nabla_\theta \mathcal{L}$. Optimization proceeds via iterative updates, such as stochastic gradient descent. The result is a static artifact---a trained model $\pi$---that implicitly represents a posterior distribution over solutions.

\paragraph{Test-Time Scaling: Optimization in Task-Defined Objective Space.}
Given a fixed, pretrained model $\pi$, test-time scaling seeks to find an optimal reasoning trace $p^*$ for a specific problem instance $Q$. This process constitutes a second, distinct optimization loop. The search occurs in a discrete, structured \textbf{task-defined objective space}, the solution space $\mathcal{P}(Q)$, which consists of all possible reasoning traces. The \textbf{objective signal} is a scalar \textbf{reward} or \textbf{value} that evaluates the quality of a trace. The optimization problem at inference is therefore:
$$
p^* = \arg\max_{p \in \mathcal{A}(\pi, Q, \mathcal{C}_{\text{infer}})} V(p),
$$
where $\mathcal{A}(\pi, Q, \mathcal{C}_{\text{infer}})$ is the search algorithm that explores a subset of $\mathcal{P}(Q)$ guided by the model's prior $\pi$ and constrained by the inference compute budget $\mathcal{C}_{\text{infer}}$, and $V(p)$ is a function evaluating the final trace. Scalable inference techniques, such as tree search, use intermediate rewards $r_s$ or partial trace values $v_i$ to dynamically allocate compute to more promising regions.

\subsection{Operationalizing Search in the Objective Space}
The conceptual shift from gradients in latent space to rewards in objective space necessitates a different class of optimization algorithms. While training relies on gradient-based methods, test-time scaling is operationalized by search procedures that can navigate complex, non-differentiable solution spaces.

\paragraph{Tree Search as a Scalable Inference Optimizer.}
Tree search methods, particularly MCTS, provide a principled framework for this optimization. They build a search tree $T_Q$ where each node $C_i$ corresponds to a partial reasoning trace $p_i$. At each node, an action selection policy balances exploiting known high-reward paths and exploring novel ones. For LLM-based search, this policy often uses a PUCT-style rule that incorporates the policy network's prior. The next action $a^*$ is selected by choosing the action that leads to the most promising child node:
$$
a^* = \underset{a \in \mathcal{A}(s_i)}{\arg\max} \left( q_j + U(C_i, C_j) \right),
$$
where $s_i$ is the state at the parent node $C_i$, and action $a$ leads to the child node $C_j$ with quality value $q_j$. The uncertainty bonus $U(C_i, C_j)$ is formulated as:
$$
U(C_i, C_j) = c_{\text{exp}} \cdot \pi(a|p_i, Q) \cdot \frac{\sqrt{n_i}}{1 + n_j}.
$$
Here, $n_i$ and $n_j$ are the visit counts of the parent and child nodes, respectively. The policy $\pi$ provides a prior probability for taking action $a$ given the history $p_i$, and $c_{\text{exp}}$ is an exploration hyperparameter. This synthesis allows the algorithm to scale reasoning performance effectively with the allocated inference compute budget.

\subsection{Decomposing the Objective Space: Prompt and Answer Spaces}

The task-defined objective space, over which test-time search operates, is not monolithic. It can be productively decomposed into two distinct, hierarchically-related search spaces: the \textbf{Prompt Space} and the \textbf{Answer Space}. This decomposition clarifies the mechanisms of Chain-of-Thought (CoT) reasoning and reveals the limitations of many current test-time search methods. The overall optimization problem is thus a search for an optimal reasoning trace, which involves finding both the right algorithm and its correct execution.

\paragraph{The Prompt Space ($\mathcal{P}$): Searching for an Algorithm.}
The prompt space, $\mathcal{P}$, encompasses the set of all possible reasoning structures or ``step templates'' an LLM can adopt to solve a problem. Each template $p \in \mathcal{P}$ represents a specific strategy for externalizing and manipulating information from the model's latent state $\mathbf{h}$ into its textual output space \citep{zhang2025prompt}. In essence, selecting a template $p$ is equivalent to selecting an \textbf{algorithm}. For example, one template for a complex arithmetic task might involve explicitly tracking a running total, while another might only verbalize intermediate calculations without a canonical state representation.

The choice of template is paramount because it dictates the computational graph the model simulates through its autoregressive generation. While theoretical work suggests that a CoT-augmented Transformer can be Turing-complete \citep{li2024chain}, this potential is contingent on generating the correct computational trace; indeed, theoretical analysis shows that the complexity of the prompt space and its interaction with the answer space critically determine whether a CoT prompt works \citep{zhang2025does}. An suboptimal template can lead to an inefficient or even intractable search by failing to surface the necessary state information for subsequent steps, effectively breaking the simulated recurrence. The search for an optimal $p^* \in \mathcal{P}$ is therefore a meta-level optimization: discovering the most effective procedure for solving the task instance.

\paragraph{The Answer Space ($\mathcal{S}$): Searching for a Solution.}
For any given prompt template $p$, there exists a corresponding answer space, $\mathcal{S}_p$, which contains all possible reasoning traces (i.e., potential solutions) that can be generated by adhering to that template's structure. The complexity of navigating this space is critically conditioned on the choice of $p$. An effective template $p^*$ dramatically prunes the answer space, simplifying the path to a correct solution. Conversely, a poorly chosen template $p'$ can render the answer space vast and unstructured, making the search computationally infeasible even with a large compute budget.

Many contemporary test-time search methods, such as Tree-of-Thought \citep{yao2023tree} and Graph-of-Thought \citep{besta2024got}, operate primarily within this second level of the hierarchy. They typically fix a single, heuristically-defined prompt template (e.g., via a generic instruction like ``think step by step'') and then deploy sophisticated search algorithms to navigate the resulting answer space $\mathcal{S}_p$. These approaches excel at mitigating execution errors and exploring diverse solution paths \textit{within a fixed algorithmic strategy}. However, they do not address the foundational challenge of selecting the algorithm itself. If the governing template $p$ is flawed, even an exhaustive search of $\mathcal{S}_p$ is unlikely to yield a correct solution.

\paragraph{A Unified View of Test-Time Search.}
A comprehensive framework for test-time search must therefore account for the joint optimization over both spaces. The ultimate objective is to discover a solution trace $s^*$ that maximizes the value function $V(\cdot)$, where the search spans all possible traces allowed by all possible templates:
$$
s^* = \arg\max_{p \in \mathcal{P}, s \in \mathcal{S}_p} V(s)
$$
This formulation highlights a critical gap in current research. While significant effort has been invested in optimizing search algorithms within a given answer space $\mathcal{S}_p$, the systematic exploration of the prompt space $\mathcal{P}$ remains a largely open challenge. The true potential of test-time scaling lies not merely in executing a known algorithm more robustly, but in dynamically discovering the most effective algorithm for the specific problem at hand.

\section{Reward as a Unified Signal for RL and Search: One Objective, Two Optimizers}
\label{appendix:Reward}

In advanced AI systems, a reward signal is the fundamental currency for guiding behavior. However, its role bifurcates into two distinct yet complementary functions depending on the temporal scope of the objective: shaping a durable, long-term \textbf{policy} versus guiding a transient, short-term \textbf{plan}. This distinction is not one of paradigm but of application—whether the reward is used to permanently update the model's internal parameters (RL learning) or to direct a temporary search with fixed parameters (planning).

\subsection{RL via Policy Shaping: Internalizing Rewards for Generalization}
In Reinforcement Learning (RL), the reward signal is \textbf{internalized} directly into the model parameters, fundamentally altering the agent's behavior. Unlike transient inference-time guidance, RL distills experience into a robust policy, effectively encoding ``instincts'' that govern future actions. Formally, this process optimizes the policy parameters $\theta$ to maximize an objective $\mathcal{J}_{\text{RL}}$, which balances task-specific rewards with adherence to a set of foundational principles encapsulated by a prior policy $\pi_{\mathcal{P}}$.

The optimization objective seeks the optimal parameters $\theta^*$ that maximize the expected return while minimizing the deviation from the prior:
\begin{equation}
\label{eq:rl_objective}
\begin{split}
\theta^* = \mathop{\arg\max}_{\theta} \mathbb{E}_{\tau} \Big[ &\sum\nolimits_{t} \gamma^t R(s_t, a_t) \\[-1ex] 
&- \beta \sum\nolimits_{t} D_{\text{KL}}\big(\pi_\theta(\cdot|s_t) \,\|\, \pi_{\mathcal{P}}\big) \Big]
\end{split}
\end{equation}
where $\tau$ represents a trajectory of length $T$, and $D_{\text{KL}}$ is the Kullback-Leibler divergence serving as a regularization term. The coefficient $\beta$ controls the strength of this constraint, preventing the model from ``reward hacking'' by diverging too far from the foundational principles (e.g., fluency or safety). In this framework, the reward function acts as a long-term teacher, shaping the agent's generalizable capabilities for unseen challenges.

\subsection{Search via Deliberative Planning: Externalizing Rewards for Specificity}
Conversely, during test-time search, the reward signal functions as an \textbf{external, ephemeral guide}. It directs a deliberative process, like Monte Carlo Tree Search (MCTS), to navigate the solution space for a single, immediate task. The reward evaluates candidate action sequences (plans), allowing the system to identify a high-quality solution for the specific problem at hand. For a given task with a specific external reward function $R_{\text{ext}}$, the goal is to find an optimal plan $p^*$ that maximizes a combination of this external signal and an internal, path-dependent heuristic $\mathcal{H}_{\theta}$ provided by the frozen model.

The optimal plan $p^*$ for a state sequence $s_0, s_1, \dots, s_T$ resulting from the plan's actions is found by solving:
\begin{equation}
\begin{aligned}
p^* = \mathop{\arg\max}_{p \in \mathcal{P}_{\text{plan}}} \bigg[ 
&\sum_{t=0}^{T-1} \gamma^t R_{\text{ext}}(s_t, a_t) \\
&+ \mathcal{H}_{\theta}(s_T, p) \bigg]
\end{aligned}
\end{equation}
where the heuristic $\mathcal{H}_{\theta}$ is not just a simple state evaluation but a complex function of the final state $s_T$ and the path $p$ taken, potentially incorporating penalties for path irregularity or deviation from the model's learned priors:
\begin{equation}
\begin{aligned}
\mathcal{H}_{\theta}(s_T, p) = \, &V_\theta(s_T) \\
&- \beta \cdot \log \left( \int_{\tilde{p} \in \mathcal{N}(p)} e^{-\mathcal{E}(\tilde{p})/\tau_c} \, d\tilde{p} \right)
\end{aligned}
\end{equation}
Here, $V_\theta(s_T)$ is the model's intrinsic value estimate, while the second term acts as a complexity penalty based on the ``free energy'' over a neighborhood of paths $\mathcal{N}(p)$, discouraging overly surprising or convoluted solutions. Crucially, this feedback is discarded once the task is complete; the model's underlying parameters $\theta$ remain untouched. This makes the reward an ideal tool for \textbf{task-specific, localized objectives} without the risk of corrupting the model's general-purpose policy.

\subsection{A Symbiotic Framework}
Policy shaping and deliberative planning are not competing paradigms but
complementary components of a unified decision-making framework. The
RL-trained policy provides a strong inductive bias, supplying reusable
heuristics that constrain the search space, while search performs focused
deliberation to adapt these heuristics to the current problem instance. This
interaction can be formalized as a bi-level optimization, where an outer
learning loop anticipates the outcome of an inner planning process over a task
distribution $\mathcal{I} \sim \mathcal{D}$.

The objective is to learn policy parameters $\theta^*$ that maximize the true
task reward $R_{\text{true}}$ of plans produced by search:
\begin{equation}
\label{eq:symbiotic_objective}
\begin{aligned}
\theta^* &= \mathop{\arg\max}_\theta\;
\mathbb{E}_{\mathcal{I} \sim \mathcal{D}}
\bigg[
R_{\text{true}}\bigg( \\
&\quad \mathop{\arg\max}_{p \in \mathcal{P}_{\text{plan}}} 
\Big( \sum_{t=0}^{T-1} \gamma^t
R_{\text{ext},\mathcal{I}}(s_t, a_t) \\
&\quad + \mathcal{H}_\theta(s_T, p) \Big)
\bigg)
\bigg].
\end{aligned}
\end{equation}

This formulation highlights the coupling between learning and planning: the
outer optimization shapes a heuristic $\mathcal{H}_\theta$ that is maximally
useful for inner-loop search, while the inner loop must generate plans that
optimize the external objective $R_{\text{true}}$.

\clearpage
\begin{figure*}[!b]
    \centering
    \resizebox{0.96\textwidth}{!}{
    \begin{forest}
        forked edges,
        for tree={
            grow=east,
            reversed=true,
            anchor=base west,
            parent anchor=east,
            child anchor=west,
            base=left,
            font=\large,
            rectangle,
            draw=hidden-black,
            rounded corners,
            align=left,
            minimum width=4em,
            edge+={darkgray, line width=1pt},
            s sep=3pt,
            inner xsep=2pt,
            inner ysep=4pt,
            line width=1.1pt,
            ver/.style={rotate=90, child anchor=north, parent anchor=south, anchor=center},
        },
        where level=1{text width=9.5em,font=\normalsize,}{},
        where level=2{text width=11.5em,font=\normalsize,}{},
        where level=3{text width=12em,font=\normalsize,}{},
        [Taxonomy, ver
            [MCTS for \\ Direct Test-Time \\ Enhancement
                [General Reasoning \\ \& Problem Solving
                    [{ 
                        Discriminator-based Tree Search~\citep{chen2024tree}, 
                        Interpretable Contrastive MCTS~\citep{gao2024interpretable}, \\
                        RoT~\citep{hui2024rot}, 
                        LiteSearch~\citep{wang2024litesearch}, 
                        MindStar~\citep{kang2024mindstar}, \\
                        MARCO-O1~\citep{zhao2024marco}, 
                        Everything of Thoughts~\citep{ding2023everything}, 
                        CoAT~\citep{pan2025coat} 
                    }, leaf3]
                ]
                [Mathematical Reasoning
                    [{ 
                        MCTS Self-Refine~\citep{zhang2024accessing}, 
                        Energy Function MCTS~\citep{xu2023no}, 
                        OVM~\citep{yu2023ovm},  \\
                        LLaMA-Berry~\citep{zhang2025llama}, 
                        Automated Process Supervision~\citep{luo2024improve}, \\
                        Constrained MCTS~\citep{lin2025leveraging},  
                        Markov Chain of Thought~\citep{yang2024markov} \\
                    }, leaf3]
                ]
                [Code Generation \\ \& Software Engineering
                    [{ 
                        RTL Code Gen MCTS~\citep{delorenzo2024make}, 
                        RethinkMCTS~\citep{li2024rethinkmcts},  \\
                        Verified Multi-step Synthesis~\citep{brandfonbrener2024verified},
                        VerMCTS~\citep{brandfonbrener2024vermcts},    \\
                        Code World Models MCTS~\citep{dainese2024generating}, 
                        Planning in NL~\citep{wang2024planning},  \\
                        O1-Coder~\citep{zhang2024o1}, 
                        SRA-MCTS~\citep{xu2024sra}, 
                        SWE-Search~\citep{antoniades2024swe}, \\
                        SWE-Debate~\citep{li2025swe},  
                        MCTS-Judge~\citep{wang2025mcts}, \\
                        APRMCTS~\citep{hu2025aprmcts}, 
                        MCTS-Refined CoT~\citep{wang2025mcts}
                    }, leaf3]
                ]
                [LLM Agents \\ \& Interactive Environments
                    [{ 
                        LATS~\citep{zhou2023language}, 
                        SeLa~\citep{chi2024sela}, 
                        BIDA~\citep{yu2025bida}, \\
                        WebPilot~\citep{zhang2025webpilot}, 
                        Prompt-based MCTS~\citep{yu2023prompt}, 
                        MASTER~\citep{gan2025master},  \\
                        SE-Agent~\citep{lin2025se}, 
                        WebSynthesis~\citep{gao2025websynthesis},  \\
                        AgentSwift~\citep{li2025agentswift}, 
                        HALO~\citep{hou2025halo} 
                    }, leaf3]
                ]
                [Knowledge-Intensive \\ \& RAG Tasks
                    [{ 
                        KNOT-MCTS~\citep{wu2023knot}, 
                        Search-in-the-Chain~\citep{xu2024search}, \\
                        Contrastive RAG~\citep{gu2025toward}, 
                        RARE~\citep{tran2024rare}, 
                        CoRaG~\citep{wang2024corag}, \\
                        RITEK~\citep{huang2024ritek}, 
                        AirRAG~\citep{feng2025airrag}, 
                        KBQA-O1~\citep{luo2025kbqa}, \\
                        MCTS-KBQA~\citep{xiong2025mcts}, 
                        Hierarchical RAG~\citep{dou2025enhancing}, \\ 
                        Explainable MCTS~\citep{kowalski2025towards}, 
                        KCTS~\citep{choi2023kcts}, \\
                        RAG-Star~\citep{jiang2024rag},
                        FREESON~\citep{kim2025freeson} 
                    }, leaf3]
                ]
                [Multimodal Reasoning
                    [{ 
                        Mulberry~\citep{yao2024mulberry}, 
                        Progressive Multimodal Reasoning~\citep{dong2024progressive}, \\
                        Dyfo~\citep{li2025dyfo} 
                    }, leaf3]
                ]
            ]
            [MCTS for \\ Self-Improvement \\ via Data Generation
                [ Self-Improvement \\ Foundational Frameworks
                    [{rStar-Math~\citep{guan2025rstar}, 
                    Alphazero-like tree-search~\citep{feng2023alphazero}, \\
                    Curriculum Preference Learning~\citep{wang2024towards}, 
                    Agent Q~\citep{putta2024agent},\\
                    Iterative Preference Learning~\citep{xie2024monte}, 
                    Imagination, Searching, and Criticizing~\citep{tian2024toward}, \\
                    Mutual Reasoning~\citep{qi2024mutual}, 
                    AlphaMath Almost Zero~\citep{chen2024alphamath}, 
                    CPL~\citep{wang2024cpl}, \\
                    Step-level Value Preference Optimization~\citep{chen2024step}, TreeRPO~\citep{yang2025treerpo}, \\
                    Data Influence-Oriented Tree Search~\citep{shi2025efficient}, 
                    MCTS-Refined CoT~\citep{wang2025mcts},  \\
                    ASTRO~\citep{kim2025astro}, 
                    }, leaf ]
                ]
                [General Capabilities \\ \& Alignment
                    [{PPL-MCTS~\citep{chaffin2021ppl}, Value-Guided MCTS~\citep{liu2023don}, ARGS~\citep{khanov2024args},  \\  PromptAgent~\citep{wang2023promptagent}, STAIR~\citep{zhang2025stair}, \\ 
                    Dynamic Rewarding~\citep{singla2024dynamic},  
                    Evolutionary Space Search~\citep{li2024optimizing},  \\
                    APRMCTS~\citep{hu2025aprmcts}}, leaf ]
                ]
                [Scientific \\ \& Specialized Domains
                    [{Self-Play Approach~\citep{guo2024can}, Named Entity Matching~\citep{volkova2024novel},  \\Synthetic Data Generation~\citep{locowic2024synthetic},  
                    Monte Carlo Thought Search~\citep{sprueill2023monte},  \\
                    STRATEGIST~\citep{light2024strategist}, Fast and Slow Thinking~\citep{cheng2025think},  Peptune~\citep{tang2025peptune}, \\
                    Multi-Agent Sampling~\citep{ye2024multi}, 
                    Step-level~\citep{ma2025step}, \\
                    Process Reward-guided Tree Search~\citep{park2024ensembling}, {Think\&Cite}~\citep{li2024think},  
                    SAPIENT~\citep{du2024sapient}, \\
                    Automatic Heuristic Design~\citep{zheng2025monte}, 
                    Trans-Zero~\citep{zou2025trans}, \\
                    Prompt-Based MCTS~\citep{duan2025prompt}, {MedS$^3$}~\citep{jiang2025meds},  
                    MCTSr-Zero~\citep{lu2025mctsr}, \\
                    K-MSE~\citep{zhuang-etal-2025-boosting} 
                    Stepwise Domain Knowledge-Driven Reasoning~\citep{liu2025towards} \\
                    IRIS~\citep{garikaparthi2025iris}, 
                    ChemAgent~\citep{wu2025chemagent}}, 
                    leaf ]
                ]
                [Multimodal Applications
                    [{MCTS-guided Sample Selection~\citep{wang2025sota}, MMC~\citep{liu2025mmc}, MM-PRM~\citep{du2025mm}}, leaf ]
                ]
            ]
            [Advanced Topics and \\ Hybrid Approaches
                [Multi-Agent and \\ Collaborative Search
                    [{ 
                    Multi-agent sampling~\cite{ye2024multi}, \\
                    Process reward-guided tree search~\cite{park2024ensembling}, Webpilot~\cite{zhang2025webpilot}, 
                    MASTER~\cite{gan2025master}, \\
                    Data Influence-Oriented Tree Search~\cite{shi2025efficient}, 
                    Multi-LLM collaborative search~\cite{yang2025multi}, \\
                    KompeteAI~\cite{kulibaba2025kompeteai}, 
                    SWE-Debate~\cite{li2025swe}, \\
                    AniMaker~\cite{shi2025animaker}, 
                    HALO~\cite{hou2025halo}}, leaf4]
                ]
                [Reward Model Design \\ and Optimization
                    [{rStar-Math~\cite{guan2025rstar}, 
                    AlphaZero-like tree-search~\cite{feng2023alphazero}, \\
                    Iterative preference learning~\cite{xie2024monte}, 
                    Step-level q-value models~\cite{zhai2025enhancing}, \\
                    Curriculum Preference Learning~\cite{wang2024towards},
                    Interpretable contrastive MCTS~\cite{gao2024interpretable}, \\
                    Energy function guided MCTS~\cite{xu2023no}, 
                    AlphaMath Almost Zero~\cite{chen2024alphamath}, \\
                    CPL~\cite{wang2024cpl}, 
                    OVM~\cite{yu2023ovm}, 
                    Value-Guided MCTS~\cite{liu2023don},\\
                    Step-Level reward model~\cite{ma2023let}, 
                    Step-level value preference optimization~\cite{chen2024step}, \\
                    Automated process supervision~\cite{luo2024improve}, 
                    MCTS-boosted mathematical reasoning~\cite{ma2025step}, \\
                    {Think\&Cite}~\cite{li2024think}, 
                    STAIR~\cite{zhang2025stair}, 
                    MT-RewardTree~\cite{feng2025mt}, \\
                    Hierarchical Multi-Step Reward Models~\cite{wang2025towards}, 
                    ProMed~\cite{ding2025promed}, \\
                    TreeRPO~\cite{yang2025treerpo}, 
                    Unifying RL and Search-Based TTS~\cite{jin2025your},\\
                    Re-ranking Reasoning Context with Tree Search~\cite{yang2025re}
                    }, leaf4]
                ]
                [Search Efficiency \\ and Dynamics
                    [{Tree search for agents~\cite{koh2024tree}, 
                    Discriminator-dependent tree search~\cite{chen2024tree}, \\
                    Information Directed Tree Search~\cite{chandak2024information}, 
                    RoT~\cite{hui2024rot}, 
                    LiteSearch~\cite{wang2024litesearch},  \\
                    BoostStep~\cite{zhang2025booststep}, 
                    Everything of Thoughts~\cite{ding2023everything}, \\
                    Dual process of fast and slow thinking~\cite{cheng2025think}, 
                     \\
                    BFS-Prover~\cite{xin2025bfs}, 
                    MCTS-Judge~\cite{wang2025mcts}, \\
                    Adaptive branching tree search~\cite{inoue2025wider}, 
                    Retro-Search~\cite{lu2025retro}, \\
                    Bilevel MCTS~\cite{asai2025bilevel}, 
                    Test-Time Depth Adaptation~\cite{li2025skip}, \\
                    Abstraction dropping methods~\cite{schmocker2025time}, 
                    SIGMA~\cite{ren2025sigma}, 
                    AgentSwift~\cite{li2025agentswift}, \\
                    FREESON~\cite{kim2025freeson}, 
                    Structural Entropy Guided Agent~\cite{wei2025structural}
                    }, leaf4]
                ]   
            ]
        ]
    \end{forest}
    }
    \caption{A comprehensive taxonomy of MCTS.}
    \label{fig:MCTS-taxonomy}
\end{figure*}
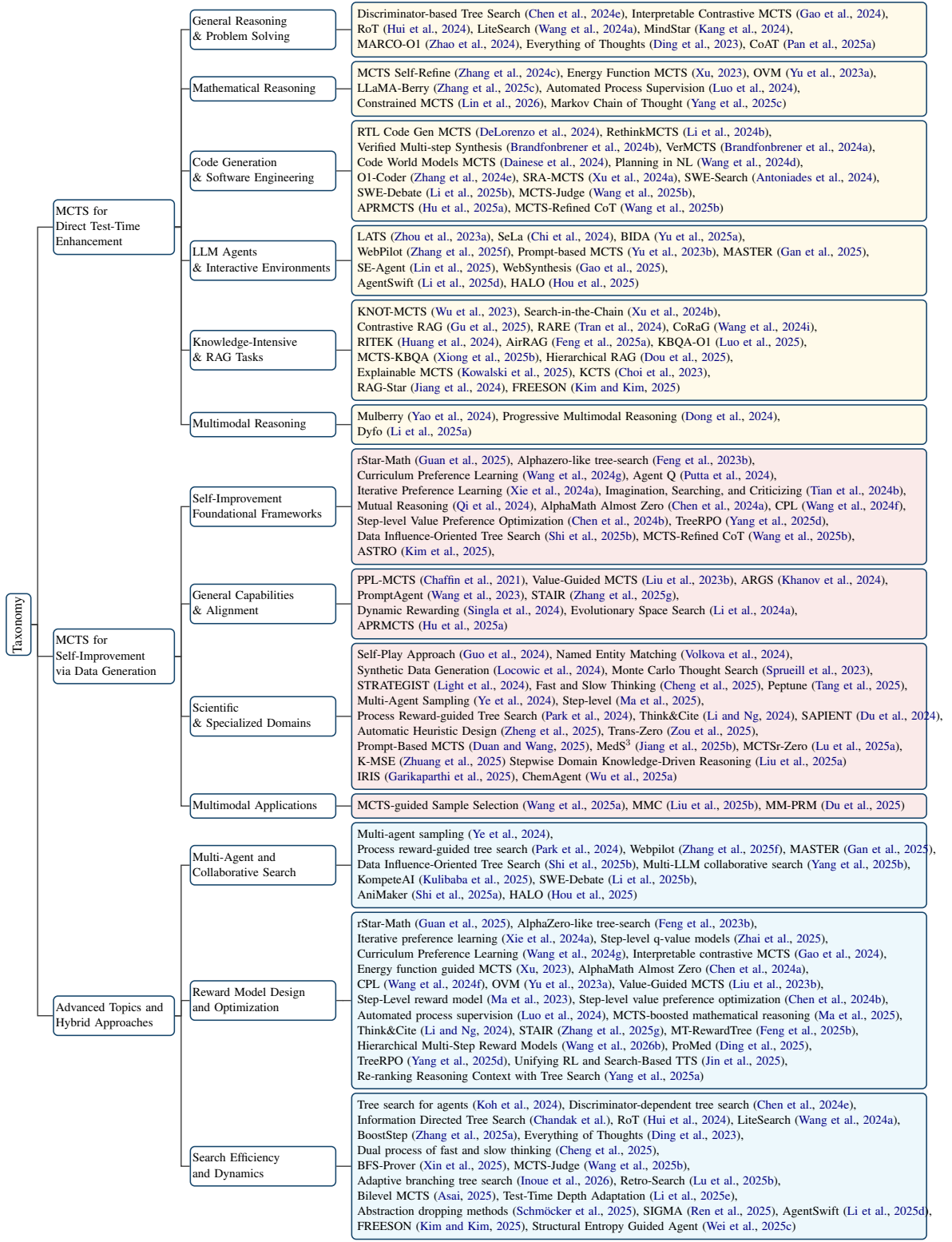

\section{Monte Carlo Tree Search (MCTS)}
\label{appendix:MCTS}


\subsection{Unified Notation and Problem Formation}
We adopt the notation conventions introduced in ReST-MCTS$^*$ \citep{zhang2024rest} to formalize MCTS in the context of LLM reasoning in a {\color{red}\textit{unified manner}}. This approach ensures that all the articles surveyed adhere to a {\color{red}\textit{consistent notation system}} (with minor adjustments to accommodate unique designs), allowing for a clear comparison of their methods without the reader having to navigate the discrepancies in notation.

We first introduce the table of notations:

\begin{table*}[ht]
\centering
\begin{adjustbox}{max width=\textwidth}
\begin{tabular}{>{\columncolor{lightblue}\centering}p{0.15\textwidth}>{\columncolor{lightblue}}p{0.8\textwidth}}
\arrayrulecolor{mycustomblue}
\toprule[1.2pt]
\noalign{\vspace{-3pt}}
\rowcolor{mycustomblue}
\multicolumn{1}{>{\centering\columncolor{mycustomblue}}p{0.15\textwidth}}{\textcolor{white}{\textbf{Symbol}}} & 
\multicolumn{1}{>{\centering\columncolor{mycustomblue}}p{0.8\textwidth}}{\textcolor{white}{\textbf{Definition}}} \\
$Q$ & Input question or problem for which reasoning is being performed \\[0.5em]
$c$ & User prompt or conditioning input used to bias the reasoning traces \\[0.5em]
$a_i$ & Reasoning action at step $i$ generated by the LLM (policy network), where $a_i \in \mathcal{A}$ \\[0.5em]
$s_i$ & Reasoning state at step $i$ resulting from action $a_i$ \\[0.5em]
$p_i$ & Partial reasoning trace up to step $i$, defined as $p_i = [s_1, s_2, \ldots, s_i]$ \\[0.5em]
$r_{s_i}$ & Single-step reward for state $s_i$, measuring its quality independent of previous states \\[0.5em]
$v_i$ & Value of partial solution $p_i$, indicating its potential to reach a correct final answer \\[0.5em]
$T_Q$ & Search tree for problem $Q$, where each node uniquely identifies a reasoning trace \\[0.5em]
$\pi$ & Policy model (LLM) used to generate reasoning steps during tree search \\[0.5em]
$V_\theta$ & Value model that computes partial trace values: $v_i = V_\theta(p_i)$ \\[0.5em]
$R_\theta$ & Reward model that generates single-step rewards: $r_{s_i} = R_\theta(s_i)$ \\[0.5em]
$\mathcal{A}$ & Action space available at state $s_i$, representing all possible next actions \\[1em]
\multicolumn{1}{>{\columncolor{lightblue}\centering}p{0.15\textwidth}}{$C_i$} & 
\multicolumn{1}{>{\columncolor{lightblue}}p{0.8\textwidth}}{\textbf{Search tree node}, represented as $C_i = (t_i, n_i, q_i)$ where:} \\[0.3em]
& $\bullet$ $t_i$: tree node that identifies $C_i$ \\[0.3em]
& $\bullet$ $n_i$: Visit count of node $C_i$, tracking exploration frequency \\[0.3em]
& $\bullet$ $q_i$: Quality value of the partial solution at node $C_i$, indicating its potential to lead to a correct answer \\[-3pt]
\bottomrule[1.2pt]
\end{tabular}
\end{adjustbox}
\caption{Unified Notations for MCTS-Based Methods in LLM Reasoning}
\label{tab:all_notations}
\end{table*}

With this set of notations defined, a search problem in LLM based reasoning can be generalized as finding the correct solution \textit{or} the optimal reasoning trace $p^\prime = [s^\prime_1, s^\prime_2, \cdots, s^\prime_n] $ for a given problem $Q$.

We categorize approaches for finding correct final solution  (a specific terminal state $s^\prime$)  as goal-driven. Goal-driven methods focus primarily on arriving at the correct final answer for given reasoning problems, paying less attention to the reasoning trace that leads to it. In contrast, approaches that aim to identify good or optimal reasoning steps for a given problem are categorized as step-driven. Step-driven methods not only seek to find the correct solution but also emphasize discovering high-quality intermediate steps that contribute meaningfully to the reasoning process and minimize the reasoning distance.

In the search process, the reasoning LLM acts as a policy network $\pi(\cdot|Q, c)$.  where it generates a sequence of reasoning steps or actions to solve the problem $Q$, under a given instruction prompt $c$. The sequence of generated state-action pairs by $\pi(\cdot|Q, c)$  is denoted as $[s_1, a_1, s_2, a_2, s_3, a_3, \cdots, s_n]$, where $s_1$ is the initial state (often a dummy answer or system prompt) and $s_n$ is the terminal state. The terminal state $s_n$ is reached  when \texttt{[eos]} (i.e. end of sequence) token is produced, which may signify the generation of a final answer (correct or incorrect) or the exhaustion of the step limit (e.g. max context length).

Note that, unlike most other reinforcement learning (RL) problems, where an action $a_i$ leads to different states $s_{i+1}$ based on a state transition probability, a reasoning action $a_i$ in LLM-based reasoning deterministically leads to a fixed next reasoning state. This deterministic nature is due to the structure of reasoning (with rare exceptions). As a result, we clarify the usage of certain notations, which may differ from those in typical RL formulations:

\begin{itemize}
    \item A reasoning trace, or partial solution, $p_i$, can be expressed in two equivalent forms:
    \[
    p_i = [s_1, a_1, s_2, a_2, s_3, a_3, \ldots, s_i]
    \]
    or 
    \[
    p_i = [s_1, s_2, s_3, \ldots, s_i].
    \]
    The first form treats actions as distinct from states, while the second combines actions and resulting states into $s$. There is no inherent difference between the two representations, as LLM outputs both $s_i$ and $a_i$ into a sentence in each reasoning step during Chain of Thought. Some looks at it separately  (such as RAP) while others take a joint view (such as ReST-MCTS$^*$).

   \item Unlike traditional RL, where the reward is calculated based on the state-action pair, denoted as $R(a, s)$, and depends on the different state transitions resulting from action $a$, the reward of a single LLM reasoning step can be evaluated based on either the action $a_i$ or the resulting state $s_{i+1}$, or even on state action pairs $(s, a)$, due to the deterministic nature of reasoning (each $a$ deterministically determines $s$).

\end{itemize}

For \textit{simplicity}, we typically consider $s_i$ to be a natural language sentence generated as one chain-of-thought (CoT) reasoning step. Consequently, $p_i = [s_1, s_2, s_3, \ldots, s_i]$ represents a CoT trace consisting of $i$ sentences generated in $i$ sequential steps by LLMs.

During reasoning, a given reasoning state $s_i$ can transition to different next reasoning states $s_{i+1}$, deterministically,  depending on the different action $a_i$ that is chosen (from the action space $\mathcal{A}$) by the LLM policy $\pi$, forming a tree structure, denoted as $T_Q$.

Monte Carlo Tree Search (MCTS) optimizes the search for the reasoning trace $[s_1, s_2, \ldots, s_n]$ in $T_Q$ to find correct answers. Each partial solution trace $p_i = [s_1, s_2, \ldots, s_i]$ forms a unique path (or even node) in this tree, associated with its estimated value $v_i$ and visit count $n_i$. The value $v_i$ defines how promising such partial trace is to reach the correct answer. MCTS process is guided by this promising indicator $v_i$. 

Unsurprisingly, the design and computation of $v_i$ become one of the most critical challenges in search algorithm design for LLM reasoning. Our survey places particular emphasis on the methods used to design the value function $V(\cdot)$ in each of the surveyed papers.

All of the search to be discussed here is done in \textit{Answer Space} of problem $Q$, for the discussion of searching in \textit{Prompt Space} of LLM, refer to Section.

\subsection{Practitioner's Guide: Task-oriented MCTS guide}
\label{app:practitioner_guide}

\noindent
We observe that optimal search configurations---specifically node granularity, evaluation signals, and backpropagation logic---are distinct functions of the task domain's reward sparsity and error propagation characteristics. Table~\ref{tab:task_guide} synthesizes these domain-specific primitives.

\noindent
\textbf{Mathematical Reasoning: Mitigating Variance via Trace-Based Search.}
In mathematical domains, the primary challenge is \textit{error accumulation}, where a single logical fault invalidates the subsequent trajectory. Consequently, relying solely on Outcome Reward Models (ORMs) induces high variance due to "false positives" (correct answers derived from flawed reasoning).
\begin{itemize}
    \item \noindent\textbf{Topology \& Evaluation:} We recommend \textbf{Trace-based nodes} ($p_i=[s_1...s_i]$), enabling the value function to condition on the full derivation history rather than the immediate state. Evaluation should leverage \textbf{Process Reward Models (PRMs)} to verify intermediate steps. In the absence of trained PRMs, methods like \textbf{MCTSr} effectively substitute the model with LLM-based self-refinement.
    \item \noindent\textbf{Backup Dynamics:} The objective is \textit{robustness}. Practitioners should employ \textbf{Average} or \textbf{Sum} backup rules rather than Maximization. A reasoning path is only reliable if the density of correct rollouts is high, thereby filtering out lucky guesses.
\end{itemize}

\noindent
\textbf{Code Generation: Exploiting Binary Oracles.}
Code generation is distinct from reasoning due to the availability of a deterministic oracle (the compiler/test suite). The search objective shifts from maximizing expected utility to ensuring the \textit{existence} of a solution.
\begin{itemize}
    \item \noindent\textbf{Topology \& Evaluation:} \textbf{Terminal-State nodes} are sufficient, as the intermediate logic is often opaque until execution. The primary signal is \textbf{Execution Feedback (ORM)}. Advanced implementations (e.g., \textbf{RethinkMCTS}) integrate verbal feedback from failed tests into the prompt state for subsequent iterations.
    \item \noindent\textbf{Backup Dynamics:} Because the reward signal is binary (pass/fail), \textbf{Max} backup updates ($Q_i \leftarrow \max(Q_i, r_{new})$) are optimal. Finding a single passing solution satisfies the task requirements; the average quality of failed attempts is irrelevant to the final utility.
\end{itemize}

\noindent
\textbf{RAG \& Knowledge Tasks: The Weakest Link Principle.}
Knowledge-intensive tasks require a strict logical conjunction between retrieval relevance and answer correctness. A high-fidelity answer derived from irrelevant documents constitutes a hallucination.
\begin{itemize}
    \item \noindent\textbf{Topology \& Evaluation:} The search space should be modeled via \textbf{State-Action nodes} explicitly separating ``Retrieval'' actions from ``Reasoning'' actions. Evaluation demands a \textbf{Hybrid} signal: a PRM for document relevance and an ORM for factual consistency.
    \item \noindent\textbf{Backup Dynamics:} To enforce factual integrity, we recommend \textbf{Min-based aggregation} ($V(s) = \min(r_{steps})$), as utilized in HiAR-ICL. This enforces a ``weakest link'' logic, ensuring that a hallucination or retrieval failure in any single step penalizes the value of the entire reasoning chain, preventing the propagation of grounded but irrelevant text.
\end{itemize}

\noindent
\textbf{Autonomous Agents: Lookahead in Latent World Models.}
Agents operate in partially observable environments where actions induce irreversible state transitions. MCTS here serves as a planner using the LLM as a simulator.
\begin{itemize}
    \item \noindent\textbf{Topology \& Evaluation:} Nodes must represent \textbf{State-Action pairs} $(s_t, a_t)$, where the LLM functions as a \textbf{World Model} predicting $s_{t+1}$. Effective rewards are composite: $r_t = r_{prob}^\alpha \cdot r_{utility}^{1-\alpha}$, balancing the prior likelihood of an action (naturalness) with its task utility.
    \item \noindent\textbf{Backup Dynamics:} Given the long search horizons, getting stuck in local optima is a significant risk. Practitioners should increase exploration constants ($c_{puct}$) and employ \textbf{Max of Averages} for backup, isolating the single best plan from a diverse set of simulations.
\end{itemize}

\subsection{Advanced Topics and Hybrid Approaches for MCTS}

As the field matures, researchers are exploring more sophisticated techniques that refine the core search paradigm, create better reward signals, and combine multiple methodologies.

\subsubsection{Multi-Agent and Collaborative Search}
Instead of a single LLM performing a search, these approaches use multiple LLM agents that collaborate, debate, or take on specialized roles to solve a problem more effectively. This paradigm shifts from a monolithic searcher to a coordinated team, enabling more robust and diverse problem-solving. For instance, some frameworks use MCTS to orchestrate multiple agents, dynamically adjusting their number and communication based on task complexity \cite{gan2025master}. Others employ hierarchical structures with specialized agents for high-level planning, role design, and low-level execution \cite{hou2025halo}. In competitive settings, such as software issue resolution, multi-agent debate frameworks encourage diverse reasoning paths and lead to more consolidated solutions \cite{li2025swe}. Another collaborative approach, the Mixture-of-Search-Agents (MOSA), leverages the collective expertise of multiple LLMs by combining their independent explorations with iterative refinement, which helps mitigate the limitations of any single model \cite{yang2025multi}.

\subsubsection{Reward Model Design and Optimization}
The success of any search algorithm hinges on the quality of its reward function. This area focuses on designing, training, and analyzing reward models that can accurately guide the search process. A significant trend is the shift from coarse, outcome-based rewards to more granular, step-level feedback. Process-Supervised Reward Models (PRMs) provide this step-by-step guidance, improving reasoning in tasks like mathematics and code generation \cite{ma2023let}. However, annotating these steps is costly, leading to automated data collection pipelines that use MCTS to generate large-scale, step-level supervision data efficiently \cite{luo2024improve}. Research also explores alternatives, such as Outcome-supervised Value Models (OVMs), which are trained only on final outcomes but effectively learn to assess the potential of incomplete reasoning paths, acting as a value function for planning \cite{yu2023ovm}. More advanced hybrid approaches unify reinforcement learning and search by demonstrating that a reward function learned via RL can serve as an ideal PRM for guiding search, eliminating the need for labeled process data \cite{jin2025your}. Other innovations include Hierarchical Reward Models (HRMs) that evaluate both individual steps and their coherence in sequence \cite{wang2025towards} and comprehensive frameworks for building domain-specific reward models, such as for machine translation \cite{feng2025mt}. Analysis of these models reveals counterintuitive findings; for instance, step-level reward models are more adept at assessing the logical coherence of mathematical language than the nuances of natural language descriptions \cite{ma2025step}. A particularly effective design is to \emph{ground} rewards in executable feedback and to combine complementary views of quality: BoostAPR trains dual reward models on execution results to guide automated program repair, showing that heterogeneous, verifiable reward signals are more reliable than a single learned critic \cite{li2026boostapr}.

\subsubsection{Search Efficiency and Dynamics}
A major challenge for tree search is its high computational cost. These works focus on making the search process more efficient and adaptive. To reduce wasted computation, methods like LiteSearch introduce dynamic node selection and node-level exploration budgets based on guidance from a value network \cite{wang2024litesearch}. Algorithmic enhancements, such as bilevel MCTS, can achieve amortized $O(1)$ runtime for node selection, significantly speeding up planning in domains with deep search trees \cite{asai2025bilevel}. Another strategy is to guide the search more intelligently; Information Directed Tree Search (IDTS), for example, uses a Bayesian approach to quantify the information gain from different feedback types, steering the search toward more informative paths \cite{chandak2024information}. The search process can also be made more dynamic and adaptive. Adaptive Branching MCTS (AB-MCTS) dynamically decides at each node whether to "go wider" by expanding new candidates or "go deeper" by refining existing ones, effectively generalizing repeated sampling \cite{inoue2025wider}. Some approaches even adapt the model's architecture at inference time, creating a custom "chain-of-layers" for each sample by skipping or repeating layers from the pretrained model as needed \cite{li2025skip}. Other works focus on improving the quality of reasoning within the search; BoostStep, for instance, enhances single-step reasoning through a step-aligned in-context learning mechanism that provides more relevant examples \cite{zhang2025booststep}. For MCTS variants that use abstractions to simplify the search space, new methods have been proposed to dynamically drop these abstractions in time-critical settings to ensure optimal performance \cite{schmocker2025time}.

\subsection{MCTS for Direct Test-Time Enhancement}
\label{sec:inference_enhancement}

This category includes methods that use Monte Carlo Tree Search (MCTS) primarily to improve the quality of the LLM's output for a single, given prompt at inference time, without updating the model's weights. These approaches treat the generation of a solution as a sequential decision-making problem, where the MCTS algorithm explores a tree of possible reasoning steps or text segments to find an optimal path. The core idea is to leverage lookahead planning to overcome the greedy, left-to-right nature of standard autoregressive decoding, thereby enhancing the model's performance on tasks that require strategic thinking, exploration, or backtracking.

\subsubsection{General Reasoning \& Problem Solving}
This area focuses on creating domain-agnostic frameworks to enhance the fundamental reasoning capabilities of LLMs. Research here aims to make MCTS-based inference more efficient, interpretable, and robust. For instance, some works seek to improve search efficiency by designing more lightweight algorithms or dynamic resource allocation strategies, reducing the substantial computational overhead typically associated with tree search \cite{wang2024litesearch, gao2024interpretable}. Others incorporate meta-cognitive strategies like reflection, where the model learns from previous search experiences within the same problem to avoid repeating mistakes, effectively summarizing successful strategies to guide future steps \cite{hui2024rot}. Another line of inquiry investigates the core components and limitations of tree search, finding that its effectiveness is often contingent on the accuracy of a reward model or discriminator that evaluates intermediate steps \cite{chen2024tree}. To broaden the search space and emulate human-like associative thinking, methods like Chain-of-Associated-Thoughts (CoAT) integrate MCTS with dynamic memory modules, allowing the model to incorporate new information during the reasoning process \cite{pan2025coat}. These general-purpose enhancements treat complex problem-solving as a formal search task, building frameworks that integrate external knowledge and planning capabilities to handle open-ended challenges \cite{ding2023everything, zhao2024marco, kang2024mindstar}.

\subsubsection{Mathematical Reasoning}
Mathematics provides an ideal testbed for MCTS because its problems have clear, verifiable solutions, which simplifies the design of effective reward functions. This verifiability allows for precise feedback on the correctness of intermediate reasoning steps or the final outcome. Many approaches in this domain focus on improving the quality of the reasoning path. For example, MCT Self-Refine (MCTSr) integrates a self-correction mechanism directly into the MCTS loop, allowing the LLM to refine its own reasoning steps during exploration \cite{zhang2024accessing}. Similarly, LLaMA-Berry employs a pairwise preference reward model to globally evaluate and compare different reasoning trajectories, guiding the search toward more promising solutions \cite{zhang2025llama}. Other works focus on the efficiency and scalability of the search process. To handle long chains of thought without excessive computational cost, Markov Chain of Thought (MCoT) compresses previous steps into a concise state representation \cite{yang2024markov}. Some methods circumvent the need for expensive, step-by-step human annotations by training value models on final outcomes alone \cite{yu2023ovm} or by using MCTS to automate the collection of process supervision data \cite{luo2024improve}. To further refine the search, techniques like Constrained MCTS (CMCTS) limit the action space to more rational steps \cite{lin2025leveraging}, while others use lightweight energy functions as path verifiers to guide the search without additional model fine-tuning \cite{xu2023no}.

\subsubsection{Code Generation \& Software Engineering}
In this domain, MCTS is employed to navigate the vast and complex search space of possible code implementations. A significant advantage here is the availability of immediate, objective feedback from external tools like compilers, unit tests, and formal verifiers, which can serve as powerful reward signals. Several works leverage this feedback to guide the search toward correct and efficient code. For instance, RethinkMCTS searches over the reasoning process (i.e., the "thoughts" behind the code) and uses detailed execution feedback to refine erroneous thoughts and steer the search \cite{li2024rethinkmcts}. Going a step further, VerMCTS generates formally verified programs by using a logical verifier to check the correctness of partial programs at each node in the search tree, providing strong guarantees of soundness \cite{brandfonbrener2024verified}. The application of MCTS is broad, spanning from hardware design, where it optimizes for power, performance, and area (PPA) in RTL code \cite{delorenzo2024make}, to complex, repository-level software engineering tasks. In these larger-scale scenarios, multi-agent frameworks like SWE-Search and SWE-Debate use MCTS to manage self-improvement mechanisms and coordinate patch generation \cite{antoniades2024swe, li2025swe}. Beyond code generation, MCTS is also used for automated program repair (APRMCTS) \cite{hu2025aprmcts} and even for evaluating code correctness in an LLM-as-a-Judge paradigm (MCTS-Judge) \cite{wang2025mcts}. These methods often improve performance by searching over abstract plans rather than raw code, which helps generate more diverse and effective solutions \cite{wang2024planning}. Orthogonal to search, reinforcement learning offers a complementary route to the same goal whenever execution feedback is available: BoostAPR optimizes program repair with execution-grounded rewards supplied by dual reward models \cite{li2026boostapr}, while Evo-CuRL organizes training over code lineage graphs so that a curriculum of increasingly difficult software-engineering reasoning tasks is mastered progressively \cite{tan2026evocurl}. Both reinforce the view that verifiable, execution-based supervision is the key ingredient for reliable code reasoning.

\subsubsection{LLM Agents \& Interactive Environments}
For LLM agents operating in interactive environments, where a sequence of decisions is required to achieve a goal, MCTS provides a principled planning mechanism to explore possible action trajectories. These agents must navigate dynamic states and often rely on environmental feedback to guide their choices. A common approach is to use the LLM itself as both a world model to predict future states and a policy to suggest promising actions, effectively combining the LLM's commonsense knowledge with the structured exploration of MCTS \cite{zhao2023large, yu2023prompt}. This paradigm has been successfully applied to complex web navigation tasks, where tree search allows agents to perform explicit exploration and multi-step planning, significantly improving success rates on benchmarks like VisualWebArena and WebArena \cite{koh2024tree, zhang2025webpilot}. To manage the immense search space, some frameworks use learned world models to create simulated environments for efficient planning \cite{gao2025websynthesis} or leverage learned skills to prune the action space \cite{xie2025mirage}. The versatility of MCTS also extends to specialized domains such as automated machine learning (AutoML), where agents like SELA explore different pipeline configurations \cite{chi2024sela}, long-horizon robotic manipulation with progress-aware policy learning \cite{liu2026palm}, and conversational agents, where MCTS helps plan dialogue actions to ensure conversations are both goal-oriented and compliant with predefined procedures \cite{li2024planning}. These frameworks, like Language Agent Tree Search (LATS), unify reasoning, acting, and planning, often incorporating self-reflection to enhance decision-making \cite{zhou2023language}. Beyond planning, the long-term memory of LLM agents itself calls for governance mechanisms, from persisting verifier signals for memory governance \cite{wang2026memguard} to state-evolution attribution watermarking that safeguards long-term memory systems \cite{zhang2026memmark}. Embodied instantiations push these requirements further, since the action space is continuous, contact-rich, and costly to sample in the real world: Focus-Then-Contact uses affordance priors to first localize where to act and then refines contact-rich manipulation with residual reinforcement learning, substantially accelerating real-world robot learning \cite{qiao2026focus}, while SignBot studies human-to-humanoid sign language interaction, where the agent must perceive, plan, and generate temporally extended gestures in tight synchrony with a human partner \cite{qiao2025signbot}.

\subsubsection{Retrieval-Augmented Generation (RAG) \& Knowledge-Intensive Tasks}
In knowledge-intensive tasks, MCTS enhances RAG by transforming the typically static, one-shot retrieval process into a dynamic and iterative reasoning loop. Instead of retrieving all necessary information at the beginning, MCTS-based approaches strategically decide when to query an external knowledge source and what to ask for at each step of the reasoning process. This allows the LLM to build a solution incrementally, using retrieved information to verify facts, fill knowledge gaps, and correct its trajectory. Frameworks like SearChain and RAG-Star explicitly model this process, using MCTS to explore a tree of reasoning steps where each node can trigger a retrieval and verification action \cite{xu2024search, jiang2024rag}. This dynamic integration of retrieval and reasoning is crucial for mitigating hallucinations and improving factual accuracy, especially in complex multi-hop question answering \cite{wu2023knot, choi2023kcts,min2025unihgkr}. The search can be structured to navigate complex knowledge bases \cite{luo2025kbqa, xiong2025mcts, huang2024ritek} or to select an optimal combination of retrieved text chunks to feed into the LLM's context \cite{wang2024corag}. Some innovative approaches, like FREESON, even empower the LLM to perform the retrieval itself by traversing the corpus using a specialized MCTS, eliminating the need for a separate retriever model \cite{kim2025freeson}. This tight coupling of search and retrieval enhances the deliberative reasoning capabilities of LLMs, allowing smaller models to tackle complex knowledge-intensive tasks effectively \cite{hu2025mcts, dou2025enhancing}. The same retrieve-then-verify principle also underpins fact-sensitive applications beyond question answering, such as fake news detection, where a retrieval-augmented multi-agent pipeline gathers and cross-checks external evidence before committing to a verdict \cite{li2026retrieval}.

\subsubsection{Multimodal Reasoning}
For tasks that require reasoning over both text and other modalities like images or video, MCTS serves as a powerful tool to explore the complex interplay between different data types. It helps to structure the reasoning process by breaking down a multimodal problem into a sequence of steps, where each step can involve grounding textual concepts in visual evidence. For example, the AR-MCTS framework uses an active retrieval mechanism within the MCTS loop to fetch relevant supporting insights from a hybrid-modal corpus at each reasoning step, ensuring that the generated explanation is well-supported by both visual and textual facts \cite{dong2024progressive}. Other approaches, such as AStar, leverage MCTS in a training-free manner to first abstract a library of high-level reasoning patterns, or "thought cards", from a small set of example problems. During inference, the most relevant thought card is retrieved to provide a strategic scaffold for solving a new multimodal problem, guiding the model's reasoning process without requiring extensive fine-tuning. Some works also explore using multiple models in a collaborative MCTS framework to jointly search for the best reasoning path, leveraging collective intelligence to tackle difficult multimodal questions \cite{yao2024mulberry, yu2025core3d}. By systematically exploring how to combine and re-rank multimodal reasoning contexts, these methods make vision-language models more robust and capable of handling complex, multi-step visual reasoning \cite{yang2025re}. Beyond visual reasoning, recent work probes the cognitive abilities of MLLMs through scientific exams spanning perception, understanding, and reasoning \cite{zhou2025scientists}, pushes toward cognitive supersensing in multimodal LLMs \cite{li2026toward}, improves spatial reasoning in VLMs via fine-grained preference optimization \cite{shen2026fine}, and unifies 3D understanding and generation through elastic semantic anchoring \cite{yu2026elsa3d}. Understanding and mitigating object hallucination in vision-language models, e.g., by uncovering its dual-pathway circuits, further improves the reliability of multimodal reasoning \cite{liu2026dual}. Retrieval augmentation is likewise being coupled with multi-agent coordination to verify cross-modal claims, as in multimodal fake news detection \cite{li2026retrieval}.

\subsection{MCTS for Self-Improvement via Data Generation}
\label{sec:self_improvement}

This powerful paradigm uses MCTS not just to find a single good answer, but to generate high-quality reasoning trajectories. These trajectories are then used as synthetic data to fine-tune the LLM or a reward model, creating a virtuous cycle of self-improvement.

\subsubsection{Foundational Self-Improvement Frameworks}
These papers introduce the core methodologies for using MCTS within a self-training loop, often inspired by reinforcement learning concepts like AlphaZero and preference optimization. A central theme is the creation of a self-evolutionary cycle where a policy model (the LLM) and a value/reward model are iteratively improved. For example, frameworks like rStar-Math and AlphaLLM use MCTS to perform extensive rollouts, generating vast amounts of verified, step-by-step reasoning data that is then used to train both the LLM and a process preference model \cite{guan2025rstar, tian2024toward}. This AlphaZero-like approach, where the model learns from its own planned-out explorations, can be adapted to various tasks and model sizes, leveraging a learned value function to guide the search more effectively than relying on a pretrained LLM's priors alone \cite{feng2023alphazero}. The data generated from MCTS rollouts is often formatted into preference pairs (i.e., comparing a better reasoning step to a worse one) and used with algorithms like Direct Preference Optimization (DPO) to update the model's policy \cite{xie2024monte, chen2024step}. This process can be entirely self-contained, as demonstrated by frameworks like AlphaMath, which automatically generate both process supervision and step-level evaluation signals without any human or superior-model annotations \cite{chen2024alphamath}. These methods often focus on learning from both successful and unsuccessful trajectories to enhance generalization \cite{putta2024agent, yuan2025agent} and use the search process to explicitly find and correct errors, thereby teaching the model robust recovery skills \cite{kim2025astro, wang2024towards}.

\subsubsection{General Capabilities \& Alignment}
MCTS is used to generate synthetic data for enhancing core LLM capabilities and ensuring alignment with human values~\cite{cao2023multi,chen2021self,zhao2025timeseriesscientist,chen2021adaptive,sun2025docagent,wen2025beyond,you2025uncovering,cao2024multi,guo2026csrv2,wei2026think,guo2026no}. This includes optimizing prompts, where frameworks like PromptAgent treat prompt engineering as a strategic planning problem and use MCTS to explore the space of possible instructions, learning from errors to generate expert-level prompts \cite{wang2023promptagent}. A similar search-based optimization can be used for tuning-free self-alignment, crafting optimal alignment instructions at inference time without costly model updates \cite{singla2024dynamic}. In the context of safety, MCTS can generate step-level reasoning data to teach models how to identify and mitigate risks, balancing helpfulness and harmlessness \cite{zhang2025stair}. The data generation process can also be used for instruction tuning, where MCTS helps explore the "evolutionary space" of instructions to synthesize high-quality, diverse, and complex training data \cite{li2024optimizing}. By generating data from MCTS trajectories that include both successes and recoveries from failure, models can be trained to be more robust and reflective agents. Some methods guide generation with a discriminator to ensure outputs adhere to constraints like non-toxicity \cite{chaffin2021ppl}, while others leverage the value model from a prior alignment process (like PPO) to guide the search \cite{liu2023don, khanov2024args}.

\subsubsection{Scientific \& Specialized Domains}
The self-improvement paradigm is being adapted to a wide array of specialized domains~\cite{liu2023medical,zhou2023survey,hu2025survey,zhang2026postergen,liang2025slidegen,xiong2025quantagent,shen2026decoding,Huang2026HarmonyCellAS}. This includes generating high-quality synthetic tabular data \cite{locowic2024synthetic} and augmenting classical ensembles such as random forests with LLMs for few-shot tabular learning \cite{yang2026forestllm}, creating data for multi-agent collaboration \cite{ye2024multi}, and developing domain-specific models through self-evolution, such as for clinical reasoning in medicine \cite{jiang2025meds}. In conversational AI, MCTS-generated dialogue plans are used to train more strategic and effective recommender agents \cite{du2024sapient}. The approach is also used at a meta-level, for tasks like discovering optimal heuristics for optimization problems \cite{zheng2025monte} or even optimizing hyperparameters for fine-tuning \cite{volkova2024novel}. In strategic domains like game-playing, MCTS guides the learning of high-level strategies through self-play simulations \cite{guo2024can, light2024strategist}. While some applications use MCTS strictly for test-time guidance in specialized areas like therapeutic peptide generation \cite{tang2025peptune} or catalyst design \cite{sprueill2023monte}, the broader trend is to use the explored trajectories to create a feedback loop that continually improves the model's domain-specific expertise.
Similarly, in molecular structure elucidation, K-MSE~\cite{zhuang-etal-2025-boosting} leverages MCTS to enhance LLMs with a knowledge base and a molecule-spectrum scorer, significantly improving their chemical reasoning capabilities. This is also seen in multilingual translation, where MCTS is used to generate synthetic data without parallel corpora \cite{zou2025trans}, and in educational applications for generating personalized test questions \cite{wu2025personalized}.

\subsubsection{Multimodal Applications}
The data generation paradigm extends to multimodal contexts, where MCTS is used to enhance the reasoning capabilities of Vision-Language Models (VLMs)~\cite{feng2022kergnns,you2021self,liu2023llmrec,zhou2023attention,you2021knowledge,you2022end,you2020towards,liu2021aligning,you2021mrd,you2020contextualized,you2024calibrating}. To overcome the lack of fine-grained supervision in multimodal reasoning, MCTS-based pipelines can automatically generate millions of step-level annotations for training powerful process reward models (PRMs) without human labeling \cite{du2025mm}. Another approach involves creating a multimodal actor-critic framework where MCTS guides an actor model to explore diverse reasoning paths. An annotator model then compares pairs of paths-one leading to a correct outcome and one to an incorrect one-to generate critique data that teaches the VLM to correct its own errors \cite{liu2025mmc}. An alternative, data-efficient strategy uses MCTS to quantify the difficulty of visual reasoning samples by measuring the number of search iterations required to solve them. This allows for the selection of a small but highly informative subset of challenging examples for reinforcement fine-tuning, achieving state-of-the-art performance with significantly less data \cite{wang2025sota}. Beyond reasoning enhancement, multimodal LLMs are also applied to perceptual judgment tasks such as learning human-perceived fakeness in AI-generated videos \cite{fu2025learning}, and to knowledge-faithful generative perception, where coverage-guided cross-scale re-indexing addresses cases where generated images look right but retrieve wrong \cite{dong2026generated}.

\section{Informed Search Based Method }
\label{appendix:Informed}

To enhance the reasoning capabilities of Large Language Models beyond simple sequential generation, researchers have increasingly turned to informed search algorithms. This paradigm structures problem-solving as a tree traversal, where heuristic guidance helps navigate vast and complex solution spaces efficiently. Early frameworks such as Tree-of-Thoughts (ToT) adapted classical algorithms like Breadth-First Search (BFS) and Depth-First Search (DFS), using the LLM itself to evaluate intermediate 'thoughts' and prioritize promising reasoning paths. Building on this, more recent approaches have implemented A* search, a more sophisticated heuristic method, to further optimize exploration. Methods like ToolChain* and Q* exemplify this trend by designing intricate cost and heuristic functions that incorporate memory, self-consistency, and learned value estimates to guide the search for optimal solutions. This section explores these key informed search strategies, detailing how they formalize and direct the LLM's reasoning process.

\subsection{Informed BFS/DFS}
The Tree-of-Thoughts (ToT) framework \citep{yao2023tree} enables Large Language Models (LMs) to systematically explore multiple reasoning paths. It formulates problem-solving as a tree search, where each node is a state $s=[x, z_{1\dots i}]$ comprising the input $x$ and a sequence of thoughts $z_{1\dots i}$ generated thus far. The ToT framework is defined by four key components: problem structuring, thought generation, state evaluation, and a search strategy.

The framework first \textbf{decomposes} the problem into intermediate steps. Then, at each step $i+1$, a generator $G(p_\theta,s,k)$ produces $k$ candidate thoughts from a given state $s=[x, z_{1\dots i}]$ using an LM $p_\theta$. This generation occurs via two distinct methods: (1) \textbf{sampling} $k$ independent and identically distributed (i.i.d.) thoughts from a Chain-of-Thought (CoT) prompt, a method effective for expansive thought spaces (e.g., text generation); or (2) \textbf{proposing} thoughts sequentially using a "propose prompt" to prevent redundancy, which is better suited for constrained reasoning tasks. To guide the search, an evaluation function $V(p_\theta, S)$ leverages an LM $p_\theta$ to provide heuristic assessments of progress for a set of states $S$. The evaluation can be performed in two modes: (1) a \textbf{value-based} approach, where each state is scored independently, yielding a scalar or categorical assessment; or (2) a \textbf{voting-based} approach, where the LM selects the most promising state from the set $S$.

ToT implements two primary search algorithms. The \textbf{informed Breadth-First Search (BFS)} algorithm emulates a beam search, maintaining a beam of $b$ states at each step. This process constrains the number of states at any depth to $b$, avoiding exponential growth and making it efficient for problems with a fixed depth $T$. In contrast, the \textbf{informed Depth-First Search (DFS)} algorithm explores a single path until its value, as determined by the evaluator, falls below a threshold, at which point the path is pruned.

Building on these foundational search strategies, recent works have adapted BFS-style exploration for a variety of specialized domains. In causal discovery, LLM-guided BFS has been employed to efficiently uncover causal graphs from both textual knowledge and observational data, using dynamic scoring and active learning to navigate the hypothesis space \citep{jiralerspong2024efficient, susanti2025can, zanna2025uncovering}. Beyond structured discovery, researchers have also explored the LLM's intrinsic capacity for search. For instance, the Autonomous Tree-Search (ATS) paradigm demonstrates that LLMs can execute a BFS-like exploration internally with a fixed system prompt, eliminating the need for external control logic \citep{zhang2023autonomous}. Other work has proposed LLM-First Search (LFS), where the model itself dynamically decides whether to broaden the search (go wider) or deepen the current path, offering a more adaptive alternative to the fixed beam width of ToT-BFS \citep{herr2025llm}. In more fundamental architectural explorations, a novel paradigm called Coconut (Chain of Continuous Thought) has shown that by reasoning in a continuous latent space, LLMs can implicitly perform BFS to explore multiple reasoning steps simultaneously \citep{hao2024training}. For highly structured domains like automated theorem proving, BFS-Prover integrates Best-First Search with an expert iteration framework, achieving state-of-the-art results by strategically filtering problems and refining its policy with Direct Preference Optimization (DPO) \citep{xin2025bfs}.

\subsection{A*}
To mitigate the computational overhead associated with methods like Monte Carlo Tree Search (MCTS), recent work has explored A*-based search algorithms. These methods have been particularly prominent in robotics, where frameworks like LLM-A* leverage the commonsense knowledge of LLMs to generate heuristics for path planning, synergizing the precise pathfinding of A* with the global reasoning of LLMs \citep{meng2024llm}. Notably, ToolChain* \citep{zhuang2023toolchain} and Q* \citep{wang2024q} apply A* search at inference time for general reasoning tasks. 

These methods guide exploration using a specialized cost function $f(n) = g(n) + h(n)$, which prioritizes nodes that appear to be on the most promising path to a solution. This function balances the cost of the path taken so far, $g(n)$, with an estimated cost to reach the goal, $h(n)$. The primary innovation in methods like ToolChain* \citep{zhuang2024toolchain} and Q* \citep{wang2024qimprovingmultistepreasoning} lies in constructing composite heuristics for $g(n)$ and $h(n)$ from diverse, LLM-relevant signals. The key components used to formulate these cost functions are summarized in Table~\ref{tab:astar_heuristics}. The key components used to formulate these cost functions are summarized in Table~\ref{tab:astar_heuristics}.

\begin{center}
\textbf{ToolChain*}
\end{center}

In ToolChain*, the cost function for a node $n$ is the standard A* formulation, $f(n) = g(n) + h(n)$, where $g(n)$ is the \textbf{cumulative cost} from the start node to $n$, and $h(n)$ is a heuristic estimate of the \textbf{future cost} to the goal. The cumulative cost $g(n)$ is the sum of single-step costs over all ancestors of $n$, denoted $an(n)$. Each single-step cost is derived from two value functions, $g_{t,1}$ and $g_{t,2}$, whose outputs are bounded in $[0,1]$. The cost is formulated as the geometric mean of the complements of these values. The cumulative cost is thus:
\begin{equation}
\label{eq:toolchaingn}
g(n) = \sum_{i\in \{an(n),n\}}(1-g_{t,1}(i))^\alpha \cdot (1-g_{t,2}(i))^{1-\alpha},
\end{equation}
where the hyperparameter $\alpha$ weights the contribution of each value function.

The first value function, $g_{t,1}(n)$, is task-specific and draws from a \textbf{long-term memory} $\mathcal{M}$, which is initialized with seed demonstrations and augmented with successful plans discovered during search. Each memory entry $m_j$ is a plan sequence $(s_{j,0}, a_{j,1}, \dots, a_{j,T_j})$. This function evaluates the current plan $s_n$ by computing its maximum longest common subsequence (LCS) score against all plans in memory: $g_{t,1}(n)=\max_{m_j\in\mathcal{M}}\frac{\text{LCS}(s_n,m_j)}{\min(L(s_n),L(m_j))}$, where $L$ is the sequence length. The second value function, $g_{t,2}(n)$, is based on \textbf{self-consistency frequency}. It measures the frequency with which node $n$ is proposed as the next step across $k$ independently sampled reasoning paths, reflecting its reliability.

The \textbf{future cost} $h(n)$ is formulated analogously to $g(n)$:
\begin{equation}
\label{eq:toolchainhn}
h(n) = \sum_{i\in \{an(n),n\}}(1-h_{t,1}(i))^\beta \cdot (1-h_{t,2}(i))^{1-\beta},
\end{equation}
where $\beta$ is the geometric mean weight. The first heuristic, $h_{t,1}(n)$, leverages the \textbf{long-term memory} $\mathcal{M}$. For an action node $n$, it finds the action $a$ in each memory plan $m_j$ with the highest lexical similarity to $n$. The heuristic is the sum of these actions' relative positions: $h_{t,1}(n)=\sum_{m_j\in\mathcal{M}} \mathbf{1}_{{a\in m_j}}\frac{pos(a,m_j)}{T_j}$. The second heuristic, $h_{t,2}(n)$, is an \textbf{LLM imagination score}. An LLM generates a plausible future plan toward a target node $n_T$, and the heuristic value is the ratio of the current path length to the total imagined path length: $h_{t,2}(n)=\frac{|an(n)|}{|an(n_T)|}$, where $|an(\cdot)|$ is the number of ancestors. A higher score signifies closer proximity to the goal.

\begin{center}
\textbf{Q*}
\end{center}

In Q*, the cost function is $f(n) = g(n) + \lambda h(n)$, where $\lambda$ is a weighting hyperparameter. The accumulated cost $g(n)$ is an aggregation of process-based rewards for the current node and its ancestors: $g(n) = \text{Agg}(\{\mathcal{R}(s) \mid s \in an(n) \cup \{n\}\})$. The reward function $\mathcal{R}$ can be derived from human feedback, ground-truth labels, predefined rules, or LM logit scores. The aggregation function, $\text{Agg}$, can be chosen from $\{\max, \min, \sum, [-1]\}$, where $[-1]$ indicates selecting the reward of the last node.

The heuristic cost $h(n)$ is a Q-function that estimates the expected future reward. As an exhaustive search over subsequent steps is intractable, the heuristic is approximated by taking the maximum Q-value among the top-$k$ actions proposed by the LLM policy $\pi_\theta$: $h(n) = \max_{a_t \in \text{top-k}(\pi_\theta(\cdot|n))} Q(n, a_t)$. A primary challenge is estimating optimal Q-values when the frozen policy $\pi_\theta$ is suboptimal. The authors propose three methods for learning a proxy Q-value model: (1) offline reinforcement learning on curated data, (2) learning from MCTS rollouts, or (3) distillation from a stronger LLM. However, this approach may have limited generalization, and the anticipated computational savings are not guaranteed.

\section{Unified Evaluation and Compute Accounting for Tree-Search}
\label{appendix:evaluation_compute}

To characterize the current capabilities of Tree-Search Test-Time Scaling (TTS), we select \textbf{mathematical reasoning} as the representative domain. We prioritize this domain because, unlike open-ended generation, mathematical problems offer deterministic success criteria, enabling high-resolution analysis. While our case study focuses on GSM8K and MATH, the fragmentation it reveals in reporting and compute accounting is systemic~\citep{kaplan2020scaling,hoffmann2022chinchilla,snell2024testtime}. Consequently, the framework we propose here is deliberately \textbf{domain-agnostic} and intended as a reusable standard.

\subsection{The Landscape of Mathematical Reasoning and the Infeasibility of Retrospective Comparison}

\begin{figure*}[h]
    \centering
    \includegraphics[width=0.85\linewidth]{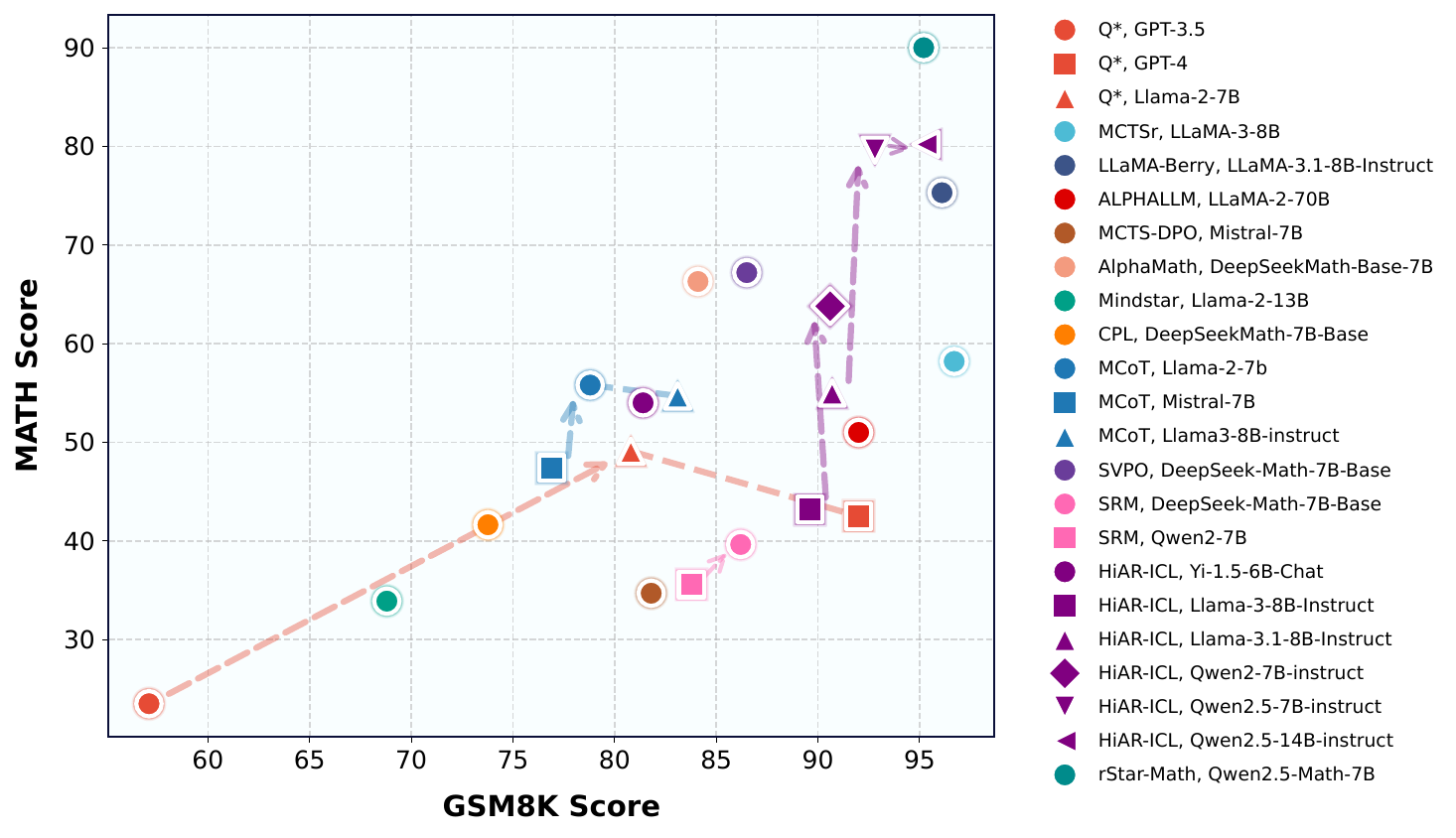}
    \caption{Performance landscape of tree-search methods across GSM8K and MATH. \textbf{Caveat:} The scatter plot aggregates reported metrics from heterogeneous experimental setups. Due to missing data on verifier costs and unstandardized compute budgets in the original papers, \textbf{re-computing these data points under a unified FLOPs standard is impossible}. This visualization conveys the qualitative state-of-the-art rather than a controlled iso-compute ranking.}
    \label{fig:model_performance}
\end{figure*}

To concretely visualize the state of the art, we examine canonical benchmarks where tree-structured decoding has shown substantial gains~\citep{xie2024mctsdpo,ha2025dsgmcts,guan2025rstar}. As visualized in Figure~\ref{fig:model_performance}, MCTS-based variants like MCTSr and rStar-Math populate a Pareto frontier that dominates standard baselines, reinforcing a form of \emph{model--search equivalence} where smaller models with search rival larger static models.

However, we emphasize that \textbf{a strictly fair, compute-normalized comparison of existing literature is currently infeasible}. Unlike controlled studies~\citep{snell2024testtime}, published tree-search papers exhibit substantial methodological heterogeneity that prevents retrospective normalization. \textbf{First}, verifier costs are frequently opaque; many methods employ deep neural Reward Models without reporting the associated token overhead ($T_{\text{eval}}$), making it impossible to calculate total FLOPs without original logs. \textbf{Second}, hardware platforms diverge significantly (e.g., A100 clusters vs.\ consumer GPUs), rendering wall-clock comparisons invalid. \textbf{Third}, baselines span massive parameter scales ($\sim$7B to 70B+), preventing simple step-based comparisons.

\paragraph{Logical Implication:} Constructing a truly apples-to-apples ranking under a unified protocol would require re-implementing and re-evaluating all surveyed methods from scratch. Such an undertaking constitutes a comprehensive \emph{benchmarking study} in its own right, distinct from the scope of this \emph{methodological survey}. Therefore, rather than attempting an imprecise retrofit of past results, we propose a forward-looking protocol to resolve this fragmentation in future work.

\subsection{Proposed Protocol: A Universal Framework for Compute Accounting (SCRP)}
\label{subsec:compute_recipe}

To address the systemic issues identified above, we propose the \textbf{Standardized Compute-Reporting Protocol (SCRP)}. This protocol provides a minimal, actionable recipe for comparability without requiring retroactive adjustments to baseline data.

\paragraph{Unified Resource Vector and FLOPs Abstraction.}
We disentangle computation by defining a unified budget vector
$\mathbf{B} = (C_{\text{policy}}, C_{\text{eval}}, C_{\text{verify}}, T_{\text{wall}})$,
which separates policy generation, node evaluation, and external verification.
To enable hardware-agnostic comparison, we use FLOPs as the primary independent
variable. For a dense transformer with $P$ parameters, inference cost is
approximated as $C \approx 2PT$. Accordingly, the total compute cost for an
instance $x$ is
\begin{equation}
\label{eq:compute_cost}
\begin{aligned}
\mathcal{C}_{\text{total}}(x) \approx\;&
2 P_{\text{policy}} T_{\text{policy}}(x)
+ 2 P_{\text{eval}} T_{\text{eval}}(x) \\
&+ C_{\text{verify}}(x),
\end{aligned}
\end{equation}
where $T_{\text{policy}}$ and $T_{\text{eval}}$ denote cumulative token counts,
and $C_{\text{verify}}$ captures symbolic or execution-based verification cost.

\paragraph{Standardized Metrics.}
Based on this budget, we recommend reporting three key metrics: (1) \textbf{Budgeted Accuracy (Pass@FLOPs)}, defined as $Q(b) = \mathbb{E}[\text{Acc} \mid \mathcal{C}_{\text{total}} \leq b]$, which explicitly visualizes the trade-off between search depth and accuracy; (2) \textbf{Tokens-per-Solved (TpS)}, a model-agnostic proxy for search algorithm efficiency; and (3) \textbf{Parallelism Efficiency}, the ratio between theoretical FLOPs and realized wall-clock speedup. Adopting SCRP allows future research to produce naturally comparable compute-performance curves, eliminating the opacity that currently plagues the field.

\section{Practitioner's Guide with Unified Notation}
\label{appendix:Walkthrough}

\begin{table}[!ht]
\centering
\caption{Comparison of MCTS Node Representations and Evaluations}
\begin{adjustbox}{max width=\textwidth}
\begin{tabular}{@{}lccc@{}}
\toprule
\textbf{Model Name} & \textbf{Tree Node} & \textbf{Node Evaluation} & \textbf{Evaluation Need} \\ \midrule

\textbf{ReST-MCTS*} & 
\({\color{red}p_i} = (s_1, s_2, \ldots, s_i)\) & 
\( {\color{blue}v_i}= V_\text{LLM}({\color{red}p_i})\) & 
\makecell{Current reasoning trace \\ \( p_i = (s_1, s_2, \ldots, s_i\)) } \\ \midrule

\textbf{RAP} & 
\( {\color{red}(a_i, s_i)}\) & 
\makecell{ ${\color{blue}r_i} = R({\color{red}a_i, s_i}) =r_{i,1}^\alpha \cdot r_{i,2}^{1-\alpha} $} & 
\makecell{Current state-action} \\ \midrule

\textbf{LLaMA-Berry} & 
{\color{red}\(s_i^{\text{terminal}}\)} & 
\makecell{\( {\color{blue}r_i} = \alpha R_{\text{local}}({\color{red}s_i^{\text{terminal}}}) + (1-\alpha) R_{\text{global}}({\color{red}s_i^{\text{terminal}}})\)} & 
\makecell{Current and all previously\\ explored  solution nodes in $T_Q$} \\ \midrule
\textbf{MCTSr} & 
\({\color{red}s_i^{\text{terminal}}}\) & 
\makecell{\({\color{blue}r_i} = \frac{1}{2} \left( \min\limits_{\text{$j \in$ evaluating $n$ times}} R^j_\text{LLM}({\color{red}s^\text{terminal}_i}) + \frac{1}{n} \sum_{j=1}^{n} R^j_\text{LLM}({\color{red}s^\text{terminal}_i}) \right)\) \\ re-evaluating \(n\) times to improve robustness.} & 
\makecell{current solution $s^\text{terminal}_i$} \\
\midrule

\textbf{TS-LLM} & 
{\color{red}\( (a_i, s_i)\)} & 
\makecell{
\(  
\begin{cases} 
{\color{blue}v_i} = V_{\text{LLM}}(p_i = (a_1, s_1, \cdots,{\color{red} a_i, s_i})), & \text{if } s_i \neq \text{solution node}, \\ 
{\color{blue}r_i} = R_{\text{ORM}}({\color{red}a_i^\text{terminal}, s_i^{\text{terminal}}}), & \text{if } s_i = \text{solution node}.
\end{cases}
\)} & 
\makecell{All history states \\ $p_i$ = ($s_1$, $s_2$, $\cdots$, $s_i)$ } \\ \midrule

\textbf{ALPHALLM} & 
{\color{red}\( (a_i, s_i)\)} & 
\makecell{
\(  
\begin{cases} 
v_i = V^\text{future}(p_i = (a_1, s_1, \cdots,{\color{red} a_i, s_i}), & \text{if } s_i \neq \text{solution node}, \\ 
{\color{blue}r_i} = R_{\text{PRM}}({\color{red} s_i}), & \text{if } s_i \neq \text{solution node},
\\
{\color{blue}r_i^\text{terminal}} = R_{\text{ORM}}({\color{red} s_i^{\text{terminal}}}), & \text{if } s_i = \text{solution node}.
\end{cases}
\)}
 & 
\makecell{ current state \(s_i\), \\ history $p_i$ of $s_i$} \\ \midrule

\textbf{PG-TD} & 
\( {\color{red}(a_i, s_i)}\) & 
\makecell{
\({\color{blue}r_i} = 
\begin{cases} 
0, & \text{if } s_i \neq \text{solution node}, \\ 
\text{test cases pass rate: TEST($p_i = ({a_1, s_1, \cdots, \color{red}s_i}$) )}, & \text{if } s_i = \text{solution node}.
\end{cases}
\)} & 
\makecell{$p_i = (a_1, s_1, a_2, s_2, \cdots, s_i)$ \\ and test cases provided  } \\
\midrule
\textbf{rStar} & 
${\color{red}(a_i, s_i)}$ & 
${\color{blue}r_i} =
\begin{cases}
0, & \text{if } s_i \neq \text{solution node}, \\
$\text{mutual agreement rate $R_\text{voting}({\color{red}s_i^\text{terminal}})$}$, & \text{if $s_i$ = \text{solution node}}
\end{cases}$ & 
\makecell{Current state-action \\ $(a_i, s_i)$ }\ \\
\midrule
\textbf{RethinkMCTS} & 
$(a_i, s_i)$ & 
${\color{blue}v_i} =
\begin{cases}
\text{test cases pass rate: TEST($p_i = ({a_1, s_1, \cdots, \color{red}s_i}$) ) }, & \text{if } 0 \leq \text{TEST}(p_i) < 1, \\
\alpha \cdot \text{TEST}(p_i) + \beta \cdot V_\text{LLM}(p_i), & \text{if } \text{TEST}(p_i) = 1,
\end{cases}$ & 
\makecell{History  trace $p_i$ and  \\ public test cases } \\ \midrule
\textbf{HiAR-ICL (PRM)} & 
$(a_i, s_i)$ & 
$r_i = R_{\text{PRM}}(s_i)$ & 
\makecell{Current state-action \\ $(a_i, s_i)$ } \\ 
\midrule

\textbf{HiAR-ICL (ORM)} & 
$(a_i, s_i)$ & 
\makecell{
\({\color{blue}r_i} = 
\begin{cases} 
0, & \text{if } s_i \neq \text{solution node}, \\ R_\text{ORM}(s_i^\text{terminal}), & \text{if } s_i = \text{solution node}
\end{cases} \)}& 
\makecell{Current state-action \\ $(a_i, s_i)$} \\ 
\midrule

\textbf{HiAR-ICL (Self-Consistency)} & 
$(a_i, s_i)$ & 
\makecell{
\({\color{blue}r_i} = 
\begin{cases} 
0, & \text{if } s_i \neq \text{solution node}, \\ \text{VOTING}(s_i^\text{terminal}), & \text{if } s_i = \text{solution node}
\end{cases} \)} & 
\makecell{Terminal solution \\ $s_i^{\text{terminal}}$ } \\ 
\midrule

\textbf{Agent-R} &
{\color{red} $s_i$} &
\makecell{Average reward \( {\color{blue}Q(s_i)} \) from rollouts; \\ based on final environmental \\ reward \( {\color{blue}r(\tau)} \in [0, 1] \)} &
\makecell{Final reward \(r(\tau)\) for a \\ complete MCTS rollout} \\ 
\midrule

\textbf{Retro-Search} &
{\color{red}\(s_i\)} &
{\color{blue}\(V(s_i) = \gamma^{N-i}R(a(s_i), a^{*})\)} &
\makecell{Complete trajectory continuation \\ from step \( s_i \) to final answer} \\ 
\midrule

\textbf{MASTER} &
\( {\color{red}s_i} \) &
\makecell{\( ({\color{blue}r_{0,i}}, {\color{blue}c_{0,i}})= V_\text{LLM-self-eval}({\color{red}s_i})\)} &
\makecell{Current agent's full context \\ (Solution + Validation)} \\
\midrule

\textbf{AB-MCTS} &
\({\color{red}t_{out}}\) &
\makecell{ \({\color{blue}p(r | \{r_n\}_{\text{descendants}})}\) \\ (Posterior predictive dist.) } &
\makecell{History of scores from \\ descendant nodes} \\
\midrule

\textbf{SELT} &
\makecell{\(v\) containing state \({\color{red}s}\) \\ which is the reasoning path} &
\( {\color{blue}\Delta}= \text{Score}_{\text{LLM}}({\color{red}s}, s_{\text{represent}})\) &
\makecell{Current answer \({\color{red}s}\) and \\ representative answers \\ from clusters \(s_{\text{represent}}\)} \\
\midrule

\textbf{TRANS-ZERO} &
\({\color{red}y_i}\) &
\({\color{blue}r(y_i)} = \max_{\omega \in \{x_\omega\}}(S(x_\omega, x_{\text{src}}))\) &
\makecell{Original source text \(x_{\text{src}}\) and \\ the set of reconstructions \(\{x_\omega\}\) \\ from multilingual rollouts} \\ 
\midrule


\textbf{CMCTS} &
\( {\color{red}s_i}\) &
\makecell{\( {\color{blue}r_i} = Q({\color{red}s_{i-1}, a_i}) + V({\color{red}s_i})\) \\ where Q and V are from PRM} &
\makecell{Current state \(s_i\) and \\ previous state-action \((s_{i-1}, a_i)\)} \\

\bottomrule
\end{tabular}
\end{adjustbox}
\label{tab:mcts_Node}
\end{table}

\begin{table}[!ht]
\centering
\caption{Comparison of MCTS value $Q$ update and visit count $n$ update}
\begin{adjustbox}{max width=\textwidth}
\begin{tabular}{@{}lccc@{}}
\toprule
\textbf{Model Name} & \textbf{Tree Node $t_i$} & \textbf{Update (back-propagate) \(Q_i\) using node value} & \textbf{Update $n_i$} \\ \midrule

\textbf{ReST-MCTS*} & 
\(p_i = (s_1, s_2, \ldots, s_i)\) & 
$Q^\text{update}_i \leftarrow \frac{\sum\limits_{j \in \text{Children}(p_i)} n_{j} \cdot {\color{blue}v_{j}}}{\sum\limits_{j \in \text{Children}(p_i)} n_{j}}$ & 
$n_i = n_i + 1 $ \\ \midrule
\textbf{RAP} & $(a_i, s_i)$ & $Q^\text{update}_i \leftarrow$  $\max\limits_{\text{each roll out}\,\substack{s_i, a_i, {\color{blue}r_i}, \ldots,  s_l, a_l, {\color{blue}r_n}, s^\text{terminal}_{l+1}}} \text{avg}({\color{blue}r_i}, {\color{blue}r_{i+1}}, \ldots, {\color{blue}r_n})$ & $n_i = n_i + 1$ \\ \midrule
\textbf{LLaMA-Berry} & $s_i^\text{terminal}$ & \makecell{for $j \in$ Children($s_i^\text{terminal}$) : \\ \ \ \ \ \ \ \ \ $Q^\text{update}_i \leftarrow (1-\gamma)\cdot {\color{blue}Q_i} + \gamma \cdot  {\color{blue}r_j}$ }\\ \midrule
\textbf{MCTSr} & $s_i^\text{terminal}$ & $Q_i^\text{update} = \frac{1}{2} \left( {\color{blue}Q_i} + \max\limits_{j \in \text{Children}(s_i)} {\color{blue}r_j} \right) $\\ \midrule
\makecell{\textbf{TS-LLM (MCTS-$\alpha$)} \\ \textbf{(MCTS-Rollout)}}  & $(a_i, s_i)$ &   \makecell{
\(  Q_i^\text{update} \leftarrow
\begin{cases}
Q_i + \gamma {\color{blue}v_n} & \text{if rollout final $s_n$} \neq \text{solution node}, \\ 
Q_i + \gamma {\color{blue}r_n} & \text{if rollout final $s_n$} = \text{solution node}
\end{cases}
\)}\\ \midrule

\textbf{ALPHALLM} & $(a_i, s_i)$ & \makecell{
$Q_i^\text{update} \leftarrow Q_i + \beta_{\text{v}} \cdot {\color{blue}v_i} + \beta_{\text{PRM}} \cdot {\color{blue}r_i}$ + \\ 
$\beta_{\text{ORM}} \cdot \mathbb{E}_{s^\texttt{terminal}_m \sim \pi_\text{LLM}(s_i)}[{\color{blue}r_m^\text{terminal}}]$ \\ roll-out to terminal node $s_m^\text{terminal}$ n times \\ to estimate expected reward of ORM values.} \\ \midrule

\textbf{PG-TD} & $(a_i, s_i)$ & \makecell{$Q_i^\text{update} \leftarrow \max( Q_i, {\color{blue}r_m})$ \\ where $s_m$ is terminal roll-out state} \\ \midrule
\textbf{rStar} & $(a_i, s_i)$ & \makecell{ $Q_i^\text{update} \leftarrow  {\color{blue}Q_i} + {\color{blue}r_m}$ \\ where $s_m$ is terminal roll-out state} \\\midrule
\textbf{RethinkMCTS} & 
$(a_i, s_i)$ & 
$Q_i^{\text{update}} \leftarrow \max\limits_{\text{j $\in$ Children($(a_i, s_i)$}} (Q_i, {\color{blue}r_j})$
 \\\midrule
\textbf{HiAR-ICL (PRM)} & 
$(a_i, s_i)$ & 
$Q_i^{\text{update}} \leftarrow \alpha \cdot Q_i + (1-\alpha)\cdot  \min ( Q_i, r_{i+1} )$ & 
$n_i \leftarrow n_i + 1$ \\ 
\midrule

\textbf{HiAR-ICL (ORM)} & 
$(a_i, s_i)$ & 
\makecell{ $Q_i^\text{update} \leftarrow \alpha \cdot Q_i + (1 - \alpha) \cdot {\color{blue}r_m}$ \\ where $s_m$ is terminal roll-out state} & 
$n_i \leftarrow n_i + 1$ \\ 
\midrule

\textbf{HiAR-ICL (Self-Consistency)} & 
$(a_i, s_i)$ & 
\makecell{ $Q_i^\text{update} \leftarrow \alpha \cdot Q_i + (1 - \alpha) \cdot {\color{blue}r_m}$ \\ where $s_m$ is terminal roll-out state} & 
$n_i \leftarrow n_i + 1$ \\ 
\midrule

\textbf{Agent-R} &
$s_i$ &
$Q_i^\text{update} \leftarrow \frac{Q_i \cdot n_i + {\color{blue}r(\tau)}}{n_i + 1}$ &
$n_i = n_i + 1$ \\ 
\midrule

\textbf{Retro-Search} & \(s_i\) & 
\makecell{Greedy trajectory replacement, not value backpropagation. \\ 
Replaces if \({\color{blue}V(s_\text{new})} > {\color{blue}V(s_\text{old})}\). } & N/A \\ 
\midrule

\textbf{MASTER} & $s_i$ & $Q_i = {\color{blue}c_{0,i}} \cdot {\color{blue}r_{0,i}} + (1-{\color{blue}c_{0,i}}) \cdot \frac{1}{n_i}\sum_{n=1}^{n_i} {\color{blue}r_n}$ & \makecell{Only on backpropagation: \\ $n_i = n_i + 1$} \\
\midrule

\textbf{AB-MCTS} & $t_{out}$ & \makecell{Update posterior parameters. E.g., for Beta dist: \\ $\hat{\alpha} \leftarrow \tilde{\alpha} + \sum {\color{blue}r_n}$, $\hat{\beta} \leftarrow \tilde{\beta} + \sum (1 - {\color{blue}r_n})$} & \makecell{Implicitly tracks number of \\ observations for posterior update} \\
\midrule

\textbf{SELT} &
\(v\) containing state \(s\) &
\(Q(v) \leftarrow Q(v) + {\color{blue}\Delta}\) &
\(N(v) \leftarrow N(v) + 1\) \\
\midrule

\textbf{TRANS-ZERO} & \(y_i\) & $Q_i^\text{update} \leftarrow Q_i + {\color{blue}r_j}$ \text{ (where } \(r_j\) \text{ is from child node)} & $n_i = n_i + 1$ \\
\midrule


\textbf{CMCTS} & $s_i$ & \(Q_i^{\text{update}} \leftarrow \frac{\sum (\sum_{j=i}^{T} {\color{blue}r_j})}{n_i} \) & $n_i = n_i + 1$ \\

\bottomrule
\end{tabular}
\end{adjustbox}
\label{tab:mcts_evaluations}
\end{table}

\begin{table}[!ht]
\centering
\caption{Reward Model Training: Input Generation, Label Construction, and Datasets}
\begin{adjustbox}{max width=\textwidth}
\begin{tabular}{@{}lcccc@{}}
\toprule
\textbf{Model Name} & 
\makecell{\textbf{Reward Model} \\ \textbf{Input Generation}} & 
\makecell{\textbf{Reward Label} \\ \textbf{Generation}} & 
\textbf{Reward Model} & 
\makecell{\textbf{Reward Model} \\ \textbf{Train Dataset}} \\ 
\midrule

\textbf{ReST-MCTS*} & 
\makecell{
1. SciInstructQuestion, co-training \\ both answers and correct reasoning \\ traces. \\
2. Dataset with only answers; \\ reasoning traces are collected using \\ DFS-based reasoning traces \\ (with Mistral-7B model), generated \\ through breadth-first or depth-first \\ search on verified search trees. 
} & 
\makecell{
For correct reasoning traces, \\ a \textit{cumulative quality value} \\ \(v_k\) is assigned to each \\ partial trace \(p_k\), computed as \\ \(v_k = k / K\), where \(k\) is \\ the current step, and \(K\) is \\ the total number of steps \\ in the correct reasoning trace. \\ For incorrect traces, \(v_k\) values \\ are penalized or set to 0, \\ reflecting the trace's inability \\ to lead to a correct solution.
} & 
\(V(p)\) & 
\makecell{MATH, \\ SciInstructQuestion} \\ 
\midrule

\textbf{TS-LLM (Value Model)} & 
\makecell{
Rollouts generated using \\ a supervised fine-tuned policy \\ (LLaMA2-7B). \\
Inputs are partial trajectories \\ \(p_i = (s_1, s_2, \dots, s_i)\) from \\ sampled reasoning traces \\ across tasks like GSM8K \\ and Game24.
} & 
\makecell{
Target labels are calculated using \\ Temporal Difference (TD-\(\lambda\)) \\ or Monte Carlo (MC) methods. \\
TD-\(\lambda\) uses a weighted sum of \\ \(n\)-step returns and bootstraps \\ with the predicted reward \\ \(V(s_T)\) at the terminal state. \\ MC estimates the full cumulative \\ return directly as the sum of \\ rewards from \(s_i\) to \(s_T\).
} & 
\(V(p)\) & 
\makecell{GSM8K, \\ Game24 Rollouts} \\ 
\midrule

\textbf{TS-LLM (ORM)} & 
\makecell{
Terminal states \(s^{\text{terminal}}\) \\ generated from sampled rollouts \\ of a fine-tuned policy model.
} & 
\makecell{
Labels are assigned using a binary \\ reward based on solution correctness \\ and quality. Correct solutions \\ receive \(+1\), while incorrect ones \\ are penalized with \(-1\). \\ Rewards are derived from a \\ task-specific reward function.
} & 
\(R(s^{\text{terminal}})\) & 
\makecell{GSM8K, \\ Game24 Rollouts} \\
\midrule
\textbf{ALPHALLM (Value Model)} & 
\makecell{
Reasoning traces \\ are sampled from policy LLM (LLaMA).
} & 
\makecell{
Reward \( v_t \) is computed using \\ Temporal Difference (TD) or Monte Carlo \\ (MC) methods by reaching solution nodes\\ and assign scores based on expected correctness \\ of roll-out trace from each node. 
} & 
\( v_{\pi} \) & 
\makecell{GSM8k Game24 \\ and PrOntoQA} \\ 
\midrule

\textbf{ALPHALLM (PRM)} & 
\makecell{
reasoning traces are reused from \\ the value function. Inputs are sampled \\ node values \( (s_t, a_t) \).
} & 
\makecell{
Immediate rewards \( r^{\text{PRM}}_t \) are assigned \\ using prefix sampling with \\ textual templates for intermediate \\ correctness assessments.
} & 
PRM & 
\makecell{GSM8k Game24 \\ and PrOntoQA} \\ 
\midrule

\textbf{ALPHALLM (ORM)} & 
\makecell{
Terminal states \( s^{\text{terminal}} \) from \\ sampled trajectories are used.
} & 
\makecell{
Labels \( r^{\text{ORM}} \) are binary (\(+1\) for correct, \\ \(-1\) for incorrect solutions).
} & 
ORM & 
\makecell{GSM8k Game24 \\ and PrOntoQA} \\ 

\midrule

\textbf{CMCTS} &
\makecell{Uses a pre-trained PRM. \\ Input generation is external \\ to the CMCTS framework.} &
\makecell{Uses a pre-trained PRM. \\ Label generation is external \\ to the CMCTS framework.} &
\makecell{PRM for \\ \(Q(s, a)\) and \(V(s)\)} &
\makecell{External to framework \\ (depends on pre-trained \\ PRM, e.g., from [58]).} \\

\bottomrule
\end{tabular}
\end{adjustbox}
\label{tab:reward_model_training}
\end{table}

\begin{center}
\textbf{ReST-MCTS$^*$}
\end{center}



ReST-MCTS$^*$~\citep{zhang2024rest} adopts a \textbf{\textit{step-driven}} approach, emphasizing the \textbf{\textit{discovery of high-quality reasoning traces and the optimization of intermediate steps}}. Its novelty lies in integrating MCTS with process reward guidance to automatically generate high-quality traces without manual annotation. These traces are then used to iteratively train improved reward and policy (LLM) models. This approach significantly enhances reasoning trace quality, achieving superior performance on datasets such as MATH, GPQA, and CEval compared to baselines that do not leverage MCTS.\\



\textbf{Evaluator-Modeling:}  
A \textbf{\textit{separate}} LLM (Mistral), distinct from the reasoning policy LLM, is used as the value model $V(p_i)$ to \bolditalic{evaluate partial reasoning traces} $p_i = [s_1, s_2, \ldots, s_i]$. The model is fine-tuned on DFS-searched reasoning data with automatically labeled quality scores, capturing the likelihood of $p_i$ leading to a correct solution. During MCTS rollouts, $V(\cdot)$ evaluates each $p_i$ and this value is stored in each node directly. 

\textbf{Evaluation-Function Design:}  
ReST-MCTS$^*$ introduces a weighted reward $w_i$, instead of standard $r_i$, to better reflect the quality and contribution of a single reasoning step (state) $s_i$. The weighted reward is defined as:
\[
w_i = \frac{1 - v_{i-1}}{m_i + 1}(1 - 2r_i),
\]
where $v_{i-1}$ is the value of the previous trace, and $m_i$ is a heuristic measure of the remaining reasoning distance. This approach prioritizes steps closer to the solution, improving exploration of promising paths.

\textbf{MCTS Design:}  
ReST-MCTS$^*$ primarily follows standard MCTS design but introduces a self-critic mechanism during the MCTS expansion phase. Self-critic is also used in deterning termination node in $T_Q$. 

\begin{center}
\textbf{RAP}
\end{center}

RAP~\citep{hao2023reasoning} (\textit{Reasoning as Planning with World Models}) adopts a \textbf{\textit{goal-driven}} approach, focusing on efficiently navigating reasoning paths to directly \bolditalic{achieve correct solutions}. This work innovates by re-purposing the \textit{same} LLM as both a reasoning policy (producing action $a_i$) and a \textit{world model} to simulate state transitions (producing $s_i$ and reward returning ($R(a_i)$ model). By combining MCTS with state-action rollouts guided by the world model, RAP demonstrates improved efficiency and accuracy across tasks, achieving strong performance on benchmarks like Blocksworld and logical reasoning datasets while reducing reliance on pretraining or additional reward models.

\textbf{Evaluator-Modeling:}
A \textbf{\textit{repurposed}} LLM  (LLaMA in this work), which is the same model as policy (reasoning) LLM, is used to generate per-step reward. This model generates not only next state $s_t$ but also returns  reward for each newly derived state action pair $(s_t, a_t)$. This world model (same as reward model $R$) capture both action likelihood and task-specific progress when calculating reward $r_i$. Evaluation occurs at each node ($(s_t, a_t)$ in this case) and focuses solely on the current state-action pair, rather than the full trace, ensuring efficient and lightweight evaluation. 

\textbf{Evaluation-Function Design:}  
RAP introduces a novel per-step reward \(r_i\) = $R(s_i, a_i)$ that combines action likelihood and task-specific evaluations to assess the quality of each reasoning step. The reward is calculated using a weighted geometric mean of two components:  
\begin{equation}
    r_t = r_{t,1}^\alpha \cdot r_{t,2}^{1-\alpha}
\end{equation}
where
\(r_{t,1}\) is the log-probability of the action \(a_t\) as predicted by the policy model, reflecting its confidence in the chosen action. \(r_{t,2}\) is a task-specific score evaluating how well the predicted next state \(s_{t+1}\) (generated by the world model) aligns with the task's objectives. This is often derived using heuristics or domain-specific metrics. 

\bolditalic{The value $Q$  of current node is calculated by using the max value of average future reward of each rolled-out future trace starting from this node }:
\begin{equation}
   Q_i = \max_{\substack{s_t, a_t, r_t, \ldots,  s_l, a_l, r_l, s_{l+1}}} \text{avg}(r_t, \ldots, r_l).
\end{equation}
where $s_t, a_t, r_t, \ldots,  s_l, a_l, r_l, s_{l+1}$ is one specific rollout trace starting from current node $(s_t, a_t)$. 
this value $v$ will be used in MCTS selection. 

\textbf{MCTS-design:} RAP uses this $v$ value for selection process in MCTS and sticks with standard MCTS design for expansion and backprogration.

\begin{center}
\textbf{LLaMA-Berry}
\end{center}

LLaMA-Berry~\citep{zhang2024llama} is strong \textbf{\textit{goal-driven}} approach, emphasizing the refinement of complete reasoning solutions.  It changes the way of building the reasoning tree $T_Q$ by using only final result  $s^\texttt{terminal}$ as each node, and each child node is a refinement of parent solution. It also innovates the evaluation framework by using\textit{Pairwise Preference Reward Model (PPRM)}, using global win-loss matrices and local adjacency comparisons to rank these solutions. By leveraging critiquing and rewriting during tree expansion, LLaMA-Berry ensures accurate self-refinement of solutions, achieving robust performance across explored solution paths.

\textbf{Evaluator-Modeling: }
A \textbf{\textit{fine-tuned small LLM (Gemma2-2B-Instruct)}} serves as the Pairwise Preference Reward Model (PPRM). PPRM, which is $R_\theta(s^\texttt{terminal})$ in out notation system, predicts the quality ($r_i$) of  single state solutions by comparing directly with other solutions. It is trained using ranked reasoning traces and produces a win-loss matrix \(M(i, j)\), where \(M(i, j)\) encodes the preference between two solutions \(s_i\) and \(s_j\). During evaluation, the PPRM gives a ranked score when each new solution is derived. 

\textbf{Evaluation Function Design:}
LLaMA-Berry introduces an evaluation function that combines a \textbf{local reward} (\(v_{\text{local}}\)) and a \textbf{global reward} (\(v_{\text{global}}\)) of a derived solution $s_i$, defined as:
\[
Q_i = R_{\text{local}}(s_i^{\text{terminal}}) + R_{\text{global}}(s_i^{\text{terminal}})
\]

The local value evaluates the quality of \(s_i^{\texttt{terminal}}\) based on its immediate neighbors in the reasoning tree. Pairwise comparisons are performed with the preceding solution \(s_{i-1}^{\text{terminal}}\) and the subsequent solution \(s_{i+1}^{\text{terminal}}\). Using the Pairwise Preference Reward Model (PPRM), the local value is calculated as:
\[
R_{\text{local}}(s_i^{\text{terminal}}) = \frac{1}{2} \left( \text{PPRM}(s_i^{\text{terminal}}, s_{i-1}^{\text{terminal}}) + \text{PPRM}(s_i^{\text{terminal}}, s_{i+1}^{\text{terminal}}) \right)
\]

The global value assesses \(s_i^{\text{terminal}}\) based on its position within the entire set of explored solutions in $T_Q$. A win-loss matrix \(M(i, j)\) is constructed using pairwise comparisons between all solutions:
\[
M(i, j) = 
\begin{cases} 
1 & \text{if } s_i \text{ is preferred over } s_j, \\
-1 & \text{if } s_j \text{ is preferred over } s_i, \\
0 & \text{if they are equally preferred.}
\end{cases}
\]
The global value is computed using the Enhanced Borda Count (EBC), which aggregates the win-loss scores for each solution:
\[
R_{\text{global}}(s_i^{\text{terminal}}) = \sum_{j} M(i, j)
\]
where the sum is taken over all other solutions \(s_j\) in the reasoning tree.

\textbf{MCTS design:} LLaMA-Berry follows standard MCTS principles with significant enhancements in its \textit{expansion}, \textit{evaluation}, and \textit{backpropagation} processes. During expansion, it employs a critiquing and rewriting mechanism that iteratively refines each solution at the node, improving the quality and accuracy of expanded results. The evaluation phase integrates both global win-loss matrix calculations and local adjacency comparisons to determine the combined quality \(Q_i\) of a solution node. For backpropagation, LLaMA-Berry introduces an updated formula to refine \(Q_i\) by aggregating the values from subsequent child nodes in the tree. This ensures that the backpropagated \(Q_i\) reflects both the cumulative utility of explored paths and the quality of individual solutions. 

\begin{center}
\textbf{MCTSr}
\end{center}

MCTSr~\citep{zhang2024accessing} is also a strong \textbf{\textit{goal-driven}} approach that emphasizes generating and refining complete reasoning solutions. It structures the reasoning tree \(T_Q\) where each node represents a terminal solution \(s^\texttt{terminal}\), and edges denote iterative refinement attempts. The key innovation in MCTSr lies in its evaluation framework, where rewards for nodes are computed using a combination of minimum and average reward values from multiple resampling attempts. This approach ensures robustness and fairness in evaluating solutions, significantly improving performance in complex reasoning tasks.

\textbf{Evaluator-Modeling:}
MCTSr uses the same LLM (LLaMA) for reasoning and evaluation, where the evaluation model assigns rewards \(R(s^\texttt{terminal})\) to each terminal solutions through a resampling process. The evaluator repeat its rewarding process \(n\) times for each \(s^\texttt{terminal}\), assigning scores between -100 and 100, which are then aggregated to calculate the final node value. This integration avoids the need for a separately trained reward model, instead leveraging the inherent capabilities of the LLM.

\textbf{Evaluation Function Design:}
MCTSr introduces an evaluation function for terminal solutions that combines the minimum reward from resampling and the average reward over all resampled attempts:
\[
Q_i = \frac{1}{2} \left( \min R(s_i^{\texttt{terminal}}) + \frac{1}{n} \sum_{j=1}^{n} R(s_i^{\texttt{terminal}}) \right)
\]
Here, \(R(s_i^{\texttt{terminal}})\) represents the reward assigned to \(s_i^{\texttt{terminal}}\) for a single evaluation, and \(n\) is the number of sampling of rewards attempts. This formula balances the robustness of worst-case performance (\(\min R\)) with the overall quality (\(\text{avg} R\)) of the solution.

\textbf{MCTS Design:}
MCTSr follows the standard MCTS structure with refinements across the selection, evaluation, and backpropagation phases. 
Each child node represents a refined version of its parent solution, generated by iteratively rewriting \(s^\texttt{terminal}\) using the same LLM during expansion phase.
MCTSr employs an updated backpropagation formula that aggregates the values from child nodes, refining the parent node's value \(Q_i\):
\[
Q'(s_i) = \frac{1}{2} \left( Q(s_i) + \max_{j \in \text{Children}(s_i)} Q(s_j) \right)
\]
This formula ensures that parent nodes reflect the best potential of their children while retaining their intrinsic value.
 Dynamic pruning is applied to remove unpromising nodes based on evaluation scores and exploration criteria, improving computation

\begin{center}
\textbf{TS-LLM}
\end{center}

TS-LLM~\citep{feng2023alphazero} is both \textbf{\textit{goal-driven}} and Step-driven, leveraging an AlphaZero-inspired MCTS framework to optimize reasoning solutions and better steps to re-train policy LLM. A key innovation in TS-LLM is its dual-model evaluation strategy, where the value model evaluates intermediate reasoning paths based on the entire trajectory \(p_i = (s_1, s_2, \cdots, s_i)\), while the Outcome Reward Model (ORM) scores terminal solutions. This design ensures that the reasoning process balances exploration of promising paths with rigorous evaluation of terminal solutions.

\textbf{Evaluator-Modeling:}
TS-LLM employs two separate models for evaluation:
a) A \textbf{value model}, which predicts \(V(p_i)\), the cumulative reward potential of an intermediate trajectory \(p_i = (s_1, s_2, \cdots, s_i)\);
b) An \textbf{Outcome Reward Model (ORM)}, which assigns a reward \(R_{\text{ORM}}(s_i^{\text{terminal}})\) to terminal nodes based on solution quality.

Both models are trained using supervised learning on collected data from MCTS rollouts. The value model focuses on evaluating reasoning trajectories, while the ORM specializes in scoring terminal solutions. 

\textbf{Evaluation Function Design:}
The evaluation function for a node \(s_i\) is determined by whether it is an intermediate or terminal node:
\[
Q_i = 
\begin{cases} 
V_{\text{value-model}}(p_i = (s_1, s_2, \cdots, s_i)), & \text{if } s_i \neq s_i^{\text{terminal}}, \\ 
R_{\text{ORM}}(s_i^{\text{terminal}}), & \text{if } s_i = s_i^{\text{terminal}}.
\end{cases}
\]
This case-based design ensures that intermediate nodes are evaluated based on their cumulative trajectory, while terminal nodes are directly evaluated for correctness and quality.

\textbf{MCTS Design:}
TS-LLM follows closely standard MCTS design, with a additional pruning mechanism. It dynamically removes unpromising nodes based on updated \(Q_i\) values and visit counts, ensuring efficient resource usage.

\begin{center}
\textbf{\centering ALPHALLM}
\end{center}

ALPHALLM~\citep{tian2024toward} is a strong \textbf{\textit{goal-driven}} framework that integrates an AlphaZero-inspired MCTS with multi-critic evaluation to optimize reasoning paths and terminal solutions. Each node in its tree represents a reasoning step paired with an action, \((s_i, a_i)\), and edges denote transitions between steps. Its core innovation lies in the weighted integration of a value model, process reward model (PRM), and outcome reward model (ORM) to evaluate both intermediate and terminal nodes. ALPHALLM demonstrates superior reasoning capabilities across tasks such as GSM8K and Game24, leveraging its multi-critic evaluation to achieve consistent improvements.

\textbf{Evaluator-Modeling:}  
ALPHALLM employs three distinct evaluation modeling:
The \textbf{value model} predicts \(V^{\text{future}}(p_i)\), capturing the potential reward of intermediate trajectories \(p_i = (s_1, s_2, \ldots, s_i)\). This model is trained using temporal difference (TD-\(\lambda\)) learning and Monte Carlo rollouts to evaluate intermediate states.  
The \textbf{Process Reward Model (PRM)} provides immediate feedback \(R_{\text{PRM}}(s_i)\) for each reasoning step \(s_i\), focusing on local step quality. PRM is fine-tuned on prefixes of reasoning traces using step-level rewards.  
The \textbf{Outcome Reward Model (ORM)} assigns $r = $\(R_{\text{ORM}}(s_i^{\text{terminal}})\) to terminal states based on the correctness and quality of solutions. ORM is trained using solution-specific labels derived from task outcomes.  

This multi-critic design allows ALPHALLM to evaluate reasoning steps both locally and globally, ensuring robust guidance during the search.

\textbf{Evaluation Function Design:}  
The node value \(Q_i\) in ALPHALLM combines contributions from all three critics as :
\[
Q_i = \beta_{\text{v}} \cdot V^{\text{future}}(p_i) + \beta_{\text{PRM}} \cdot R_{\text{PRM}}(s_i) + \beta_{\text{ORM}} \cdot \mathbb{E}_{s^\texttt{terminal} \sim \pi_\text{LLM}(s_i)}[\text{ORM}(s^\text{terminal})]
\]
Where expected values for ORM is calculated by monte carlo sampling to reach terminal (solution) state from current state $s_i$ using policy LLM. 

This formula ensures that the evaluation balances trajectory-level exploration with terminal state quality, facilitating robust reasoning and exploration.

\textbf{MCTS Design:}  
ALPHALLM sticks with standard MCTS process except for evaluation phase as we have discussed above. 

\begin{center}
\textbf{\centering PG-TD}
\end{center}
PG-TD~\citep{zhang2023planning} adopts a \textbf{\textit{goal-driven}} approach, leveraging a novel integration of Monte Carlo Tree Search (MCTS) and large language models to improve code generation quality. Its primary innovation lies in using test case execution as the evaluation metric during the generation process, rather than relying on deep learning models. By combining the Transformer's beam search probabilities with P-UCB selection for planning, PG-TD achieves significant improvements in code correctness and efficiency compared to standard decoding methods. This framework has shown strong performance across multiple benchmarks, including APPS and CodeContests, particularly in tasks requiring executable and syntactically valid code.

\textbf{Evaluator-Modeling:}
PG-TD does not use a separate deep learning model for evaluation. Instead, it evaluates nodes during MCTS rollouts by executing test cases on the final generated programs. The outcome (pass rate) of these test cases directly determines the reward. This approach simplifies evaluation while aligning it closely with the end goal of generating functional code. Since the evaluation relies purely on test results, no additional value or reward model training is required.

\textbf{Evaluation-Function Design:}
PG-TD evaluates each node in the reasoning tree by running test cases on final complete program represented by the node. The reward for $s^\text{terminal}$ is average pass rate on all test cases.

\textbf{MCTS Design:}
PG-TD introduces several innovations to standard MCTS design. The selection process uses a P-UCB algorithm, which incorporates the probabilities provided by the Transformer's beam search to balance exploration and exploitation effectively. During expansion, the tree grows by selecting the top-k most probable tokens suggested by the Transformer, reducing the likelihood of syntax errors. Additionally, caching mechanisms (tree structure caching and sequence caching) significantly improve efficiency by reusing previously computed paths and evaluations.

\begin{center}
\textbf{\centering rStar}
\end{center}
rStar~\citep{qi2024mutual} introduces a novel \textbf{\textit{goal-driven}} framework to enhance the reasoning capabilities of small language models (SLMs) by employing a self-play mutual reasoning paradigm. The core idea is to combine a \textit{generation-discrimination} process, where a target SLM generates reasoning trajectories via MCTS, and another SLM verifies the quality of these trajectories through mutual agreement. This approach is particularly effective in overcoming the limitations of SLMs, such as poor exploration and unreliable self-rewarding. Experiments demonstrate significant performance gains across reasoning tasks, with accuracy improvements on benchmarks like GSM8K, MATH, and StrategyQA, surpassing many fine-tuned models.

\textbf{Evaluator-Modeling:}
rStar employs two SLMs for a collaborative evaluation process. The generator SLM performs MCTS-based trajectory generation, while the discriminator SLM verifies the quality of trajectories. The discriminator applies a mutual reasoning consistency mechanism: it is given partial reasoning traces and asked to complete the remaining steps, validating trajectories based on whether the generator and discriminator agree on the solutions. This unsupervised approach eliminates the need for fine-tuned value models or external supervision.

\textbf{Evaluation-Function Design:}
As SLM does not perform well in partial solution evaluation, rStar applys a AlphaGo-like evaluation framework, where intemediate solutions $p_i$ with $s_i$ gets reward of 0 for simplicity, and reward only assigned if $s^\text{terminal}$ is reached, based on the mutual agreement with discriminator SLM.

\textbf{MCTS Design:}
rStar enhances MCTS with a diverse action space inspired by human reasoning processes. Actions include proposing single steps, generating sub-questions, rephrasing problems, and re-answering sub-questions, enabling broader and deeper exploration of solution trajectories. The P-UCB algorithm balances exploration and exploitation, while mutual consistency during node selection ensures robust trajectory validation.

\begin{center}
\textbf{\centering RethinkMCTS}
\end{center}

RethinkMCTS~\citep{li2024rethinkmcts} adopts a \textbf{\textit{goal-driven}} approach, enhancing reasoning-to-code performance by leveraging fine-grained feedback and refining erroneous thoughts during the search process. Its novelty lies in combining Monte Carlo Tree Search (MCTS) with a dual evaluation mechanism and introducing a "rethink" operation, which corrects reasoning errors based on execution feedback. This framework significantly improves the quality of search paths and achieves state-of-the-art performance in code generation tasks, with notable gains on benchmarks like APPS and HumanEval.

\textbf{Evaluator-Modeling:}  
RethinkMCTS employs a dual evaluation framework that uses: 1) \textbf{Public Test Cases (denoted as TEST($p_i$))}: The pass rate of public test cases is calculated to assess the correctness of the generated code.
2) \textbf{LLM Self-Evaluation ($V_{\text{LLM}}$)}: When all public test cases are passed, the LLM provides a self-assessment score to further evaluate the likelihood of correctness for private test cases.

Both components are integrated into a unified evaluation system, which ensures more robust assessments for selecting high-quality nodes during tree exploration.

\textbf{Evaluation-Function Design:}  
The evaluation combines scalar and self-assessment rewards to compute the node value $r_i$:
\[
r_i =
\begin{cases}
v_{\text{test}}, & \text{if } 0 \leq v_{\text{test}} < 1, \\
\alpha \cdot v_{\text{test}} + \beta \cdot v_{\text{llm}}, & \text{if } v_{\text{test}} = 1,
\end{cases}
\]
where $\alpha$ and $\beta$ are weighting parameters (e.g., $\alpha = 0.8, \beta = 0.2$). 

The reward ($r_i$) is assigned to each node. (for $p_i$ that's incomplete or incorrect, TEST($p_i$) will always be 0 so there's no need to differentiate the terminal node and intermediate node)

\textbf{MCTS Design:}
RethinkMCTS incorporates several innovations into the MCTS framework. During the selection phase, the P-UCB algorithm is used to balance exploration and exploitation, where verbal feedback stored at nodes influences subsequent thought refinements. In the expansion phase, nodes that fail public test cases incorporate block-level verbal feedback into the prompts, enabling the LLM to propose new thoughts and assign reasonableness scores to each. The rethink operation is employed for leaf nodes that fail public test cases, refining the current thought based on verbal feedback to correct erroneous paths and improve overall search quality. Finally, during backpropagation, node values $Q_i$ are updated using the maximum reward from child nodes, ensuring that the best paths are prioritized for future exploration. Verbal feedback is stored separately and utilized in the next expansion phase but is not directly incorporated into the scalar reward.

\begin{center}
\textbf{\centering HiAR-ICL}
\end{center}

HiAR-ICL~\citep{wu2024beyond} adopts a versatile and hybrid \textbf{\textit{goal-driven and step-driven}} approach to enhance in-context learning (ICL) by refining reasoning trajectories and leveraging three evaluation strategies: Process Reward Model (PRM), Outcome Reward Model (ORM), and Self-Consistency. This framework introduces hierarchical context construction for iterative refinements and integrates multiple evaluation paradigms to balance intermediate and terminal solution rewards. HiAR-ICL demonstrates significant improvements across reasoning and coding benchmarks, showcasing its adaptability to diverse tasks.

\textbf{Evaluator-Modeling:} HiAR-ICL employs three evaluation mechanisms tailored for different stages of the reasoning process. The Process Reward Model (PRM) evaluates intermediate reasoning steps using a pre-trained language model fine-tuned to assign rewards based on the quality of each step in the trajectory. The Outcome Reward Model (ORM) evaluates terminal solutions using a reward model fine-tuned specifically for outcome-based reasoning tasks. Finally, the Self-Consistency mechanism uses majority voting across multiple reasoning trajectories sampled from the language model to assess terminal solutions. PRM focuses on partial states as inputs and produces scalar intermediate rewards, while ORM and Self-Consistency directly assess terminal states and produce outcome-based rewards.

\textbf{Evaluation-Function Design:} Each evaluation mechanism employs unique reward aggregation strategies. PRM utilizes a Min-based aggregation method where the Process Reward Model evaluates all intermediate steps leading to a state \(s_i\) and calculates the minimum reward across all steps. This emphasizes the weakest link in the trajectory. The reward is defined as:
\[
R_{\text{PRM}}(s_i) = \min_{j=1}^{i} r_{\text{PRM}, j},
\]
where \(r_{\text{PRM}, j}\) represents the reward for the \(j\)-th intermediate step. In contrast, ORM and Self-Consistency directly evaluate terminal solutions using either the Outcome Reward Model or majority voting across sampled traces, respectively, with rewards \(R_{\text{ORM}}(s_m^{\text{terminal}})\) and \(R_{\text{voting}}(s_m^{\text{terminal}})\).

\textbf{MCTS Design:} HiAR-ICL sticks to standard MCTS designs and made changes during backprobgation stages by using a min based value backpropgating approach and a product based reward backpropgating approach.

\begin{center}
\textbf{\centering Agent-R}
\end{center}

The \textbf{Agent-R}~\citep{yuan2025agent} framework is an iterative self-training method designed to improve an agent's ability to recover from errors in interactive environments. Unlike approaches such as RAP that employ an LLM as a learned world model to predict dense, per-step rewards and next states, Agent-R performs post-hoc analysis over complete trajectories to generate corrective training data, turning failure cases into targeted revision examples that refine the underlying policy.

\textbf{Evaluator-Modeling:} Agent-R dispenses with a separately trained reward/evaluator and instead relies on two signals. First, a sparse, terminal \emph{environment reward} $r(\tau)\in[0,1]$ supplies the ground-truth success label for a trajectory $\tau$. Second, the agent's own actor model is repurposed as a critic through ``model-guided critique construction,'' where the policy is prompted to examine a failed trajectory step-by-step, labeling actions as \emph{good}, \emph{bad}, or \emph{uncertain} and identifying the first erroneous step $t'$. This self-critique localizes the source of failure without external experts or a dedicated reward model, in contrast to RAP's dense reward prediction; Agent-R leverages the actor's current competence to diagnose \emph{where} the trajectory went wrong rather than to score \emph{every} step.

\textbf{Evaluation-Function Design:} Full trajectories collected via MCTS are categorized as ``good'' $(\tau^g)$ or ``bad'' $(\tau^b)$ by thresholding their terminal reward, with parameters $\alpha$ and $\beta$ satisfying $r(\tau^b)<\beta<r(\tau^g)\le 1$ and a progressively tightened high-quality bar $\alpha$ (so that $\alpha<r(\tau^g)=r(\tau^r)$ for accepted/revision-worthy traces). The actor-derived transition point $t'$ on a bad trajectory pinpoints the earliest faulty action; Agent-R then splices the correct prefix of $\tau^b$ with the aligned suffix of a good trajectory to form a \emph{revision trajectory} $\tau^r$, e.g., $\tau^r=(\tau^b_{0:t'-1})\circ(\tau^g_{t':m})$. In this way, the evaluation function yields both a coarse trajectory-level judgment and a precise edit location that together produce training pairs emphasizing how to repair failures into successes.

\textbf{MCTS Design:} Agent-R uses Monte Carlo Tree Search not as an inference-time planner but as a data collection engine that systematically explores the action space to yield diverse good/bad trajectories for offline training. The search follows the standard stages (selection, expansion, simulation, backpropagation), with selection guided by the UCT criterion
\[
\mathrm{UCT}(s)=Q(s)+c_{uct}\sqrt{\frac{\log N_p(s)}{N(s)}},
\]
where $Q(s)$ is the average return for state $s$, $N(s)$ its visit count, and $N_p(s)$ the parent's visits; rollouts proceed under a default policy until a terminal state, whose environment reward $r(\tau)$ is then backpropagated to update $Q(s)$ by averaging and to increment $N(s)$ along the traversed path. This usage of MCTS prioritizes breadth and difficulty of experience, furnishing the raw material from which Agent-R constructs revision data that teaches the policy to self-correct.

\begin{center}
\textbf{\centering Retro-Search}
\end{center}

Retro-Search~\cite{lu2025retro} is an MCTS-inspired iterative path revision algorithm rather than a traditional tree search method.

\textbf{Evaluator-Modeling:} The evaluation does not rely on a separately trained reward model. Instead, it uses a ``revision model'' \(\hat{\mathcal{M}}\), which can either be the original reasoning model (in a self-improvement setting) or a weaker student model (in a weak-to-strong setting). This model's role is generative; it produces alternative reasoning trajectories (rollouts) from specific points in an existing path for subsequent evaluation.

\textbf{Evaluation-Function Design:} The quality of a reasoning path is determined by a deterministic value function applied to each step \(s_i\) of the trajectory. The function is defined as \(V(s_i) := \gamma^{N-i}R(a(s_i), a^{*})\), where \(a(s_i)\) is the final answer produced from the trajectory starting at step \(s_i\), \(R\) is a binary function verifying if this answer matches the ground truth \(a^{*}\), \(N\) is the total number of steps in the path, and \(\gamma\) is a decay factor that penalizes longer paths. A new path replaces an old one only if its value is strictly greater, effectively prioritizing shorter, correct solutions.

\textbf{MCTS Design:} The algorithm deviates from standard MCTS by not building an explicit tree or using selection heuristics like UCB. It performs a sequential, greedy revision of a given reasoning trace. The process identifies points where the original model switched its line of thought (e.g., using keywords like ``Alternatively''). At these points, it generates new rollouts by constraining the revision model \(\hat{\mathcal{M}}\) to continue the current thought rather than switching. The resulting trajectory is evaluated, and if it proves more efficient (i.e., has a higher value), it replaces the original path from that point onward. This process repeats for the next thought-switch in the (potentially updated) trajectory. There is no backpropagation of values; decisions are final and greedy.

\begin{center}\textbf{MASTER}\end{center}

\textbf{Evaluator-Modeling.} MASTER~\citep{gan2025master} does not train a separate evaluation model.  Instead, it repurposes the base Large Language Model (LLM) to perform self-evaluation through a structured, multi-step prompting process.  For each generated agent (node), the LLM first executes a \textbf{Validation} step, where it is prompted to verify the key facts within the agent's proposed solution.  Following this, it performs an \textbf{Assessment} step, where it is prompted to generate both a numerical score ($r_0$) indicating progress and a confidence level ($c_0$) for that score. This approach leverages the in-context reasoning capabilities of the LLM itself to serve as the evaluator, avoiding the need for model training and external datasets.  A final \textbf{Evaluation} step is applied only to terminal agents to determine if their solution is correct, which can trigger backpropagation.

\textbf{Evaluation-Function Design.} The evaluation function in MASTER is not a simple reward function but a composite procedure that yields two key values for each agent: an initial reward ($r_0$) and a confidence score ($c_0$). These are extracted from the LLM's textual output during the \textbf{Assessment} phase.  The design is intended to make the reward more reliable by first having the LLM explicitly validate the reasoning steps before assigning a score.  The confidence score ($c_0$) is a crucial component, as it is used to modulate the influence of both the initial reward and the exploration term in the system's modified UCT formula.  For terminal agents that fail the final evaluation, a reward is generated and backpropagated to penalize the preceding reasoning path.

\textbf{MCTS Design.} MASTER introduces a novel adaptation of MCTS tailored for LLMs.  The core modification is the complete \textbf{elimination of the simulation step}, which is traditionally used to estimate long-term rewards. Instead, rewards are derived directly from the LLM's self-assessment at each expansion step. The framework retains the other three MCTS procedures:
\begin{itemize}
     \item \textbf{Selection:} An agent (node) is chosen for expansion based on a modified UCT formula that incorporates the LLM's confidence ($c_0$) to dynamically weigh the initial reward and adjust the exploration term.
     \item \textbf{Expansion:} The selected agent generates a set number of child agents to explore different reasoning paths.
     \item \textbf{Backpropagation:} This step is retained but is only triggered when a terminal agent's final solution fails the evaluation.  The reward from the failed agent is then used to update the Q-values of its ancestors, allowing the system to correct for initially misallocated rewards.
\end{itemize}
This design shifts the computational resources from numerous, costly simulations to a series of refined self-evaluation steps within each node.

\begin{center}\textbf{AB-MCTS}\end{center}

\textbf{Evaluator-Modeling:} AB-MCTS~\citep{inoue2025wider} does not train or model an evaluator. Instead, it presupposes the existence of an external scoring function, $r = R(t_{out})$, which provides direct feedback on a complete, LLM-generated solution candidate $t_{out}$. This function is treated as a black box that returns a score, often normalized to $[0, 1]$, based on task-specific criteria, such as the fraction of passed test cases in a coding challenge. The core method is designed to leverage this external feedback signal for search, rather than modeling the evaluation process itself.

\textbf{Evaluation-Function Design:} The method evaluates actions at a node (either exploring deeper into an existing child's subtree or widening by generating a new child) by modeling the posterior predictive distribution of scores for future nodes. This is implemented in two ways: (1) \textbf{AB-MCTS-M} uses a Bayesian mixed-effects model where each child's subtree is a "group", sharing statistical strength to inform the score distribution of generating a new, unseen child. (2) \textbf{AB-MCTS-A} simplifies this by aggregating all existing children under a `CONT` node and representing new generation with a `GEN` node, modeling the score distribution for each with independent Bayesian models using conjugate priors (e.g., a Beta distribution for scores in $[0, 1]$) for efficient updates. The choice of action is then made via Thompson sampling from these distributions.

\textbf{MCTS Design:} The central innovation is a novel framework that dynamically decides whether to "go wider" (exploration) or "go deeper" (exploitation), enabling adaptive and theoretically unbounded branching. Unlike standard MCTS with a fixed branching factor, AB-MCTS introduces a special `GEN` node at each level of the tree, which represents the action of generating a new child candidate from the current node. The selection policy is not based on UCT but on \textbf{Thompson Sampling}, which naturally balances the choice between selecting an existing child node and selecting the `GEN` node to expand the tree's width based on the Bayesian posterior distributions of expected scores.

\begin{center}\textbf{SELT Introduction}\end{center}

The SELT (Self-Evaluation LLM Tree Search)~\citep{wu2025selt} framework introduces a novel approach to enhance the reasoning capabilities of Large Language Models (LLMs) by integrating a modified Monte Carlo Tree Search (MCTS). Its primary innovation is the elimination of external, pre-trained reward models by leveraging the intrinsic self-evaluation abilities of the LLM itself. By decomposing complex problems into atomic subtasks and employing semantic clustering to guide the evaluation, SELT aims to create a more robust, generalizable, and efficient reasoning process without the need for task-specific fine-tuning.

\textbf{Evaluator-Modeling: }
For its evaluator, SELT repurposes the foundational LLM as an intrinsic, unsupervised \textit{Scorer}, a core design choice that circumvents the dependency on external reward models. This self-evaluation is not performed in a vacuum; its effectiveness is enhanced through a dynamic, reference-based system. At each node in the search tree, SELT performs unsupervised semantic clustering on all previously generated answers to identify distinct, high-quality reasoning paths. From each cluster, a representative answer is selected, and these representatives serve as a contextual benchmark for the LLM Scorer to assess the quality of newly simulated answers.

\textbf{Evaluation-Function Design: }
The evaluation function in SELT produces a reward score, denoted as \(\Delta\), which is generated directly by the LLM Scorer during the simulation phase. Unlike methods that rely on a fixed reward function, SELT's evaluation is dynamic and context-aware. The score \(\Delta\) for a new answer is determined by the LLM's assessment of that answer against the set of representative answers curated through the semantic clustering process. This approach allows the evaluation to be grounded in the diverse and high-quality solutions discovered during the search itself, rather than an abstract or pre-trained notion of correctness. The final score \(\Delta\) is then used in the backpropagation step to update the value of all parent nodes in the traversed path.

\textbf{MCTS Design: }
SELT introduces several significant modifications to the traditional MCTS algorithm. The \textbf{Selection} phase utilizes a custom Upper Confidence Bound for Trees (UCT) formula. The exploitation term, \(S_{LLM\_Exploit}\), is redesigned using Bayesian Averaging to better handle the uncertainty inherent in LLM self-evaluation scores. The exploration term, \(S_{LLM\_Explore}\), is also adjusted to encourage deeper, more focused searches within the reasoning tree. During \textbf{Expansion}, the framework builds out a binary search tree. The \textbf{Simulation} step involves the LLM acting as a `Reasoner` to complete a reasoning path, followed by the clustering and self-evaluation process to generate the reward \(\Delta\). Finally, the \textbf{Backpropagation} phase follows a standard procedure, where the visit count \(N(v)\) and the total reward \(Q(v)\) of each node along the path are updated with the newly calculated score \(\Delta\).

\begin{center}\textbf{TRANS-ZERO}\end{center}

TRANS-ZERO~\citep{zou2025trans} operates as a goal-driven system. The entire search process is optimized to identify the single highest-quality translation for a given source input. Rewards are calculated for complete translation candidates after a comprehensive, multi-step simulation process, rather than being assigned to intermediate steps. The final output of the search is the node (translation candidate) with the highest cumulative utility, reinforcing the framework's focus on achieving a final, high-quality output.

\textbf{Evaluator-Modeling}
TRANS-ZERO does not train a dedicated evaluation model. Instead, it leverages the inherent multilingual capabilities of the base Large Language Model (LLM) for generating translation variations and employs a pre-trained, off-the-shelf text generation metric, BLEURT, to function as the evaluator. This approach bypasses the need for training a separate critic or reward model by defining evaluation as a direct measurement of semantic consistency derived from round-trip translations, making the framework self-contained and reliant only on monolingual data.

\textbf{Evaluation-Function Design}
The evaluation function computes a reward, \(r(y)\), based on the principle of multilingual semantic consistency. For any given translation candidate node, \(y\), the system performs a simulation by rolling out a temporary sub-tree. This involves translating \(y\) through a series of randomly sampled pivot languages and then back to the original source language, generating a set of reconstructions \(\{x_\omega\}\). The final reward is the maximum semantic similarity score, calculated via BLEURT, between these reconstructions and the original source text, effectively measuring how well the meaning is preserved across multiple translation steps.

\textbf{MCTS Design}
The framework introduces a novel Genetic Monte-Carlo Tree Search (G-MCTS), where each node in the tree represents a complete translation candidate. The primary innovation is in the tree expansion phase, which uses two genetic operators to foster diverse exploration. The \textbf{Merge} operator combines the current best-utility node and the best-UCB node as few-shot examples to guide an in-context translation of the original source text. The \textbf{Mutate} operator promotes creative exploration by translating a semantically similar variant of the source text-specifically, a reconstruction generated during a previous simulation-instead of the original input.

\begin{center}
\textbf{CMCTS}
\end{center}

Complementing the constrained action space, CMCTS~\citep{lin2025leveraging} integrates a set of human-like partial order rules during the simulation phase to ensure logical coherence in the reasoning chain. These rules impose constraints on the sequence of actions, such as mandating an "understand" action at the beginning and a "summary" action at the end. Further rules govern action diversity, the necessity of reflection, and the strategic use of coding actions based on reasoning depth. These rules can be used independently or in combination with the PRM to guide the search, preventing illogical or redundant state transitions. The overall MCTS process follows the standard selection, expansion, simulation, and back-propagation phases, but with these novel constraints and guidance mechanisms integrated to produce higher-quality, long-chain-of-thought reasoning.

\textbf{Evaluator-Modeling}
The CMCTS framework utilizes a pre-trained Process Reward Model (PRM) as its primary evaluator, forgoing the need for training a new model. This PRM, specifically the Qwen2.5-Math-PRM, is designed to assess the quality of intermediate reasoning steps. It functions by taking a given state \(s_t\) and a potential subsequent action \(a_{t+1}\), which are concatenated with a specialized prompt template. The model then processes this combined input to produce logits for "positive" and "negative" outcomes. This mechanism allows the PRM to provide nuanced, context-aware evaluations of reasoning quality without relying on the base Large Language Model (LLM), which is often an unreliable reward signal. The framework uses this external, specialized model to guide the search process toward more rational and effective reasoning paths.

\textbf{Evaluation-Function Design}
The evaluation function in CMCTS is bifurcated into two components: an action-value function \(Q(s_t, a_{t+1})\) and a state-value function \(V(s_t)\), both derived from the PRM. The action-value \(Q(s_t, a_{t+1})\) is calculated by applying a softmax function to the PRM's output logits, yielding the probability of a "positive" assessment for taking action \(a_{t+1}\) in state \(s_t\). Similarly, the state-value \(V(s_t)\) is computed in an action-agnostic manner, representing the intrinsic quality of a given reasoning state \(s_t\). During back-propagation, the reward for a specific transition is defined as the sum \(r_t = Q(s_{t-1}, a_t) + V(s_t)\), which combines the value of the chosen action and the resulting state. This dual-evaluation approach provides a comprehensive signal for updating the cumulative rewards of nodes in the search tree.

\textbf{MCTS Design}
The MCTS design in CMCTS introduces two primary innovations to the standard algorithm: a constrained action space and partial order rules. Unlike traditional methods where the LLM generates subsequent actions, CMCTS samples actions from four predefined, disjoint sets: \(\mathcal{A}^{\text{understand}}\), \(\mathcal{A}^{\text{reflect}}\), \(\mathcal{A}^{\text{code}}\), and \(\mathcal{A}^{\text{summary}}\). This action space constraining, applied during the expansion phase, forces the model to explore diverse and semantically rich reasoning states that are otherwise difficult to sample, such as self-correction and code-based verification. This directly addresses the issue of state-space homogenization common in other MCTS applications with LLMs.

\section{Challenges and Future of Tree-Search Methods}
\label{appendix:Challenges}

\paragraph{Search Efficiency and Intelligence}. 
Tree search algorithms, despite their power, often require significantly greater computational resources than greedy decoding, as noted by \cite{wang2024litesearch}, with resource demands exceeding 10 times that of greedy approaches in certain cases due to inefficiencies in search strategies. This high computational overhead presents a substantial barrier to the practical deployment of these methods. Algorithms like MCTSr and LLaMA-Berry, which generate multiple solutions sequentially at each node, exacerbate these resource demands. To mitigate these limitations, future research could prioritize improving the efficiency of tree search algorithms by investigating trade-offs between policy and reward models, incorporating dynamic control mechanisms, and employing effective pruning techniques to optimize tree expansion.

\paragraph{Overthinking Issues in Simple Queries.}
Task complexity is closely related to the length of reasoning chains, highlighting the need for extended cognitive processing in more difficult problems~\citep{qin2024o1,huang2025o1}. However, \cite{chen2024not} and \cite{zeng2024scaling} observe that O1-like models often overanalyze simple questions, dedicating excessive computational resources to tasks that have clear and obvious answers. For instance, a query like "3-2=?" does not require complex reasoning, yet these models may engage in unnecessary computations, wasting resources and potentially introducing errors. Forcing models to reason through such trivial tasks not only consumes valuable computational power but also causes delays. Future research should focus on methods to reduce these inefficiencies, improving models' ability to quickly recognize and handle straightforward queries while dynamically allocating computational resources across diverse problem types.

\paragraph{Self-play Between Policy Models and Reward Models.} 
Certain tree-search algorithms encounter challenges due to limited parallelism, which constrains their search speed, especially in resource-intensive settings.
As detailed in Section~\ref{appendix:MCTS}, various tree-search techniques can generate traces that are then employed to iteratively refine reward and policy models, such as ReST-MCTS and rStar-Math. This self-play paradigm is crucial for internalizing the reasoning system into the policy model, thereby endowing LLMs with sophisticated reasoning abilities~\citep{xiang2025towards}.  By internalizing tree-search reasoning into LLMs, the tree-search process can be structured within a CoT framework, facilitating sequential reasoning. This not only enhances reasoning efficiency but also mitigates parallelism limitations, thereby improving scalability. Future research should investigate strategies to optimize this self-play paradigm further, facilitating more efficient problem-solving.

\paragraph{Reward Modeling and Reward Model Training.} 
Section~\ref{appendix:MCTS} examines various MCTS-based evaluation strategies. A central element of the search strategies is the reward or evaluation model, which provides essential supervision to guide search processes effectively~\citep{lightman2023let,setlur2024rewarding,xiang2025towards}. 
Reward models are broadly categorized into two types: the Outcome Reward Model (ORM) and the Process Reward Model (PRM). Unlike outcome rewards, which deliver feedback only at the task's conclusion, process rewards provide signals at both intermediate steps and the final outcome, enabling finer-grained and more frequent supervision. Nevertheless, learning process rewards present significant challenges. For example, \cite{uesato2022solving,lightman2023let} relies on human annotators for process supervision, a costly and inherently unscalable method. While automated methods for constructing process rewards have been proposed~\citep{wang2024math,luo2024improve,wang2024multi}, they are predominantly designed for specialized areas such as mathematics and programming. These approaches struggle to generalize to broader domains, such as scientific reasoning and complex problem-solving, where human evaluation remains essential. Overcoming these limitations necessitates the development of more efficient methods to generate high-quality fine-grained rewards and scalable techniques to advance reward model capabilities, which remain open and pressing research challenges.

\paragraph{Reward Model Quality and Its Effect on Search.}
The performance and efficiency of search during testing depend on the quality of the Process Reward Model (PRM)\citep{setlur2024rewarding,xiang2025towards}. However, searches guided by an oracle verifier are more efficient than those relying on a learned PRM. Numerous studies have shown that an imperfect reward model can give rise to inverse inference scaling \citep{zeng2024scalingsearchlearningroadmap}. For instance, \cite{gao2023scaling} identified an inverse scaling effect, where expanding the search space in best-of-n search negatively impacts performance due to a distribution shift between the imperfect reward model and the policy model. These findings underscore the critical need to bridge the performance gap between oracle and learned reward models. \cite{xiang2025towards} shows that while the PRM's ability to verify complete solutions improves with additional data, a notable gap persists between trained PRMs and oracle PRMs. Therefore, understanding how scaling laws for process supervision models influence their effectiveness and efficiency in large-scale search tasks remains a pivotal challenge.

\section{The Use of Large Language Models (LLMs)}
\label{llm}

Large Language Models (LLMs) were used as assistive tools in the preparation of this work. Specifically, we employed GPT-5 to make minor edits to academic writing, such as drafting and refining sections. All scientific claims, methodological contributions, and experimental results were conceived, implemented, and validated by the authors. The authors take full responsibility for the content presented in this paper.

\end{document}